\documentclass[letterpaper, 10 pt, journal, twoside]{IEEEtran} %final RA-L version

\IEEEoverridecommandlockouts                              % This command is only needed if 
\usepackage{soul} %for highlighting

\usepackage[usenames, dvipsnames]{color} %for highlighting in colour (used with soul)
\usepackage{graphicx} %for loading graphic files
\usepackage[cmex10]{amsmath} %extra command is for IEEE format
\usepackage{amsfonts}
\usepackage{amssymb}
\usepackage{verbatim} %for \comment
\usepackage{subfigure}
\usepackage{multirow}
\usepackage{cite}
\usepackage{hhline}
\usepackage{bm}
\usepackage{tabularx,booktabs}
  \newcolumntype{Y}{>{\centering\arraybackslash}X}
\usepackage[bottom]{footmisc}

\usepackage{algorithm}
\usepackage[noend]{algpseudocode}

\usepackage[printonlyused,nolist,nohyperlinks]{acronym}
\soulregister\cite7
\soulregister\ref7

\newcommand*{\vrectangleA}{{\ooalign{\lower.3ex\hbox{$\sqcup$}\cr\raise.4ex\hbox{$\sqcap$}}}}

\begin{document}
 %\addtolength{\topmargin}{50pt}
\title{\LARGE \bf Sensor Observability Analysis for Maximizing Task-Space Observability of Articulated Robots}
%Sensor Observability Analysis of Robotic Mechanisms
%Task-Space Sensor Observability Analysis of Articulated Robotic Mechanisms
%Sensor Observability Optimization for Maximizing Task-Space Observability

\author{Christopher~Yee~Wong, \emph{Member, IEEE} and Wael~Suleiman, \emph{Senior Member, IEEE}% <-this % stops a space
%\thanks{Manuscript received July 20, 2018. }% <-this % stops a space
\thanks{This work was supported in part by the Fonds de Recherche du Qu\'{e}bec - Nature et technologies and the Natural Sciences and Engineering Research Council of Canada (NSERC). (\emph{Corresponding author: Christopher Yee Wong.})}
\thanks{C. Y. Wong and W. Suleiman are with the Universit\'{e} de Sherbooke, Sherbrooke, Canada (e-mail: \texttt{christopher.wong2}, \texttt{wael.suleiman (at) usherbrooke.ca}).}
}
%Chris Papercept PIN: 195779
%Wael Papercept PIN: 113066

\maketitle
%\thispagestyle{empty} %to remove page numbers on first page
%\pagestyle{empty} %to remove page numbers on other pages

%\linenumbers
%%%%%%%%%%%%%%%%%%%%%%%%%%%%%%%%%%%%%%%%%%%%%%%%%%%%%%%%%%%%%
%%%%%%%%%%%%%%%%%%%%%%%%%%%%%%%%%%%%%%%%%%%%%%%%%%%%%%%%%%%%%
%A one−paragraph abstract of not more than 200 words must accompany each manuscript. It should state concisely the reason for the study, what was done, what was found, what was concluded, and the relevance. 
\begin{comment}

\end{comment}
\begin{abstract}
We propose a novel performance metric for articulated robots with distributed directional sensors called the \emph{sensor observability analysis} (SOA).
These robot-mounted distributed directional sensors (e.g., joint torque sensors) change their individual sensing directions as the joints move. 
SOA transforms individual sensors axes in joint space to provide the cumulative sensing quality of these sensors to observe each task-space axis, akin to forward kinematics for sensors.
For example, certain joint configurations may align joint torque sensors in such a way that they are unable to observe interaction forces in one or more task-space axes.
The resultant sensor observability performance metrics can then be used in optimization and in null-space control to avoid sensor observability singular configurations or to maximize sensor observability in particular directions.
We use the specific case of force sensing in serial robot manipulators to showcase the analysis.
Parallels are drawn between sensor observability and the traditional kinematic manipulability; SOA is shown to be more generalizable in terms of analysing non-joint-mounted sensors and can potentially be applied to sensor types other than for force sensing.
Simulations and experiments using a custom 3-DOF robot and the Baxter robot demonstrate the utility and importance of sensor observability in physical interactions.

\end{abstract}

\begin{IEEEkeywords} %After the abstract, list three to ten terms not included in the title. 
\begin{comment}
IROS KEYWORDS:
#1 KH Formal Methods in Robotics and Automation
#2 AK Kinematics
#3 FC Robot Safety

DM Collision Avoidance
RS Methods and Tools for Robot System Design
\end{comment}
Force and tactile sensing, 
%formal methods in robotics and automation,
kinematics,
%methods and tools for robot system design,
% Reactive and Sensor-Based Planning
% Force and Tactile Sensing
physical interaction,
robot safety

%Theoretical Foundations
%
%Calibration and Identification
%Direct/Inverse Dynamics Formulation
%Formal Methods for Robotics
%Kinematics
%Optimization and Optimal Control
%Probability and Statistical Methods
\end{IEEEkeywords} 

%http://www.ieee.org/documents/taxonomy_v101.pdf
%\IEEEpeerreviewmaketitle

%%%%%%%%%%%%%%%%%%%%%%%%%%%%%%%%%%%%%%%%%%%%%%%%%%%%%%%%%%%%%
%%%%%%%%%%%%%%%%%%%%%%%%%%%%%%%%%%%%%%%%%%%%%%%%%%%%%%%%%%%%%
\section{Introduction} \label{sec:intro}
\IEEEPARstart{S}{ensors} are invaluable tools for robots as they are their way to observe themselves (introspection) and the world around them (extrospection). %, even in ways beyond the capability of humans. 
Unfortunately, sensors have limitations beyond their technical specifications, particularly directional sensors. 
Directional sensors are those with explicit axes along which measurements are performed, for example joint torque sensors, strain gauges, accelerometers, gyroscopes, distance sensors, cameras, etc. 
%load cells, force sensitive resistors, 
While optimal sensor placement is an active area of research for mobile robots and sensor networks \cite{Salaris2019TRO-PerceptionAwareTrajGeneration, Hu2022CVPR-SensorPlacementAutonomousCar}, 
the same cannot be said for articulated and reconfigurable robots \cite{Bonev2001TRA-SensorPlacementParallelRobot}.
It is often a common assumption for articulated and reconfigurable robots that, given the presence of sensors, all task-space quantities are fully observable at all times.
Consider a serial robotic manipulator with joint torque sensors at each joint; one would normally assume that the end effector (EE) forces could be reconstructed from the joint torque sensors \cite{VanDamme2011ICRA-EEforceestimationJTS, Phong2012ICRA-EEforceestimationJTS} for use in compliant control either directly \cite{Wu1980-ManipComplianceJointTorqueControl} or through machine learning \cite{Berger2020ICHR-LearningForceEstimation}.
In fact, this assumption is not always true. 
It is possible that particular robot configurations lead to cases where the joint torque sensors are unable to observe certain interaction forces at the end effector \cite{Gosselin1990TRO-SingularityAnalysisofClosedLoopKinematicChains}.
A similar assumption may be present in cases where a robot is equipped with an array of distributed distance \cite{Stavridis2020ICHR-MobileManipDistributedDistanceSensors}, proximity \cite{Xiao2022TRO-TactileWhiskers}, or contact \cite{Albini2021RAL-ExplorationUsingArmTactileSensors, Laliberte2022TRO-LinkBasedSensorforpHRI} sensors on the arms.  
If these sensors are sparsely placed, then certain robot configurations may lead to potentially unobserved directions.

If a robot unknowingly enters a configuration where observations along certain external task-space axes can no longer be made, the end result could be disastrous for the robot, the task, or the environment.
These situations must be avoided in especially critical applications such as physical human-robot interaction and minimally invasive surgery. 
Although the use of multi-axis sensors, e.g. 6-axis force-torque sensors and 3-axis accelerometers, automatically observes all possible task-space axes and renders the issue of task-space observability trivial, these advanced sensors may not always be available.
For example, 6-axis force-torque sensors are not available for many lower-cost systems given their cost or their size makes them infeasible to mount, and alternative methods must be used \cite{Hawley2019IJHR-ExtForceObsforSmallHumanoids}.
Additionally, interactions with the robot body may not be observed if the sensor is mounted at the end effector.

\begin{figure}[t]
\centering
\subfigure[]{
	\includegraphics[width=0.35\textwidth]{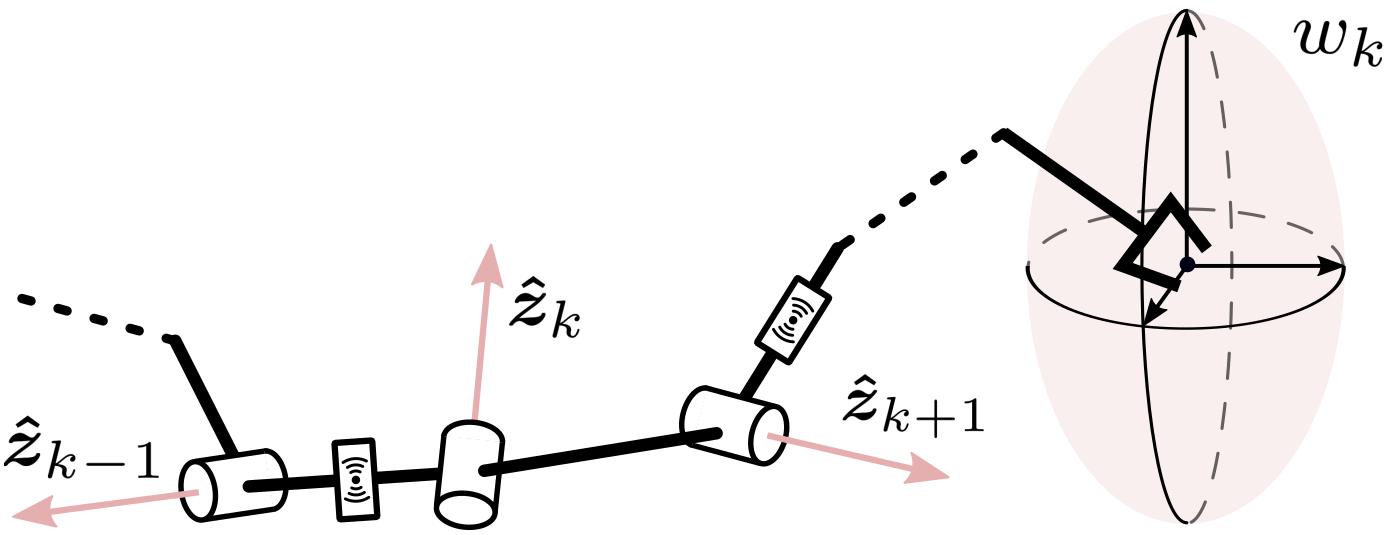}
	\label{fig:intro-kin}}
\subfigure[]{
	\includegraphics[width=0.37\textwidth]{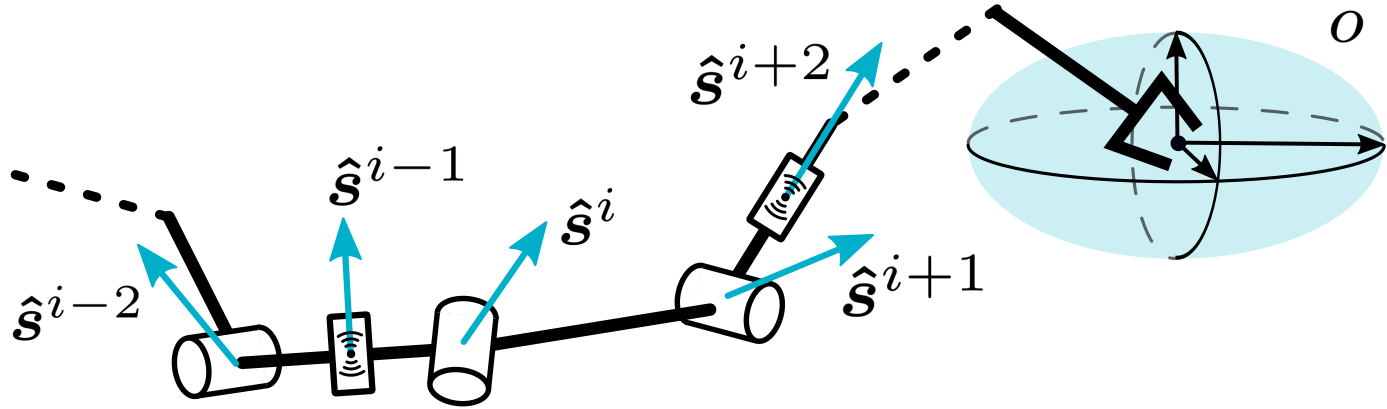}
	\label{fig:intro-sens}}
\caption{Comparison of a) joint axes $\bm{\hat{z}}_k$ and kinematic manipulability $w_k$ and b) positioning of joint-mounted and link-mounted sensors $\bm{\hat{s}}^i$ and the sensor observability $o$ and their respective ellipsoids for the same robot configuration.}
 %$\bm{\hat{s}}^{i-2} \ \bm{\hat{s}}^{i-1} \ \bm{\hat{s}}^{i} \ \bm{\hat{s}}^{i+1} \ \bm{\hat{s}}^{i+2} \ \bm{\hat{z}}_{k-1} \ \bm{\hat{z}}_{k} \ \bm{\hat{z}}_{k+1} \ w_k \ o$
\label{fig:smi-calc}
\end{figure}

Thus, a tool is required to analyse a particular robot configuration and provide a measure or index on the \textit{task-space observability} of the configuration.
Parallels can be drawn with the various kinematic performance measures used with serial robots \cite{Patel2015JIRS-SurveyManipulatorPerfMeasures}, particularly the well-known concept of kinematic robot manipulability \cite{Yoshikawa1985IJRR-Manipulability}, a quality measure of a robot's mobility and closeness to kinematic singularities.
In this paper, we extend this concept to task-space observability of robots based on the robot joint configuration and sensor placement, which may differ from the joint axes.
As this issue of task-space sensor observability is highly dependent on the sensor configuration and kinematic structure of the robot, not all robots are equally affected.
Such an analysis is especially important for robots that do not have enough sensors to cover all task space dimensions reliably and must prioritize one over others.
According to the semantics defined in \cite{Patel2015JIRS-SurveyManipulatorPerfMeasures}, sensor observability analysis is classified as a local kinematic and intrinsic performance index.

%Obs is important for under-actuated robots (if $n_q < n_t$) if you want to prioritize certain directions
%
%To analyze sensor observability, we draw parallels with the traditional robot manipulator manipulability analysis.
%There is an abundance of literature on this topic, 
%
%\hl{THIS DOES NOT HAVE TO BE FOR FORCE SENSORS, CAN BE ANY TYPE OF SENSOR??}
%

%It has often been proposed that learning by demonstration is a solution to this problem (Schaal, Ijspeert, \& Billard, 2003), and kinesthetic teaching (i.e., physical guidance of the robot by the users) has been considered as an interface for providing demonstrations (Billard, Calinon, Dillmann, \& Schaal, 2007). In this context, kinesthetic teaching typically provides data for learning task-space movements (Calinon \& Billard, 2007a) or a particular task trajectory, including a redundancy resolution (Vijayakumar, D’souza, Shibata, Conradt, \& Schaal, 2002).

%Force sensing is extremely important in performing tasks and for human-robot interaction

%%%%%%%%%%%%%%%%%%%%%%%%%%%%%%%%%%%%%%%%%%%%%%%%%%%%%%%%%%%%%
\subsection{Background}
\label{sec:background}

To note, vectors and matrices are represented by bold-faced lower case and upper case letters, respectively, whereas scalar values are not.

The traditional analytical Jacobian matrix $\bm{J(q)} \in \mathbb{R}^{{n_t} \times {n_q}}$ is defined as the matrix of first order partial derivatives relating ${n_q}$ joint-space velocities $\bm{\dot{q}}$ to ${n_t}$ task-space velocities $\bm{\dot{x}}$ \cite{Spong2020-RobotModelingandControl}:

\begin{equation}
\bm{\dot{x}} = \bm{J(q)}\bm{\dot{q}}, \ \ \bm{J(q)} = \begin{bmatrix}
  \frac{\partial x_1}{\partial q_1} & 
    \dots & 
    \frac{\partial x_1}{\partial q_{n_q}} \\[1ex] % <-- 1ex more space between rows of matrix
  \vdots & 
    \ddots & 
    \vdots \\[1ex]
  \frac{\partial x_{n_t}}{\partial q_1} & 
    \dots & 
    \frac{\partial x_{n_t}}{\partial q_{n_q}}
\end{bmatrix}
\label{eq:Jreg}
\end{equation}
%\noindent where $\bm{x}(j)$ indicates the $j$-th element of $\bm{x}$ in $\mathbb{R}^{n_t}$.
In typical cases, the task space is defined as end effector position (${n_t} = 3$) or pose (${n_t} = 6$).
For readability, we will continue the manuscript without explicitly writing the Jacobian's dependency on the vector of joints $\bm{q}$, in other words $\bm{J(q)} \rightarrow \bm{J}$. 
The Jacobian can be used as a tool to measure different properties of the robot, notably to verify whether a specific robot configuration is at a \emph{kinematic singularity}.

The \emph{kinematic manipulability index} $w_k$, commonly referred to as simply the \emph{manipulability}, is a scalar quality measure of the robot's ability move in the task space based on the current joint configuration. 
Manipulability can also be used as a scalar measure of a robot's closeness to a kinematic singularity \cite{Yoshikawa1985IJRR-Manipulability}. 
It is an important tool that allows postures to be evaluated based on their mobility:

\begin{equation}
w_k = \sqrt{det(\bm{JJ}^T)}% = \sigma_1 \sigma_2 \cdots \sigma_{n_t}
\label{eq:manip_kin}
\end{equation}
%\noindent where $\sigma_i$ are the \hl{singular} values.

%Similarly, the \emph{dynamic manipulability index} $w_d$ \cite{Yoshikawa1985ICRA-DynamicManipulability} measures the ability of the end effector to accelerate in specific directions in the task space for a given set of joint torques if the inertial effects cannot be ignored. 
%\hl{(remove $w_d$?)}

%\begin{equation}
%w_d = \sqrt{det(\bm{J}(\bm{M}^T\bm{M})^{-1}\bm{J}^T)}
%\label{eq:manip_dyn}
%\end{equation}

%\noindent where $\bm{M}$ is the inertial matrix.
The manipulability index can be exploited in different ways, typically with optimization algorithms to ensure that the robot motions stay away from any singularities \cite{Dufour2020JIRS-ManipulabilityQP}.
 %and $w_d$
While the index $w_k$ is a scalar measure, the manipulability \emph{ellipsoid}, as described in \cite{Yoshikawa1985IJRR-Manipulability, Chiu1987ICRA-ManipEllipsoids} and shown in Fig. \ref{fig:intro-kin}, is a volumetric representation of mobility for a specific robot configuration that is proportional to the length of the ellipsoid principal axes.
Mobility indicates the ease with which the end effector can move in a certain direction in task space proportional to joint motion.
As such, the manipulability ellipsoid itself can be used as a target either as the main task or as a redundancy resolution sub-task \cite{Jaquier2018RSS-TrackingManipulabilityEllipsoids, Jaquier2021IJRR-GeometryAwareManipulabilityLearning}.
Controlling the manipulability ellipsoid ensures that a certain level of manipulability is present, especially if a particular shape is desired.

The concept of manipulability has greatly evolved to encompass different calculation methods and applications since its early introduction by Yoshikawa \cite{Yoshikawa1985IJRR-Manipulability}. 
For example, some authors modified the concept to instead calculate the manipulability of the centre of mass of floating base robots \cite{Gu2015ICRA-COMManipulabilityHumanoids, Azad2017ICRA-DynManipCOM}.
Manipulability has also been extended to multi-robot closed-chain systems \cite{BicchiTRA2000-ManipulabilityClosedChain} and continuum robots \cite{Gravagne2002TRA-ManipulabilityContinuum}.

%%%%%%%%%%%%%%%%%%%%%%%%%%%%%%%%%%%%%%%%%%%%%%%%%%%%%%%%%%%%%
\subsection{Manuscript Organization and Contributions}
\label{sec:contributions}

In the same vein as the kinematic manipulability index and ellipsoid, we introduce the novel concept of \emph{sensor observability analysis} and the resulting \emph{sensor observability index} and \emph{sensor observability ellipsoid}.
 %of the \emph{sensor observability} and \emph{sensor manipulability indices}, as well as the \emph{sensor observability ellipsoid}.
The proposed concepts qualitatively evaluate, based on the current joint configuration, the cumulative ability of distributed directional sensors on an articulated robot to measure external quantities in the task space, akin to forward kinematics but for sensors. 
While the analysis of distributed sensing is thoroughly studied in sensor networks and swarm robotics, the analysis introduced here is from the viewpoint of a single multi-jointed and articulated robot.
As an example, the type of directional sensors to be analysed may include force-sensing elements, accelerometers, or distance sensors.
Sensor observability analysis would then provide a performance metric to determine if the onboard sensors are able to observe interaction forces, accelerations, or object distance in all directions, or if the current configuration is potentially blind to forces, accelerations, or objects in a certain direction. 
The derivations in this paper use force sensing as the case study for the observability of end effector forces \cite{VanDamme2011ICRA-EEforceestimationJTS, Phong2012ICRA-EEforceestimationJTS} as it is the most intuitive case.
Other types of axis-based sensors, like those mentioned above, will be explicitly developed in future work. 
The proposed formulation also allows the analysis of non-joint-mounted sensors, for example strain gauges, accelerometers, or distance sensors placed on a link, as shown in Fig. \ref{fig:intro-sens}.
We also perform simulations and experiments to demonstrate the differences between sensor observability analysis and traditional kinematic analysis and present certain cases where sensor observability is superior.
%To the best of the authors' knowledge, the framework established here has not been proposed previously,

%\hl{[insert manuscript organization here]}
%%%%%%%%%%%%%%%%%%%%%%%%%%%%%%%%%%%%%%%%%%%%%%%%%%%%%%%%%%%%%
%\subsection{Manuscript Organization}
%\label{sec:organization}

%\hl{fix references and completely rewrite. Maybe put before contributions (or merge)} \hlc{It can be before the contribution section. In this case, we can link each contribution to its section}
This paper first provides the analytical framework in Sec. \ref{sec:method} for defining task-space sensor observability based on the cumulative transformations of each individual sensor.
Discussions surrounding the peculiarities of sensor observability analysis, including analogies to the traditional Jacobian, thresholds for sensor observability, and sensor observability applied to non-traditional robot architectures, are explored in Sec. \ref{sec:specialstuff}.
Next in Sec. \ref{sec:SOoptimization}, we discuss how sensor observability can be used as a secondary task in the nullspace of the kinematics task or formulated as an optimization problem. 
Sec. \ref{sec:physicalexperiments} showcases practical implications of sensor observability analysis during physical interaction with a robot.
The utility of sensor observability is showcased through simulations and experiments on both a custom 3 degree of freedom (DOF) robot and the Baxter robot.
Finally, closing comments and future research directions are provided in Sec. \ref{sec:concl}.

A preliminary version of this paper was presented at a conference \cite{Wong2022IROS-SOIconf}. 
This article is a continuation where the concept of sensor observability is expanded to provide a fuller picture as the first major work of this framework.
Secs. \ref{sec:intro} to \ref{sec:SOI-Advantages} were previously presented in \cite{Wong2022IROS-SOIconf}, but more insights have been added throughout these sections to provide a deeper analysis of the sensor observability framework.
However, Sec. \ref{sec:noise} and onwards are all novel additions to further establish the utility of sensor observability both theoretically and practically.

 %\hlc{(detail which sections are new? (change it in the cover letter too)) (like Sensor Observability Threshold and Sensor Noise, Sec. \ref{sec:noise} and ``analysis of SOI applied to non-joint mounted robots'')} and significantly more examples of how sensor observability can be practically used.
%\hlc{see how Laliberte 2022TRO formats end of intro}
%\hlc{everything from Sec III-C onwards (SO threshold/noise) is new}

%%%%%%%%%%%%%%%%%%%%%%%%%%%%%%%%%%%%%%%%%%%%%%%%%%%%%%%%%%%%%%
%\subsection{Differences versus State-Space Observability} \label{sec:diff}
%The terms \emph{observable} and \emph{observability} used in the context of sensor observability differ from that used in state-space representation.
%State-space observability refers to \emph{state} observation---estimation of the internal system states based on the system output alone \cite{Spong2020-RobotModelingandControl}.
%Instead, sensor observability only concerns the measurement of task-space quantities and not the measurement of system states nor does it require the full system dynamics to be known.
%Further comparisons of two will be the subject of future work.

%%%%%%%%%%%%%%%%%%%%%%%%%%%%%%%%%%%%%%%%%%%%%%%%%%%%%%%%%%%%%
%%%%%%%%%%%%%%%%%%%%%%%%%%%%%%%%%%%%%%%%%%%%%%%%%%%%%%%%%%%%%
%\section{Sensor Transformation, Observability Index and Manipulability Index}
\section{Sensor Observability Analysis}
\label{sec:method}

Prior to introducing the concept of sensor observability, we would like to note that the analysis presented here assumes that joints do not have mechanical limits, and thus in this work we ignore special treatment of sensors that may be affected by joint limits. 
Furthermore, we assume that sensors are bidirectional, in the sense that they are capable of measuring along both positive and negative directions of their sensing axis (for example, laser-based distance sensors are unidirectional, whereas accelerometers and joint torque sensors are bidirectional).
Unidirectional sensors require more complex sensor axis analyses and will be addressed in future work.

%As per Newton's 3rd law ``when two bodies interact, they apply forces to one another that are equal in magnitude and opposite in direction'', 
Particularly in the case of force detection, barring dynamics and inertial effects, there must be an equal and opposite reaction force to properly detect forces, e.g. constraint forces from ground contact. 
As such, fixed base robots have full constraint forces in all directions. 
Conversely, floating base and mobile robots do not always have the luxury of perfect constraint forces. 
Friction cones must be taken into account and any slippage or lack of adequate friction forces will affect force detection and control \cite{Samadi2021RAL-HumanoidControlSlidingContacts}.
Thus, to simplify this initial analysis of sensor observability, we will only consider fixed base open kinematic chain serial manipulators for the time being to remove the question of imperfect constraint forces.
Floating base robots and slippage will be examined in future work.
A summary of the method is presented in Algorithm \ref{alg:sens_obs}.

\begin{algorithm}[t]
\caption{Summary of Sensor Observability (SO) \\ NB: Sensor $i \in 1...n_s$ and task-space axis $j \in 1...n_t$}
\begin{algorithmic}[1]

%\Procedure{Initialization}{}
    \State $\bm{\hat{s}}^{\prime,i}$ for $i \in 1...n_s$  \Comment{Def. local sensor axes (Sec \ref{sec:localsensoraxis})}
    \State $\bm{\hat{s}}^i \leftarrow \bm{R}\bm{\hat{s}}^{\prime,i}$ \Comment{Rotate to match task frame}
		\State $\bm{\tilde{s}}^i \leftarrow T_\square(\bm{\hat{s}}^i, \bm{r}^i)$ \Comment{Sensor-type transf. (Sec \ref{sec:sensortranform})}
    \State $\tilde{s}^i_j = f(\tilde{s}^i_j, s^{i,*}_j)$  \Comment{Noise thresholding (Sec \ref{sec:noise})}
		\State $\bm{S} = \begin{bmatrix} \bm{\tilde{s}}^1 \cdots \bm{\tilde{s}}^{n_s} \end{bmatrix}$  \Comment{SO matrix (Sec \ref{sec:sensobs})}
		\State $\bm{s} \leftarrow \Gamma_\square(\bm{S})$ \Comment{SO func. \& System SO (Sec \ref{sec:sensobs})} % (\S\ \ref{sec:sensobs})
		%\State $\bm{s} \leftarrow \Gamma_\square(\{\bm{\tilde{s}}^i\})$ \Comment{Calculate sens. obsv. (Sec \ref{sec:sensobs})}
		\State $o \leftarrow \prod_{j=1}^{n_t} \bm{s}_j$ \Comment{SO index (Sec \ref{sec:sensobs})}
		%\State Calculate $\bm{J_s(q)}$ \Comment{\hlc{Sensor Jacobian (Sec }\ref{sec:sensorJacob})}
		%\State Calculate $w_s$ \Comment{\hlc{Sensor manipulability index}}
		%\While{$something \not= 0$}
		%
    %\If{$condition = True$} 
        %\State Do this
        %\If{$Condition \geq 1$}
        %\State Do that
        %\ElsIf{$Condition \neq 5$}
        %\State Do another
        %\State Do that as well
        %\Else
        %\State Do otherwise
        %\EndIf
    %\EndIf
%
    %\While{$something \not= 0$}  \Comment{put some comments here}
        %\State $var1 \leftarrow var2$  \Comment{another comment}
        %\State $var3 \leftarrow var4$
    %\EndWhile  \label{roy's loop}
%\EndProcedure

\end{algorithmic}
\label{alg:sens_obs}
\end{algorithm}

%%%%%%%%%%%%%%%%%%%%%%%%%%%%%%%%%%%%%%%%%%%%%%%%%%%%%%%%%%%%%
\subsection{Local Sensor Axis $\bm{\hat{s}}^{\prime,i}$ and Rotated Sensor Axis $\bm{\hat{s}}^i$}
\label{sec:localsensoraxis}
 %as a set of unit basis vectors in $\mathbb{R}^{n_t}$
First, for each individually measured sensor axis $i \in 1...n_s$, as seen in Fig. \ref{fig:smi-calc}, we define a \emph{local} sensor axis vector $\bm{\hat{s}}^{\prime,i} \in \mathbb{R}^{n_t}$, where each element indicates whether a task-space axis is observed or not by taking on a value between $[0,1]$.
Note the difference between $n_s$, the number of sensor axes, and $n_t$, the number of task space axes.
A zero value means that that particular axis is not observed, whereas a value of one means that the axis is directly observed, i.e. the sensor axis is parallel with the task-space axis.
Values between 0 and 1 mean that the task space axis is only partially observed by an off-axis sensor\footnote{The term “partially observed“ indicates that the sensor axis is not completely in line with the task space axis. For example, if a load cell is oriented at an angle $\theta$ from the $x$-axis in the $xy$-plane and a force is applied along the $x$-axis, the sensor will only detect the component of force that is projected along the sensor axis, i.e., $F_{observed} = F_{actual} cos(\theta)$. Similarly, if a laser distance sensor is used to measure the velocity of an object, then the sensor will only detect the component of velocity that is projected along the sensor axes.}.
For example, a single one-axis joint torque sensor could be seen as an element of $ SE(3)$ with $n_t = 6$ and represented by: % a set of a single basis vector

\begin{equation}
\bm{\hat{s}}^{\prime}_{\tau z} = \begin{bmatrix} \bm{\hat{s}}_{p,{\tau z}}^{\prime} \\ \bm{\hat{s}}_{\theta, {\tau z}}^{\prime} \end{bmatrix} = \begin{bmatrix} 0 & 0 & 0 & 0 & 0 & 1 \end{bmatrix}^T
\label{eq:s_JTS}
\end{equation}

\noindent where $\bm{\hat{s}}^{\prime}_{\tau z}$ is in the local joint frame according to Denavit-Hartenberg (DH) parameters \cite{denavit1955TAM-DHparam} and $(\cdot)_p$ and $(\cdot)_\theta$ subscripts are the translational and rotational components, respectively. Similarly, a single axis load cell in the $x$-axis is represented by $\bm{\hat{s}}^{\prime}_{fx} = \begin{bmatrix} 1 & 0 & 0 & 0 & 0 & 0 \end{bmatrix}^T$.
Multi-axis sensors, e.g. a 3-axis load cell that can detect forces in the $xyz$-axes but not torques, would be represented by the set of three individual sensor axis vectors, one in each $x$-, $y$-, and $z$-axis, \textit{i.e.} $\{\bm{\hat{s}}^{\prime}_{fx}, \bm{\hat{s}}^{\prime}_{fy}, \bm{\hat{s}}^{\prime}_{fz}\}$.
%with a sensor axis representation of
%$\bm{\hat{s}}^{\prime}_{f,xyz} = \bm{\hat{s}}^{\prime}_{fx}+ \bm{\hat{s}}^{\prime}_{fy}+ \bm{\hat{s}}^{\prime}_{fz} = \begin{bmatrix} 1 & 1 & 1 & 0 & 0 & 0 \end{bmatrix}^T$
In the same vein, a 6-axis force-torque sensor would be the set of six individual sensor axes represented by $\{\bm{\hat{s}}^{\prime}_{fx}, \bm{\hat{s}}^{\prime}_{fy}, \bm{\hat{s}}^{\prime}_{fz}, \bm{\hat{s}}^{\prime}_{\tau x}, \bm{\hat{s}}^{\prime}_{\tau y}, \bm{\hat{s}}^{\prime}_{\tau z} \}$.
%$\bm{\hat{s}}^{\prime}_\text{6-axis} = \begin{bmatrix} 1 & 1 & 1 & 1 & 1 & 1 \end{bmatrix}^T$.
The reason for this separation is that it simplifies the axis normalization process during rotations and transformations.
%All individual axes will have a magnitude of 1.

It is important to note that the $n_s$ is defined as the number of individually measured sensor axes and not the number of physical sensors. 
For example, a robot with two physical 3-axis sensors would have $n_s = 6$, where sensor frames $\mathcal{F}_i \in i = \{1,2,3\}$ and $\mathcal{F}_i \in i = \{4,5,6\}$ are located at their respective physical sensors.
Defining $n_s$ in this manner simplifies the derivations that follow.

The prime symbol in $\bm{\hat{s}}^{\prime,i}$ denotes that it is defined in the local $i$-th sensor frame $\mathcal{F}_i$.
A \emph{rotated} sensor axis vector $\bm{\hat{s}}^i$ \emph{without} the prime symbol represents the set of axis vectors rotated to the task frame $\mathcal{F}_{EE}$.
For demonstration purposes, we set $\mathcal{F}_{EE}$ at the end effector, but aligned with the world frame.
All local sensor axis vectors are rotated to match the orientation of the task frame $\mathcal{F}_{EE}$, i.e. $\bm{\hat{s}}^i = \bm{R}\bm{\hat{s}}^{\prime,i}$.
%The importance of defining $\bm{\hat{s}^{\prime}}$ as a set of basis vectors is such that they are not interpreted as a single directional vector when rotated or transformed, but rather an entire frame that is being acted on.

%%%%%%%%%%%%%%%%%%%%%%%%%%%%%%%%%%%%%%%%%%%%%%%%%%%%%%%%%%%%%
\subsection{Sensor Transformation $T_\square(\bm{\hat{s}}^i, \bm{r}^i)$}
\label{sec:sensortranform}

We define the \emph{sensor transformation function} $T_\square(\bm{\hat{s}}^i, \bm{r}^i)$ as a sensor type and physics-dependent transformation that maps individual sensors from their local sensor axes to the task-space. 
For example, when discussing wrenches and force sensing, torque-sensing axes may also observe linear forces at $\mathcal{F}_{EE}$ if there exists a moment arm, analogous to $\bm{f} = \bm{\tau} \times \bm{r}$. 
Thus, a single generalized force-torque sensor $\bm{\hat{s}}^i$ would undergo the following \emph{force} sensor transformation $T_f(\bm{\hat{s}}^i, \bm{r}^i)$, designated by the subscript $f$, in the task frame:

\begin{equation}
\begin{aligned}
%\bm{\hat{s}} = \begin{bmatrix} \bm{\hat{s}}_{p} \\ \bm{\hat{s}}_{\theta} \end{bmatrix} \rightarrow 
%\bm{\tilde{s}}^i = \begin{bmatrix} \bm{\tilde{s}}_{p,i} \\ \bm{\tilde{s}}_{\theta,i} \end{bmatrix} = T_f(\bm{\hat{s}}^i) = \begin{bmatrix} \lvert\bm{\hat{s}}_{p,i}\rvert + \lvert\bm{r}_i \times \bm{\hat{s}}_{\theta,i}\rvert \\ \lvert\bm{\hat{s}}_{\theta,i}\rvert \end{bmatrix}
%\bm{\tilde{s}}_i = \begin{bmatrix} \bm{\tilde{s}}_{p,i} \\ \bm{\tilde{s}}_{\theta,i} \end{bmatrix} = T_f(\bm{\hat{s}}_i) = \begin{bmatrix} \lvert\bm{\hat{s}}_{p,i}\rvert + \sum{\lvert\bm{r}_i \times \{\bm{\hat{s}}_{\theta,i}\}\rvert} \\ \lvert\bm{\hat{s}}_{\theta,i}\rvert \end{bmatrix}
%\bm{\tilde{s}}^i = \begin{bmatrix} \bm{\tilde{s}}^i_{p} \\ \bm{\tilde{s}}^i_{\theta} \end{bmatrix} = T_{f}(\bm{\hat{s}}^i) = \begin{bmatrix} \lvert\bm{\hat{s}}^i_{p}\rvert + \frac{\lvert\bm{r}^i \times \bm{\hat{s}}^i_{\theta}\rvert}{\|\bm{r}^i \times \bm{\hat{s}}^i_{\theta}\|}  \\ \lvert\bm{\hat{s}}^i_{\theta}\rvert \end{bmatrix}
\bm{\tilde{s}}^i &= \begin{bmatrix} \bm{\tilde{s}}^i_{p} \\ \bm{\tilde{s}}^i_{\theta} \end{bmatrix} \\
&= T_{f}(\bm{\hat{s}}^i, \bm{r}^i) = 
\begin{cases}
       \begin{bmatrix} \lvert\bm{\hat{s}}^i_{p}\rvert + \bm{0} \\ \lvert\bm{\hat{s}}^i_{\theta}\rvert \end{bmatrix} \text{,} &\text{if }  \bm{\hat{s}}^i_{\theta} \times \bm{r}^i = \bm{0} \\ %      \bm{r}^i \times \bm{\hat{s}}^i_{\theta}
       \begin{bmatrix} \lvert\bm{\hat{s}}^i_{p}\rvert + \frac{\lvert\bm{\hat{s}}^i_{\theta} \times \bm{r}^i\rvert}{\|\bm{\hat{s}}^i_{\theta} \times \bm{r}^i\|}  \\ \lvert\bm{\hat{s}}^i_{\theta}\rvert \end{bmatrix} \text{,} &\text{otherwise.} \\ 
     \end{cases}
\label{eq:senstransf}
\end{aligned}
\end{equation}

%\begin{figure}[t]
%\centering
%\includegraphics[width = 0.38\textwidth]{figures/rposdefn}
%\caption{\hl{rposdefn} defined in introellipsoid}
%\label{fig:rposdefn}
%\end{figure}

\noindent where $\bm{r}^i$ is the position vector from the $i$-th sensor axis to the task frame $\mathcal{F}_{EE}$, $\lvert \cdot \rvert$ is the element-wise absolute function\footnote{The derivative of the absolute function is not defined at 0, which affects the derivative terms. Thus, practically, one should use an alternate representation to the absolute function that is smooth around 0, e.g. $\lvert x \rvert \approx x \text{tanh}(cx)$ where $c$ is a positive constant.}, and $\|\cdot\|$ is the Euclidean norm to normalize the cross product as directional analysis of sensor axes should not be influenced by the magnitude of the moment arm.
The piece-wise defined function is used in the case where $\bm{r}^i$ and $\bm{\hat{s}}^i_{\theta}$ are collinear such that $\|\bm{\hat{s}}^i_{\theta} \times \bm{r}^i\| = 0$, which would otherwise result in an undefined fraction.
Note that the hat operator $\bm{\hat{\cdot}}$ designates a locally-defined sensor axis, whereas the tilde operator $\bm{\tilde{\cdot}}$ designates the \emph{transformed} sensor axis.
%Note that the second term of $\bm{\tilde{s}}_{p,i}$ is the summation of cross products between $\bm{r}_i$ and each basis vector of the set $\{\bm{\hat{s}}_{\theta,i}\}$, rather than with the vector $\bm{\hat{s}}_{\theta,i}$ directly. \hlc{Why it can't be directly $\bm{r}_i \times \bm{\hat{s}}_{\theta,i}$?} \hl{Because } 

The method to interpret the transformed quantity $\bm{\tilde{s}}^i$ is as follows: each element of $\bm{\tilde{s}}^i$ represents a task-space axis that is observed by the various locally-defined terms of $\bm{\hat{s}}^i$ that it contains.
For example, in (\ref{eq:senstransf}), given that both $\bm{\hat{s}}^i_{p}$ and $\bm{\hat{s}}^i_{\theta}$ terms appear in the translational force term $\bm{\tilde{s}}^i_{p}$, any translational forces at the EE would be observed by both the linear and rotational axes of the $i$-th sensor (if they exist).

The use of the absolute function is two-fold: a) we assume that the sensors are bidirectional and b) it ensures that sensor axes do not subtract from each other.
It is important to note that the exact transformation $T_\square(\cdot)$ is dependent on the sensor type and the laws of physics that govern it. 
Certain transformation functions, depending on the sensor type, may simply be the identity function. 
Other types of systems and transformations will be explored in future work.
%As a note, it is possible to use non-Cartesian task spaces where orthogonality of task-space axes is something other than $90^\circ$ when applicable. 
%For example, a single-axis distance sensor is defined to be only able to observe in $10^\circ$ arcs. 
%For a 2D planar space, there could exist $n_t = \frac{360^\circ}{10^\circ} = 36$ task-space axes that are all orthogonal to each other. 
%\hlc{I'm not really sure that I understand the orthogonality here}. 
%Another example could be cameras distributed over a robot arm that each has a field of view of $70^\circ$.
%A planar space could then be defined task space axes set $45^\circ$ apart such that $T_{cam}(\bm{\hat{s}}^i, \bm{r}^i)$ would make a single transformed camera axis observe adjacent task space axes as well. 
%with $n_t = \frac{360^\circ}{60^\circ} = 10$ axes where

%\hlc{Separate $n_t$ for local sensor axes and $n_t$ for actual task space? This would be very confusing for force/torque definitions though}

%In relation to (\ref{eq:senstransf}), another potential analysis is to normalize all quantities, represented by $\|\cdot\|$, particularly the cross product term $\bm{r}_i \times \bm{\hat{s}}_{\theta,i}$. Note the non-cross product terms do not require normalization as they are already unit vectors.

%%%%%%%%%%%%%%%%%%%%%%%%%%%%%%%%%%%%%%%%%%%%%%%%%%%%%%%%%%%%%
\subsection{Sensor Observability Matrix $\bm{S}$, System Vector $\bm{s}$, Function $\Gamma_\square$, Index $o$ and Ellipsoid}
%\subsection{Sensor Observability $\bm{s}$, Function $\Gamma_\square(\bm{\tilde{s}})$, Observability Index $o$ and Ellipsoid}
\label{sec:sensobs}

We define the \emph{sensor observability matrix} $\bm{S}(\bm{q}) \in \mathbb{R}^{{n_t} \times {n_s}}$ as the matrix of column vectors of the transformed sensor axis vectors $\bm{\tilde{s}}^i$.
For readability, we will continue the manuscript without explicitly writing the dependency on $\bm{q}$, in other words $\bm{S(q)} \rightarrow \bm{S}$:

\begin{equation}
\bm{S} = \begin{bmatrix} \bm{\tilde{s}}^1 \cdots \bm{\tilde{s}}^{n_s} \end{bmatrix}
\label{eq:sensmatrix}
\end{equation}

Next, we define the overall \emph{system sensor observability vector} $\bm{s} \in \mathbb{R}^{n_t \times 1}$ as the cumulative sensing capabilities of all individual sensors of the system in the task frame $\mathcal{F}_{EE}$ with $n_t$ task axes.
The \emph{sensor observability function} $\Gamma_\square(\bm{S})$ calculates $\bm{s}$ by synthesizing all $n_s$ transformed sensor axes $\bm{\tilde{s}}^i$ according to a desired metric for analysis. 
%, where $\bm{\tilde{s}} = \{\bm{\tilde{s}}^i\}: i = 1...n_s$ is the set of all individual transformed sensors axes. 
Here, we give example definitions of $\Gamma_\square(\bm{S})$. 
Recall that $\bm{s} = \begin{bmatrix} s_1 & \cdots & s_j \end{bmatrix}^T$, where $j \in 1...n_t$.

\subsubsection{Row-wise sum function}
\begin{equation}
\bm{s} = \Gamma_{sum}(\bm{S}) = \sum_{i=1}^{n_s} \bm{\tilde{s}}^i
\label{eq:obsfunc-sum}
\end{equation}

%\subsubsection{Row-wise max function}
%\begin{equation}
%\begin{gathered}
%\bm{s} = \Gamma_{max}\left(\bm{S}\right) \\
%%\bm{s} = \Gamma_{max}\left(\bm{\tilde{s}}\right) \\
%\forall j \in 1...{n_t} : \bm{s}_j = \max_{i=1...n_s} \bm{\tilde{s}}^i_j
%\end{gathered}
%\label{eq:obsfunc-max}
%\end{equation}

\subsubsection{Row-wise $p$-norm function}
\begin{equation}
%\bm{s} = \Gamma_{p}(\bm{S}) = \sqrt[p]{\sum_{i=1}^{n_s} (\bm{\tilde{s}}^i)^p}
\bm{s} = \Gamma_{\|\cdot\|_p}(\bm{S}) \text{, where } s_j = \sqrt[p]{\sum_{i=1}^{n_s} (\tilde{s}^i_j)^p} \quad \forall j \in 1...{n_t}
\label{eq:obsfunc-L2norm}
\end{equation}

\subsubsection{Row-wise max function}
\begin{equation}
%\begin{gathered}
\bm{s} = \Gamma_{max}\left(\bm{S}\right) \text{, where } s_j = \max_{i=1...n_s} \tilde{s}^i_j \quad \forall j \in 1...{n_t} 
%\\
%\text{Recall: } \bm{s} = \begin{bmatrix} s_1 & \cdots & s_j \end{bmatrix}^T
%\bm{s} = \Gamma_{max}\left(\bm{\tilde{s}}\right) \\
%\end{gathered}
\label{eq:obsfunc-max}
\end{equation}

\noindent where the subscript $j$ in $s_j$ and $\tilde{s}_j$ indicates the $j$-th task-space axis of $\bm{s}$ and $\bm{\tilde{s}}$, respectively\footnote{Recall: superscript $i$ is for the $i$-th sensor axis, which is different from the subscript $j$ for the $j$-th task space axis (and similarly for subscript $k$ for the $k$-th joint axis, which will be defined later).}, and also corresponds to the $j$-th row of $\bm{S}$.
%the inclusion $j$ is to perform these operations separately on each task-space axis.
%These two function definitions produce different results, which are of utmost importance \hl{during optimization}. 
The sum function $\Gamma_{sum}(\cdot)$, as the name implies, performs a row-wise summation across all transformed sensor axes in $\bm{S}$ and measures the cumulative task-space sensing capabilities across all sensors.
The summation can potentially provide a measure of redundancy if multiple sensors measure the same task space axis.
One potential issue with this method is that, for the same value, the sum function does not differentiate between an axis that is directly observed by one or a few sensors, or only minimally observed by many off-axis sensors.
The lack of this differentiation may result in unintended low quality readings from non-closely aligned sensors. 
The $p$-norm function $\Gamma_{\|\cdot\|_p}(\cdot)$ partially alleviates this issue by reducing the impact of smaller values.

Conversely, the element-wise max function $\Gamma_{max}(\cdot)$ determines the maximum alignment between the individual sensor axes and each task space axis.
It provides a quality measure of how \emph{directly} a task-space axis is observed and, in a sense, its trustworthiness.
The max function $\Gamma_{max}(\cdot)$ is always bounded between $[0,1]$, where a value of $s_j = 1$ indicates that there is at least one sensor that is directly and fully observing the $j$-th task space axis, while $s_j < 1$ indicates that it is only measured indirectly by all sensors.
In all cases, $s_j \approx 0$ would indicate that the $j$-th axis is in danger of no longer being observed. 

Other sensor observability functions may be used as well, depending on the preferred analysis.
For example, a sum with minimum thresholding could potentially negate the masking effect if many low quality observations by minimally observed sensor axes are present.
Note that certain formulations of $\Gamma_\square(\cdot)$ may also use the sensor positions (in addition to the sensor orientations in $\bm{S}$) in case it is relevant, e.g. for modelling sensor-to-sensor interactions.
An example is discussed in Sec. \ref{sec:SOI-Advantages}.

Next, we define the \emph{sensor observability index} $o$:
%\hl{product or condition number-like ratio of s\_max/s\_min?}

\begin{equation}
o = \prod_{j = 1}^{n_t} s_j
\label{eq:obsindex}
\end{equation}

Analogous to the kinematic manipulability index $w_k$ in (\ref{eq:manip_kin}), the sensor observability index $o$ is a scalar quality measure of task-space observability.
If any $s_j \rightarrow 0$, then $o \rightarrow 0$, and the system is at risk of being unable to observe one or more task space axes. 
The case where $o = 0$ is called a \emph{sensor observability singularity} where the robot is in a \emph{sensor observability singular configuration}, and the system has lost the ability to observe one or more task space axes.
This situation should be avoided for risk of potentially causing failure resulting from the robot being blind in certain task space axes.
As such, $o$ can be used as an optimization variable during motion planning to avoid low quality joint configurations (examples shown in Sec. \ref{sec:SOoptimization}).
While numerical interpretation of the sensor observability index is system-dependent, it can easily used as a relative gauge of system sensor observability performance, as discussed in \cite{Patel2015JIRS-SurveyManipulatorPerfMeasures} for $w_k$.
%As the sensor observability index is more of a qualitative term, its exact numerical values are somewhat meaningless. 

\begin{figure}[t]
\centering
\subfigure[]{
	\includegraphics[width=0.43\textwidth]{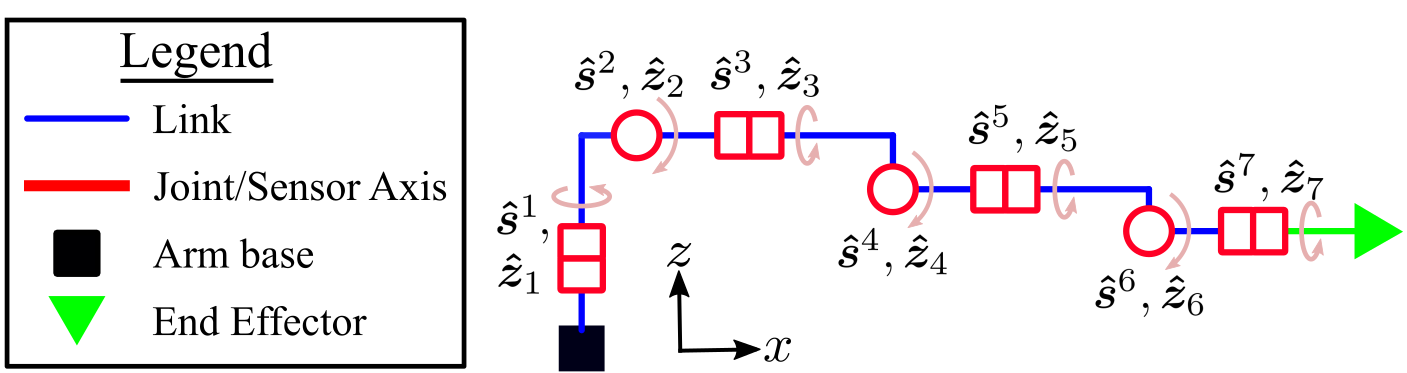}
	\label{fig:robotconfig-structure}}
\subfigure[]{
%\subfigure[$\bm{\theta} = \begin{bmatrix}\frac{\pi}{6} \ \frac{\pi}{6} \ \frac{\pi}{6} \ \frac{\pi}{6} \ \frac{\pi}{6} \ \frac{\pi}{6} \ \frac{\pi}{6} \end{bmatrix}^T$]{
	\includegraphics[height = 4.75 cm]{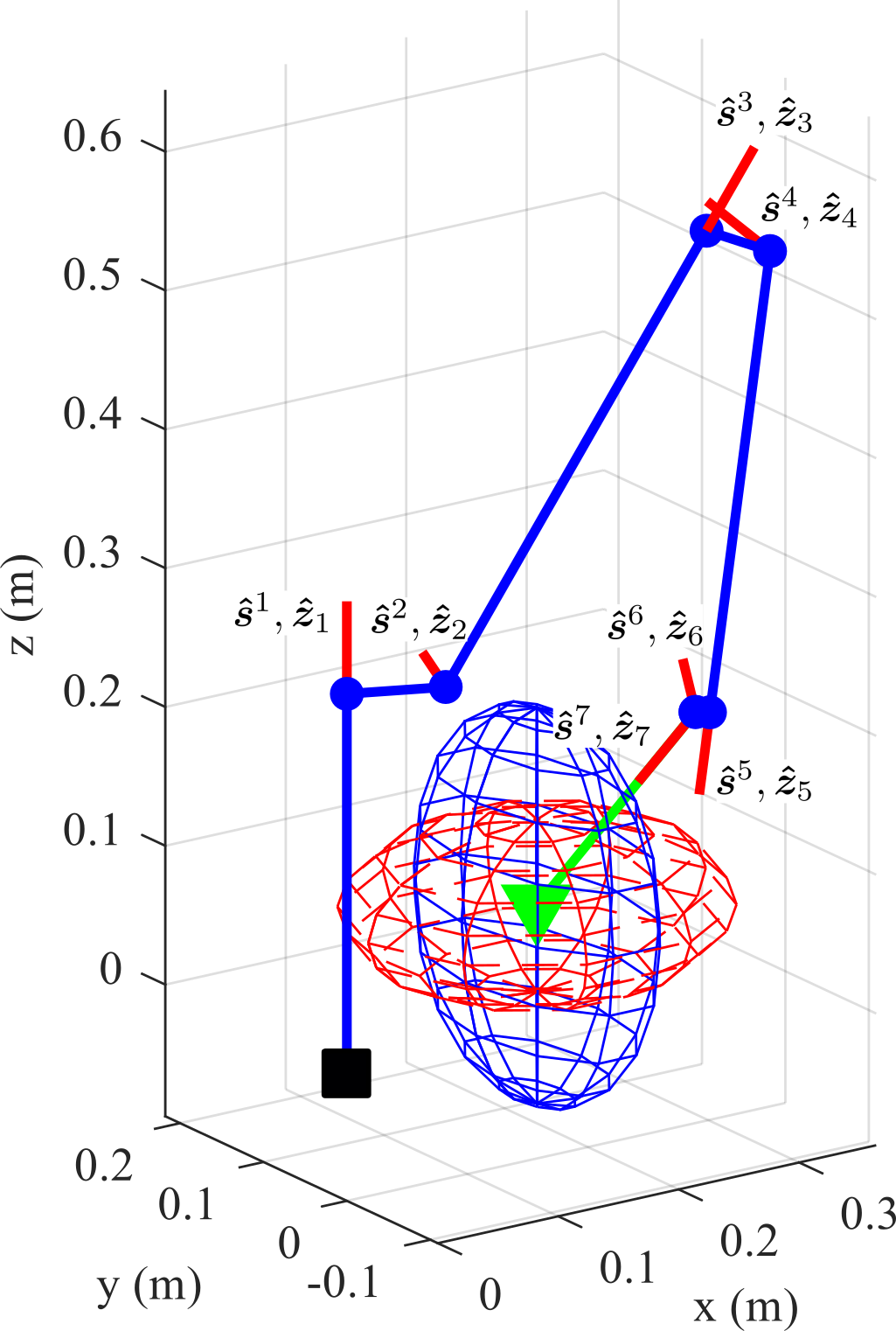}
	\label{fig:robotconfig-normal}}
\subfigure[]{
%\subfigure[$\bm{\theta} = \begin{bmatrix}\frac{\pi}{4} \ \minus\frac{\pi}{2} \ 0 \ \pi \ 0 \ 0 \ 0 \end{bmatrix}^T$]{
	\includegraphics[height = 4.75 cm]{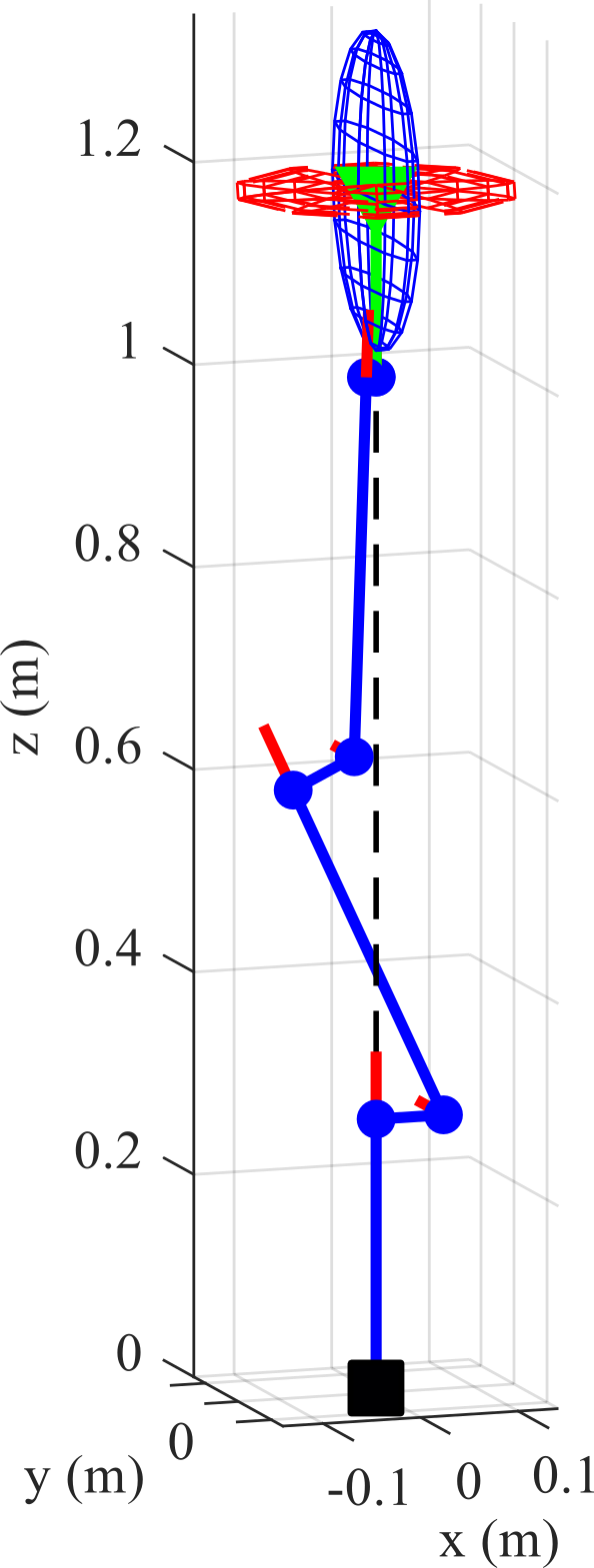}
	\label{fig:robotconfig-kinematicsingularity}}
\subfigure[]{
%\subfigure[$\bm{\theta} = \begin{bmatrix}\frac{\pi}{4} \ \minus\frac{\pi}{2} \ 0 \ \pi \ 0 \ 0 \ 0 \end{bmatrix}^T$]{
	\includegraphics[height = 4.75 cm]{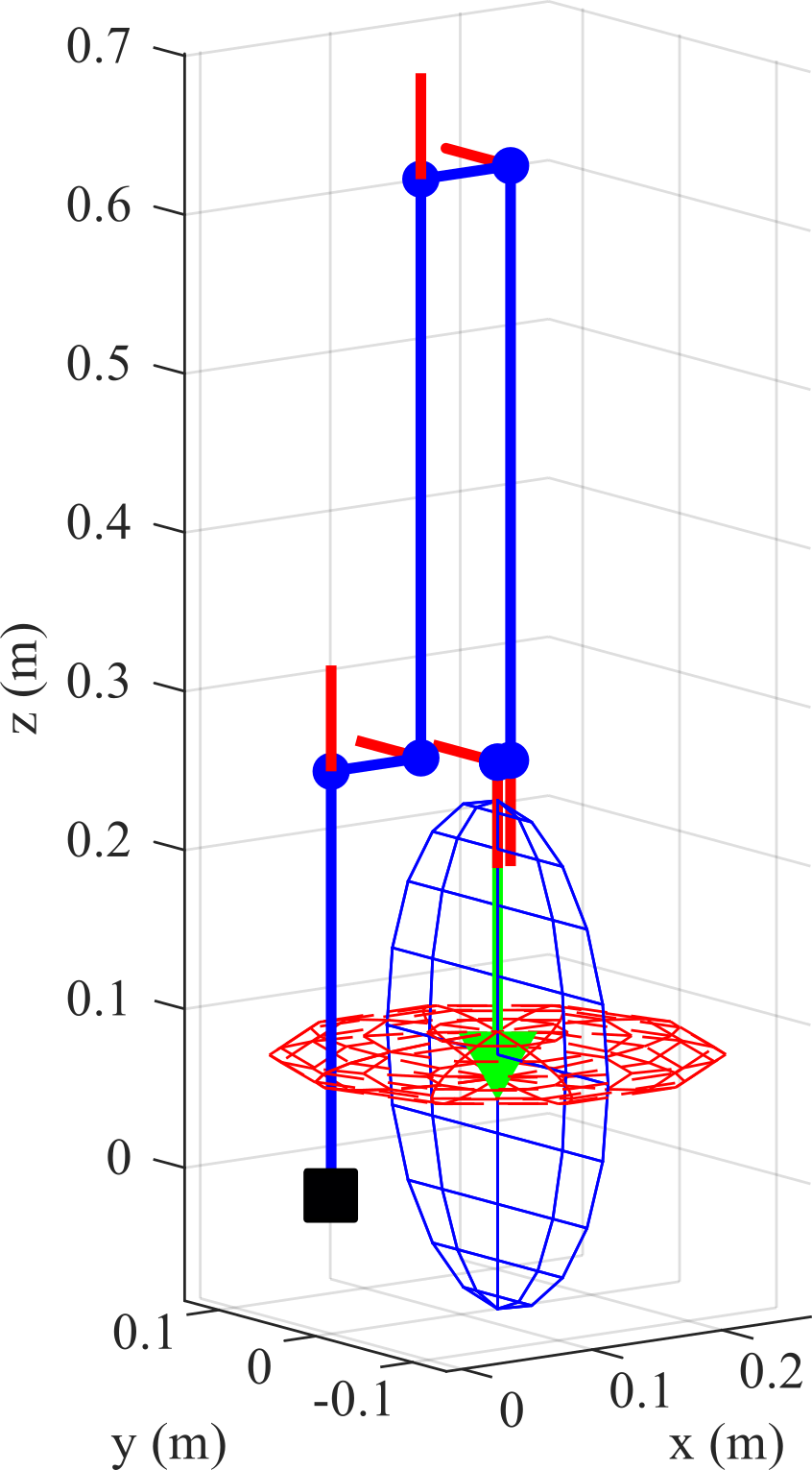}
	\label{fig:robotconfig-singularity}}
\subfigure[]{
	\includegraphics[width=0.37\textwidth]{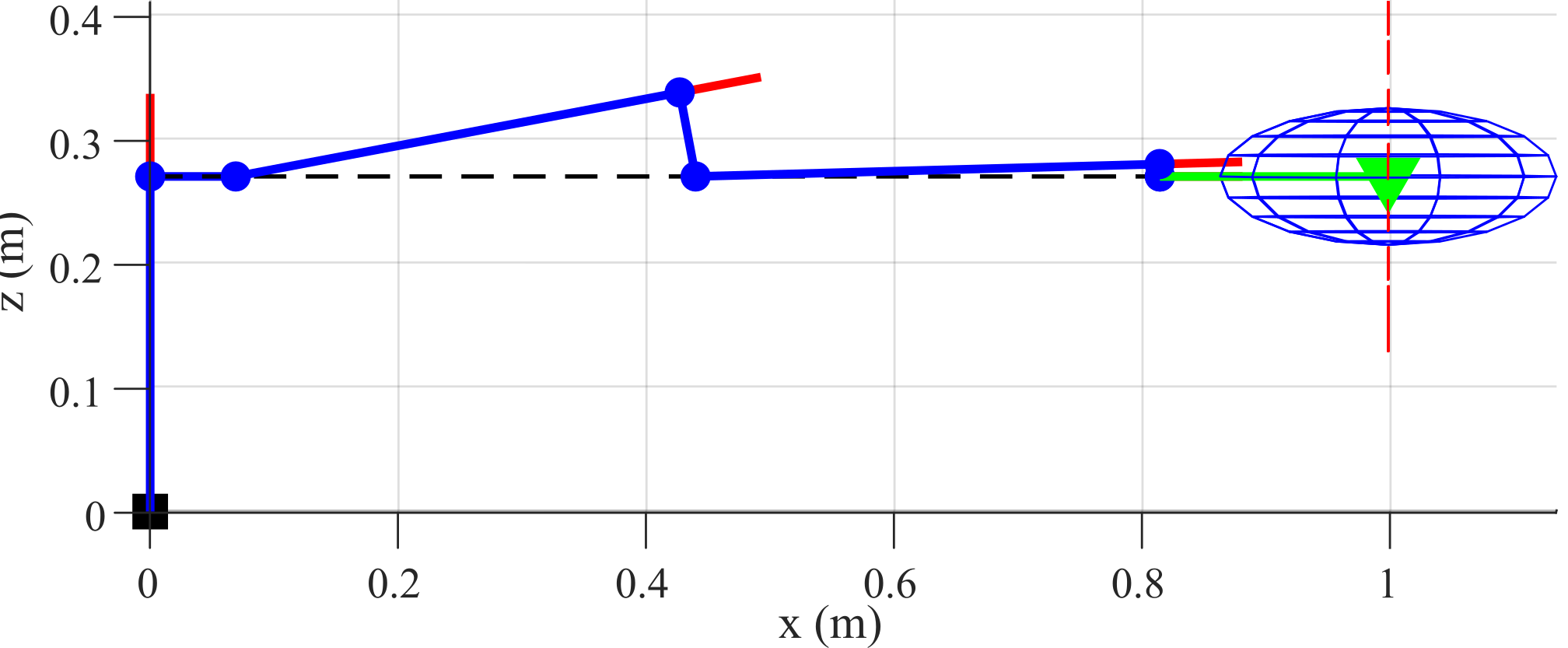}
	\label{fig:robotconfig-singsensX}}
\caption{a) Representation of a single 7-DOF Baxter robot arm with only traditional joint torque sensors in b) arbitrary configuration, c) kinematic singularity in $\theta_z$, d-e) sensor observability singularities in d) $\tau_x$ and e) $f_x$. 
%Blue lines represent the robot links, red lines are joint axes, and the green triangle is the robot end effector. Black square indicates arm base. 
Force and torque observability ellipsoids based on the sum function $\Gamma_{sum}(\cdot)$ are shown in red dashed and blue solid ellipsoids, respectively. 
Note that the ellipsoids in c) are very thin, but not completely flat, i.e. $o \neq 0$.}
\label{fig:robotconfig}
\end{figure}

Similar to the manipulability ellipsoid defined previously in Sec. \ref{sec:background}, we define the sensor observability ellipsoid in $\mathbb{R}^{n_t}$ where the principal axes are proportional to the magnitude of the task-space observability.
Fig. \ref{fig:robotconfig} showcases various joint configurations for the robot Baxter and their resulting sensor observability ellipsoid.
For visualization purposes, sensor observability is split into the force $\bm{s}_{p}$ (red dashed line ellipsoid) and torque $\bm{s}_{\theta}$ (blue solid line ellipsoid) components.
In the arbitrary configuration in Fig. \ref{fig:robotconfig-normal}, all axes are observable, as can be seen by the 3D shape of the force and torque ellipsoids. 
In the sensor observability singular configurations shown in Figs. \ref{fig:robotconfig-singularity} and \ref{fig:robotconfig-singsensX}, an axis of the sensor observability ellipsoid collapses to zero. 
This indicates that the corresponding axis, torques along the $x$-axis in Fig. \ref{fig:robotconfig-singularity} and forces along the $x$-axis in Fig. \ref{fig:robotconfig-singsensX}, is not observable by the joint torque sensors. 

\subsection{Effect of Joint Configuration on Sensor Observability}
\label{sec:simulationBaxter}

\begin{figure}[t]
\centering
\subfigure[]{
	\includegraphics[width = 0.43\textwidth]{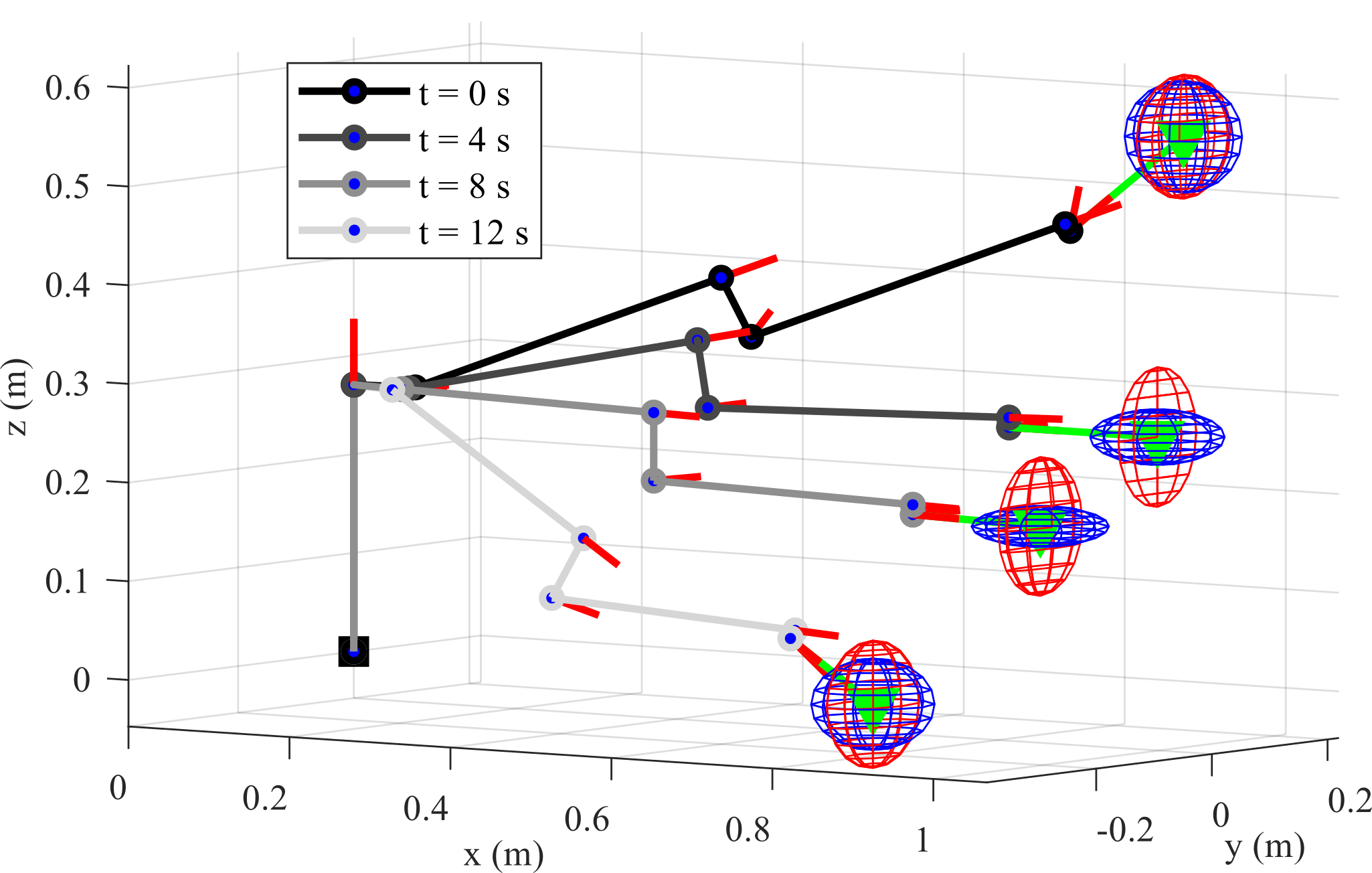}
	\label{fig:simsweep-robotconfig}}
\subfigure[]{
	\includegraphics[width = 0.44\textwidth]{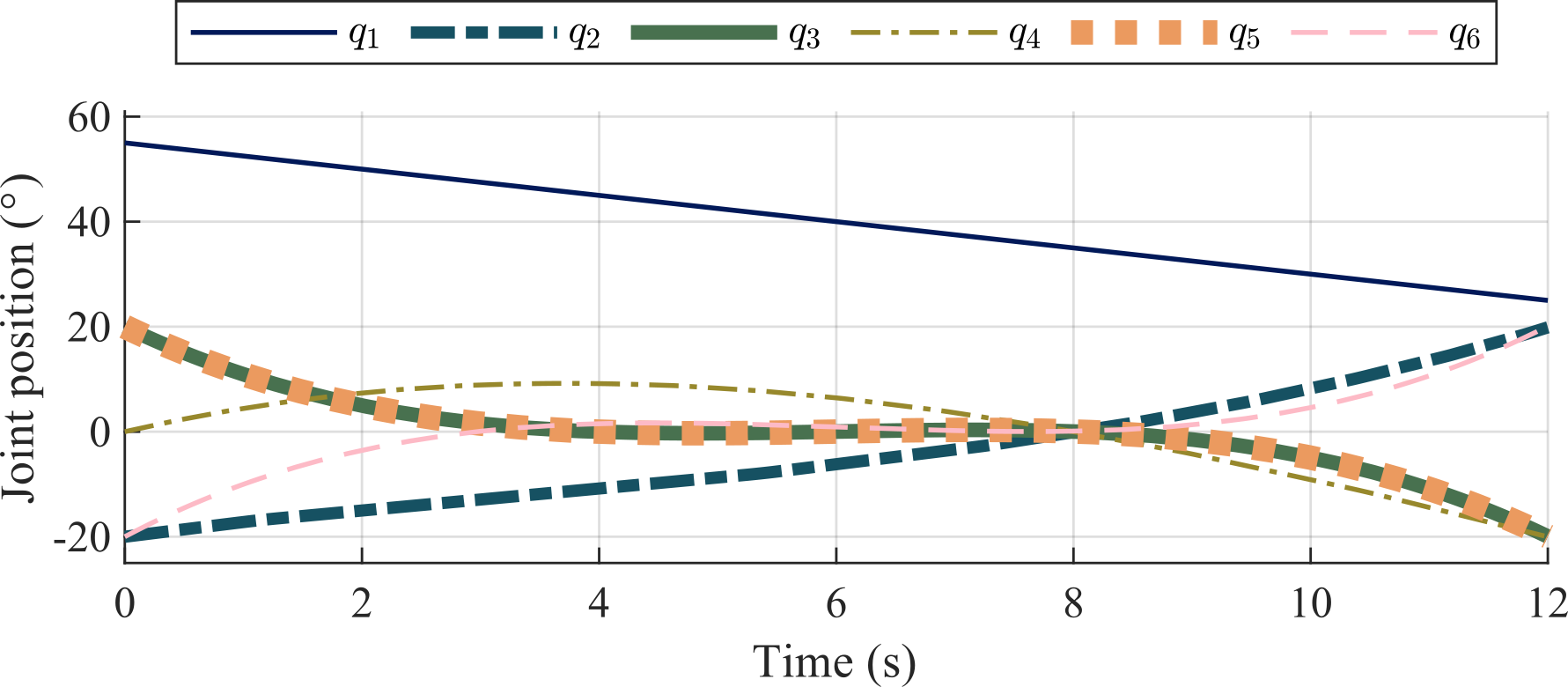}
	\label{fig:simsweep-jointpos}}
\subfigure[]{
	\includegraphics[width = 0.44\textwidth]{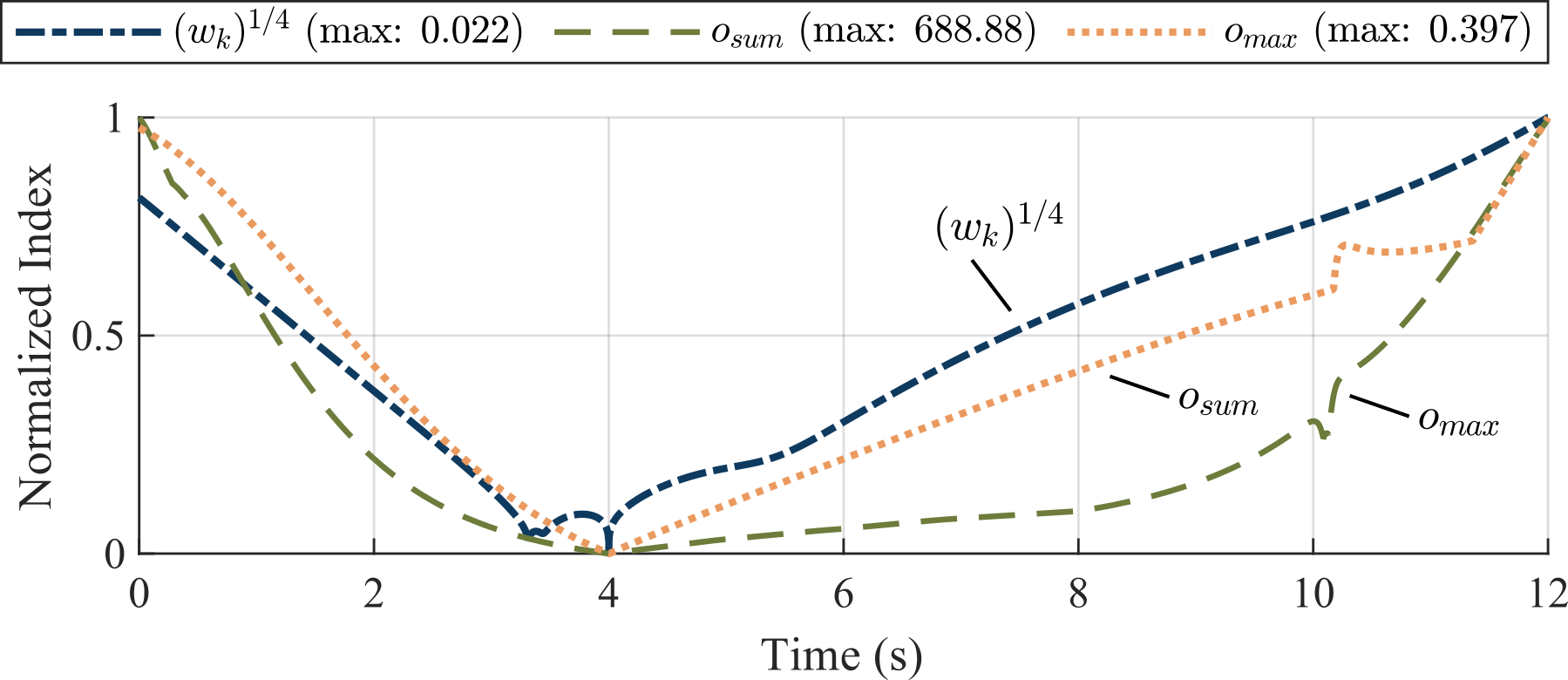}
	\label{fig:simsweep-index}}
\caption{a) Simulation of a single Baxter robot arm starting in an arbitrary position, sweeping through an observability singularity at $t = 4$ s (configuration shown in Fig. \ref{fig:robotconfig-singsensX}),
% a sensor manipulability singularity at $t = 8$ s, 
setting joints $\bm{q}_{2}$ to $\bm{q}_{6}$ to 0 at $t = 8$ s, 
and ending in an arbitrary position at $t = 12$ s. b) Joint positions of the maneuver. 
Joint $q_7$ is not shown as it is held at a constant $q_7 = 0$. 
c) Plot of the evolution of kinematic manipulability $w_k$ and 
%sensor manipulability $w_s$ (using the sum function $\Gamma_{sum}(\bm{S})$), 
observability index using the sum $o_{sum}$ and max $o_{max}$ functions. All indices are normalized to 1, but $w_k$ 
%and $w_s$ are
is further scaled with an exponential to emphasize the changes at $t = 4$ s.
%and $t = 8$ s
}
%\caption{a) Simulation of a single Baxter robot arm moving starting in an arbitrary position, sweeping through an observability singularity at $t = 4$ s (configuration shown in Fig. \ref{fig:robotconfig-singsensX}),
 %a sensor manipulability singularity at $t = 8$ s, 
%and ending in an arbitrary position at $t = 12$ s. b) Joint positions of the maneuver. 
%Joint $q_7$ is not shown as it is held at a constant $q_7 = 0$. 
%c) Evolution of the following indices: kinematic manipulability $w_k$, 
%sensor manipulability $w_s$ (using the sum function $\Gamma_{sum}(\bm{S})$), 
%observability index using the sum $o_{sum}$ and max $o_{max}$ functions. All indices are normalized, but $w_k$ and $w_s$ are further scaled with an exponential to emphasize the changes close to zero at $t = 4$ s and $t = 8$ s. \hlc{remove sensor manip -- THESE FIGURES HAVE BEEN UPDATED TO SMALLER ONES -- CHECK IF NEWER FIGURES ARE USED IN CONF}}
\label{fig:simsweep}
\end{figure}

We use the fixed-base dual-arm robot Baxter from Rethink Robotics to demonstrate the importance of sensor observability analysis. 
Each Baxter arm contains 7 degrees of freedom whose kinematic structure is shown in Fig. \ref{fig:robotconfig-structure} and described in detail in \cite{Williams2017-BaxterKinematics}. 
Each joint contains position encoders and the joints are capable of torque estimation.
Given the structure of Baxter, there exists sensor observability singular configurations, as shown in Fig. \ref{fig:robotconfig}.
%\hl{The Denavit-Hartenburg parameters are listed in Table REF}
% https://www.researchgate.net/publication/299640286_Baxter_Kinematic_Modeling_Validation_and_Reconfigurable_Representation
% https://www.ohio.edu/mechanical-faculty/williams/html/PDF/BaxterKinematics.pdf

To demonstrate the evolution of the various indices, we simulate the kinematic structure of a single Baxter arm in MATLAB and sweep through multiple configurations shown in Fig. \ref{fig:simsweep-robotconfig}.
The robot begins in an arbitrary configuration at $t = 0$ s.
At $t = 4$ s, the robot moves to the configuration shown in Fig. \ref{fig:robotconfig-singsensX}, which incurs simultaneous sensor observability and kinematic manipulability singularities $o,w_k=0$.
%, sensor manipulability $w_s,$
%At $t = 8$ s, the robot is in a configuration where it is only a \emph{sensor} manipulability singularity $w_s = 0$.
At $t = 8$ s, joints $q_2$ to $q_6$ are set to zero. %, $q_2 = ... = q_6 = 0$.
The robot finally ends in another arbitrary configuration at $t = 12$ s.
The joint angles are plotted in Fig. \ref{fig:simsweep-jointpos}.
Joint $q_7$ is not shown in the plot as it is held at a constant $q_7 = 0$ and has no effect on the results.
Fig \ref{fig:simsweep-index} plots the evolution of the kinematic manipulability index $w_k$ and sensor observability index using both the sum $o_{sum}$ in (\ref{eq:obsfunc-sum}) and max $o_{max}$ functions in (\ref{eq:obsfunc-max}) through the different robot configurations.
%, sensor manipulability index $w_s$ (using the sum function $\Gamma_{sum}(\bm{S})$) 
All indices are normalized to their respective maxima seen throughout the motion, though $w_k$ is further scaled to emphasize its evolution particularly at $t = 4$ s.
%and $w_s$ 

As expected, all indices are non-zero in the arbitrary configurations at $t = 0$ s and $t = 12$ s. 
As the robot moves towards the observability and kinematic singularity at $t = 4$ s, all indices approach zero.
This singular configuration eliminates the observability of force ($o_{sum}, o_{max} \rightarrow 0$) and translational motion ($w_k \rightarrow 0$) in the $x$-axis.
 %and the robot's ability to increase or decrease observability ($w_s \rightarrow 0$) 
%At $t = 8$ s, the robot enters another position where sensor manipulability singularity $w_s = 0$, but since observability is non-zero, this configuration is not necessarily critical.
Comparing the sensor observability indices $o_{sum}$ and $o_{max}$, both calculation methods hold somewhat similar trends. 
While $o_{max}$ has a maximum value of 1, $o_{sum}$ is theoretically unbounded but is normalized for the plot according to the maximum of 688.88 observed in this simulated motion.

%What is interesting is that the observability $o$ and the kinematic manipulability $w_k$ indices have similar trends but do not coincide exactly, despite the relationship shown in Sec. \ref{sec:parallels}. 

%\hl{While $w_s$ seems to suddenly reduce to 0 in many configurations}

%\hl{Practically, the central difference method is used to calculate the sensor Jacobian as it is on the order of 40000 times faster computationally compared to symbolically calculating the derivative on a computer equipped with Intel i5-4670k processor using the MATLAB symbolic toolbox.}
%For the path defined in Sec \ref{sec:simulation}, the mean error between the analytical and numerical solutions is 0.025\% (ignoring when $w_s = 0$).

%%%%%%%%%%%%%%%%%%%%%%%%%%%%%%%%%%%%%%%%%%%%%%%%%%%%%%%%%%%%%
%\subsection{Physical Interaction with Baxter Robot} % Example Use Case: Kinesthetic Teaching
%\subsection{\hl{Motivation for Sensor Observability Analysis}}
%\label{sec:exp-kinesteach}

\begin{figure}[t]
\centering
%\subfigure[]{
	%\includegraphics[width = 0.43\textwidth]{figures/expFS-norm}
	%\label{fig:expFS-norm}}
%\subfigure[]{
	%\includegraphics[width = 0.47\textwidth]{figures/exp-plot-norm}
	%\label{fig:expFS-norm-plot}}
%\subfigure[]{
	%\includegraphics[width = 0.43\textwidth]{figures/expFS-sing}
	%\label{fig:expFS-sing}}
%\subfigure[]{
	%\includegraphics[width = 0.47\textwidth]{figures/exp-plot-sing}
	%\label{fig:expFS-sing-plot}}
\subfigure[]{
	\includegraphics[width = 0.225\textwidth]{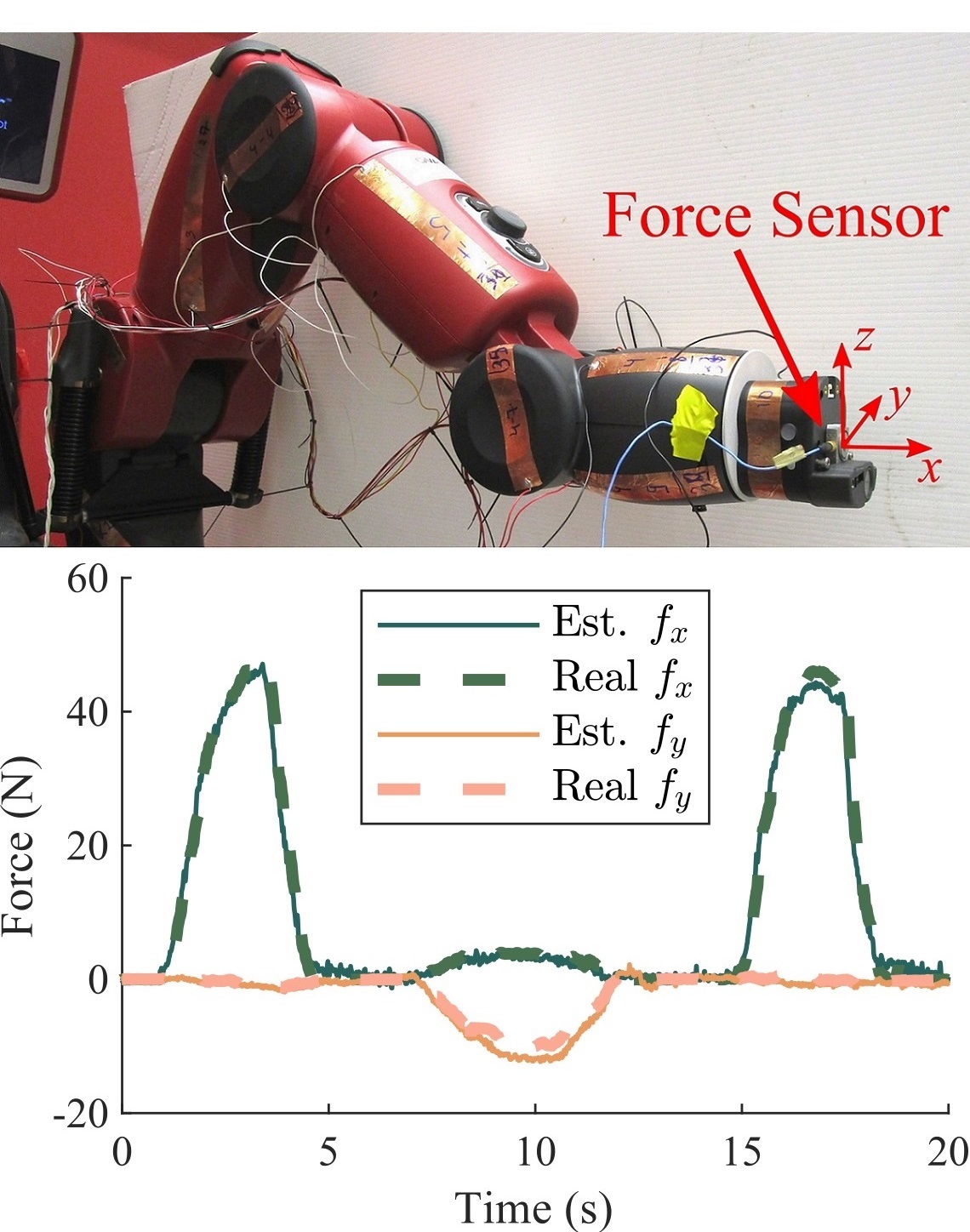}
	\label{fig:expFS-norm}}
\subfigure[]{
	\includegraphics[width = 0.225\textwidth]{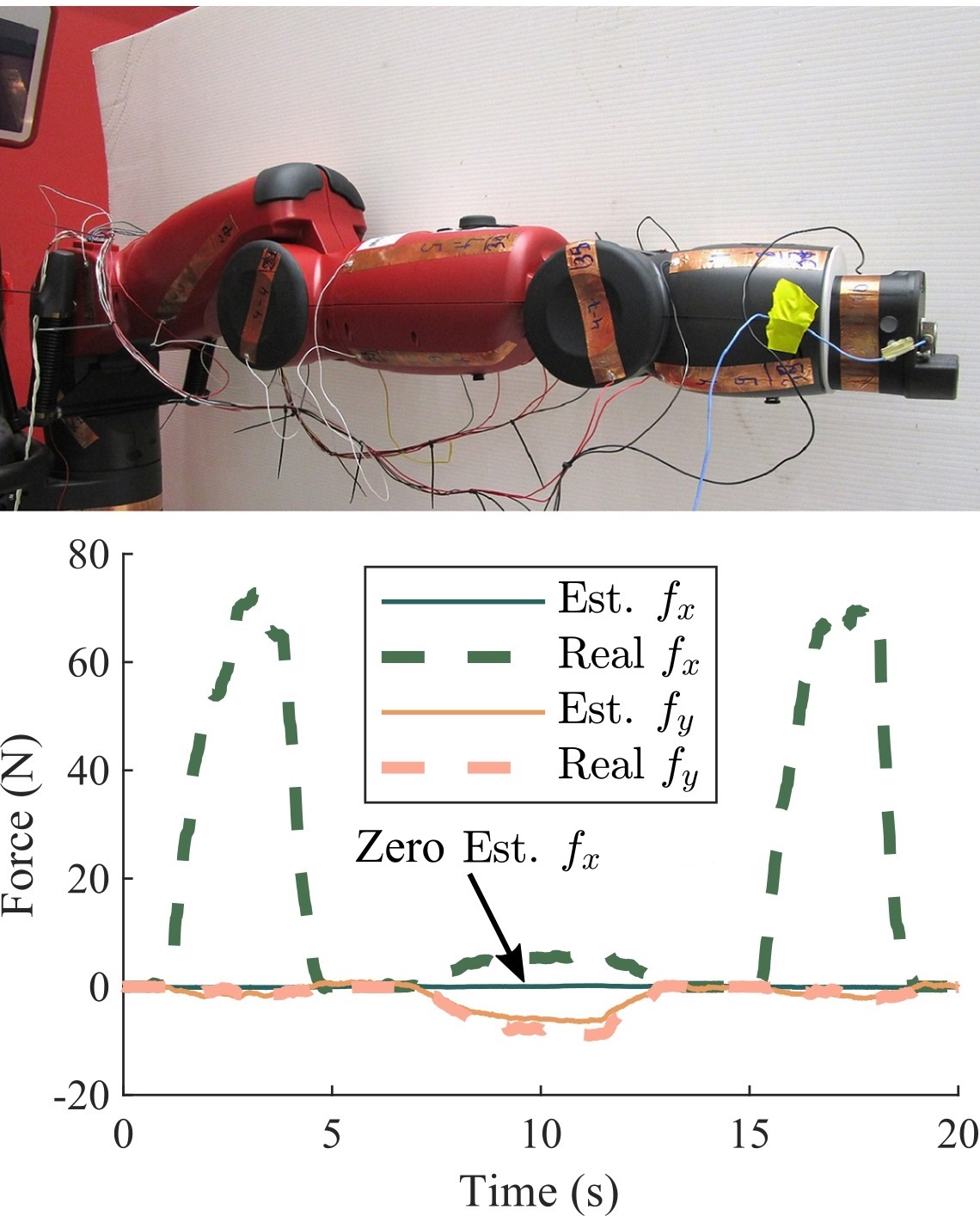}
	\label{fig:expFS-sing}}
\caption{
%Sensor observability experiments using the robot Baxter comparing end effector forces estimated by the joint torque sensors and measured using an external force sensor attached to the end effector in a-b) an arbitrary non-zero observability configuration and c-d) the observability singular configuration shown in Fig. \ref{fig:robotconfig-singsensX}. Plots b) and d) show the readings as a result of an external force applied first in $x$, then $y$, and finally $x$ again for the respective joint configurations. \hlc{FIGURES WERE UPDATED IN CONF PAPER}
Sensor observability experiments using the robot Baxter comparing end effector forces estimated by the joint torque sensors and measured using an external force sensor attached to the end effector in a) an arbitrary non-zero observability configuration and b) the observability singular configuration shown in Fig. \ref{fig:robotconfig-singsensX}. An external force is applied first in $x$, then $y$, and finally $x$ again.
Forces are fully observable in the arbitrary position in a), but forces in $x$ are not observable in the observability singular configuration in b).
}
\label{fig:expFS}
\end{figure}

%\hl{combine this section and last section, saying how this section is analysis of the singularity position}
%\hl{combine the two optimizations in a new section after singularity analysis}
%Using the robot Baxter, we illustrate the importance of sensor observability index by observing changes in the robot's ability to use its joint sensors to estimate end effector forces in both normal and observability singular configurations.
To practically illustrate the importance of sensor observability index, we observe changes in the ability of the Baxter robot to use its joint sensors to estimate end effector forces in both normal and observability singular configurations in a physical interaction experiment shown in Fig. \ref{fig:expFS}.
End effector force estimation from joint sensors is performed using the packaged Baxter API from the manufacturer and a force sensor is attached onto the end effector of the robot as shown in Fig. \ref{fig:expFS-norm} to provide a ground truth for interaction forces.
Once the robot is in position, the end effector is first pushed along the $x$-, then the $y$-, and finally $x$-axes again to observe whether the interaction forces are detected or not.

In the first scenario, shown in Fig. \ref{fig:expFS-norm}, the robot is in an arbitrary non-zero observability configuration.
The associated plot shows the robot's ability to resolve the end effector forces using the joint torque sensors.
Conversely, in the second scenario, shown in Fig. \ref{fig:expFS-sing}, the robot is in the sensor observability singular configuration shown in Fig. \ref{fig:robotconfig-singsensX}.
In this configuration, the overall system sensor observability $s_x = 0$ but $s_y \neq 0$.
In the force plot in Fig. \ref{fig:expFS-sing}, the robot is unable to observe interaction forces in the $x$-axis at $t \approx 3$ s and $t \approx 17$ s, despite the ground truth force sensor showing interaction forces. 
Forces in the $y$-axis are observed without issue. 
Off-axis forces are observed from imperfect interactions and the robot shifting during interaction.

%%%%%%%%%%%%%%%%%%%%%%%%%%%%%%%%%%%%%%%%%%%%%%%%%%%%%%%%%%%%%
%%%%%%%%%%%%%%%%%%%%%%%%%%%%%%%%%%%%%%%%%%%%%%%%%%%%%%%%%%%%%
\section{Discussions on Sensor Observability Analysis}
\label{sec:specialstuff}
%%%%%%%%%%%%%%%%%%%%%%%%%%%%%%%%%%%%%%%%%%%%%%%%%%%%%%%%%%%%%
\subsection{Special Case: Similarities to Kinematic Analysis} %Special Case - Serial Manipulator with Joint Torque Sensors
\label{sec:parallels}

In the special case where the sensor axes are collinear with the joint axes, parallels can be drawn between the sensor observability and kinematic analyses.
Let us examine a serial manipulator with only revolute joints and single-axis joint torque sensors located at each joint that are aligned with the joint axes.
This is the joint-sensor configuration of a typical serial manipulator robot.
In this specific case, each local sensor axis vector $\bm{\hat{s}}^{\prime,i} = \bm{\hat{s}}^{\prime}_{\tau z} \ \forall\ i = 1...n_s$ as in (\ref{eq:s_JTS}). 
Thus, using the force sensor transformation $T_f(\bm{\hat{s}}^i, \bm{r}^i)$ in (\ref{eq:senstransf}) and the sum-based observability function $\Gamma_{sum}(\bm{S})$ in (\ref{eq:obsfunc-sum}), the final sensor observability $\bm{s}$ has the form: 

\begin{equation}
\bm{s} = \Gamma_{sum}(\bm{S}) = \sum_{i=1}^{n_s} \bm{\tilde{s}}^i = \sum_{i=1}^{n_s} \begin{bmatrix} \frac{\lvert\bm{\hat{s}}^i_{\theta} \times \bm{r}^i\rvert}{\|\bm{\hat{s}}^i_{\theta} \times \bm{r}^i\|} \\ \lvert\bm{\hat{s}}^i_{\theta}\rvert \end{bmatrix}
%\bm{s} = \sum_{i=1}^{n_s} \begin{bmatrix} \lvert\bm{r}_i \times \bm{\hat{s}}_{\theta,i}\rvert \\ \lvert\bm{\hat{s}}_{\theta,i}\rvert \end{bmatrix}
\label{eq:kinequiv-sens1}
\end{equation}

The term $\bm{\hat{s}}^i_{p}$ is absent from (\ref{eq:kinequiv-sens1}) as it is zero for single-axis joint torque sensors $\bm{\hat{s}}^{\prime}_{\tau z}$. The summation in (\ref{eq:kinequiv-sens1}) can be rewritten in matrix form using the sensor observability matrix $\bm{S}$ multiplied by a $n_s \times 1$ vector of ones $\bm{1}$:

\begin{equation}
\bm{s} = \begin{bmatrix} 
%\lvert\bm{r}_1 \times \bm{\hat{s}}_{\theta,1}\rvert & \dots & \lvert\bm{r}_{n_s} \times \bm{\hat{s}}_{\theta,n_s}\rvert \\ 
\frac{\lvert\bm{\hat{s}}^1_{\theta} \times \bm{r}^1\rvert}{\|\bm{\hat{s}}^1_{\theta} \times \bm{r}^1\|} & \dots & \frac{\lvert\bm{\hat{s}}^{n_s}_{\theta} \times \bm{r}^{n_s}\rvert}{\|\bm{\hat{s}}^{n_s}_{\theta} \times \bm{r}^{n_s}\|} \\
\lvert\bm{\hat{s}}^1_{\theta}\rvert & \dots & \lvert\bm{\hat{s}}^{n_s}_{\theta}\rvert 
\end{bmatrix}
\begin{bmatrix} 1 \\ \vdots \\ 1 \end{bmatrix}_{n_s \times 1} = \bm{S}\bm{1}_{n_s \times 1}
\label{eq:kinequiv-sens2}
\end{equation}

For the kinematic analysis, we begin with the geometric velocity analysis \cite{Spong2020-RobotModelingandControl}:

\begin{equation}
\begin{bmatrix} \bm{v} \\ \bm{\omega} \end{bmatrix} = \sum_{k = 1}^{n_q} \begin{bmatrix} \dot{q}_k\bm{\hat{z}}_k \times \bm{r}_k \\ \dot{q}_k\bm{\hat{z}}_k \end{bmatrix}
\label{eq:kinequiv-kin1}
\end{equation}

\noindent where $\bm{v}$ and $\bm{\omega}$ are the translational and angular velocities of the end effector, $\dot{q}_k$ is the angular velocity of the $k$-th joint, and $\bm{\hat{z}}_k$ is the $k$-th joint axis where $k \in 1...n_q$ and $n_q$ is the number of joints.
Similar to (\ref{eq:kinequiv-sens1})-(\ref{eq:kinequiv-sens2}), (\ref{eq:kinequiv-kin1}) may be rewritten in matrix multiplication form using the kinematic Jacobian $\bm{J}$ and the vector of joint angular velocities $\bm{\dot{q}}$:

\begin{equation}
\begin{bmatrix} \bm{v} \\ \bm{\omega} \end{bmatrix} = \begin{bmatrix} \bm{\hat{z}}_1 \times \bm{r}_1 & \dots & \bm{\hat{z}}_{n_q} \times \bm{r}_{n_q} \\
 \bm{\hat{z}}_1 & \dots & \bm{\hat{z}}_{n_q} \end{bmatrix} \begin{bmatrix} \dot{q}_1 \\ \vdots \\ \dot{q}_{n_q} \end{bmatrix} = \bm{J}\bm{\dot{q}}
\label{eq:kinequiv-kin2}
\end{equation}

Given that the joint torque sensors $\bm{\hat{s}}^i_{\theta}$ and joint axes $\bm{\hat{z}}_k$ are unit vectors and collinear, we then in fact have $\bm{\hat{s}}^i_{\theta} = \bm{\hat{z}}_k$, $\bm{r}^i = \bm{r}_{k}$, and $n_q = n_s$.
Thus, equations (\ref{eq:kinequiv-sens2}) and (\ref{eq:kinequiv-kin2}) have very similar form despite differences in normalization where $\bm{S} \approx \bm{J}$ for a standard serial manipulator with joint torque sensors on each rotational joint.
To understand this relationship, joint axes could potentially be thought of as velocity measurement sensors.
A similar analysis holds for prismatic joints paired with single axis load cell.

Despite the similarity in form of (\ref{eq:kinequiv-sens2}) and (\ref{eq:kinequiv-kin2}), it is important to note that a sensor observability singularity does not necessarily imply kinematic singularity and vice versa.
%, especially in a general case where sensors other than joint torque sensors are used. 
For example, the configuration shown in Fig. \ref{fig:robotconfig-kinematicsingularity} is a kinematically singular configuration where axes 1 and 7 are collinear and $w_k = 0$, but is not an observability singularity $o \neq 0$ as the red force ellipsoid is thin but not flat. 
Conversely, the joint configurations shown in Figs. \ref{fig:robotconfig-singularity} and \ref{fig:robotconfig-singsensX} are both observability and kinematic singularities, demonstrating potential overlaps between the two indices for this particular robot and sensor configuration.

\begin{figure}[t]
\centering
\includegraphics[width=0.41\textwidth]{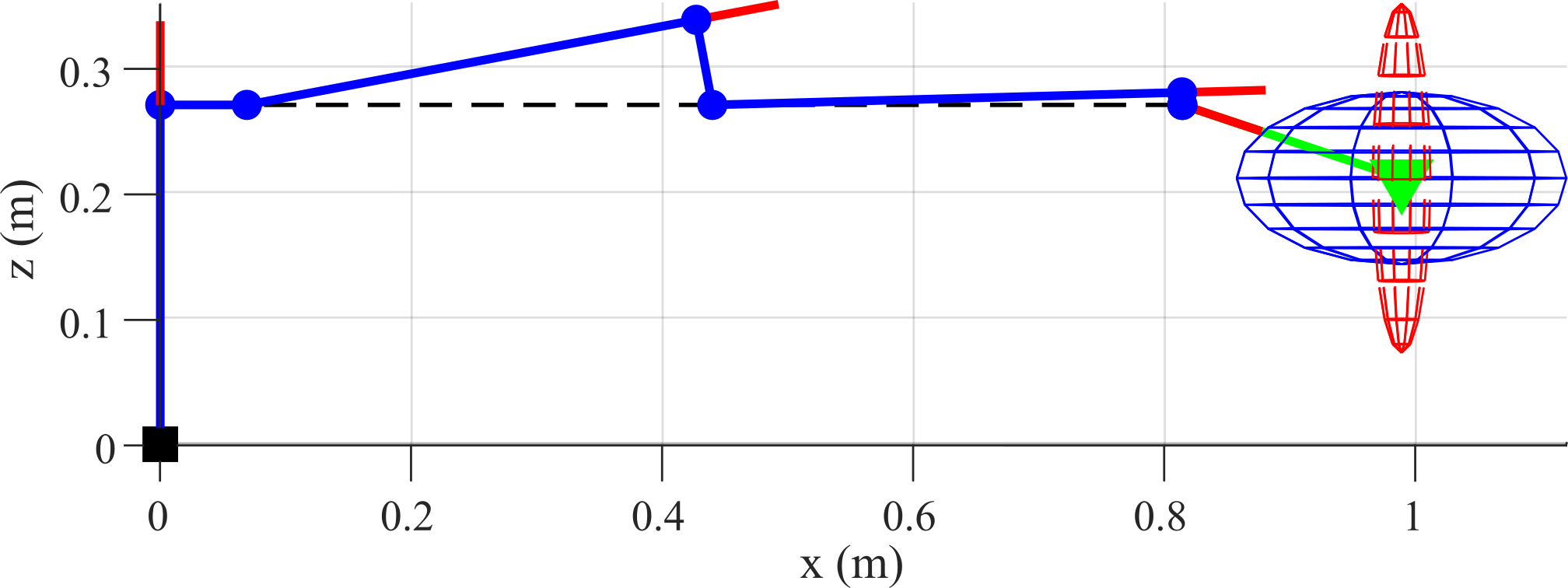}
\caption{Configuration where a non-zero null space vector exists for $\bm{J}^T$, but it is not a sensor observability singularity as $o \neq 0$. This configuration is similar to Fig. \ref{fig:robotconfig-singsensX} but $q_6$ tilts the end effector is slightly downwards.}
\label{fig:robotconfig-nullspace}
\end{figure}

In this special case for serial manipulator robots, it is sometimes possible to extract similar information using the end effector force and joint torque relationship $\bm{\tau} = \bm{J}^T \bm{f}$ and examining the null space of $\bm{J}^T$.
%, it is possible determine if any task space axes are unobservable.
The existence of non-zero null space vectors $\bm{J}^T$ indicates the possibility of having zero joint torques despite non-zero end effector forces, but it can be caused by two distinct cases.
One case is a sensing deficiency in the same manner as sensor observability. 
The other case occurs when end-effector forces and torques balance each other out and result in zero readings at the joints.
For example, in the configuration shown in Fig. \ref{fig:robotconfig-nullspace}, a null space analysis of $\bm{J}^T$ indicates that a force applied at the end effector in the $x$-axis can be nullified with a balancing torque in the $y$-axis, which result in zero joint torques.
Despite the existence of a non-zero null space vector for $\bm{J}^T$ in this configuration, this pose is not an observability singularity as $o_{sum} = 57.63$ (so $o \neq 0$).
Thus, even in the special case where the sensor axes are collinear with the joint axes, null space analysis of the Jacobian cannot replace sensor observability analysis.
The reason is due to the use of absolute values in (\ref{eq:senstransf}) and (\ref{eq:kinequiv-sens1}) during sensor observability analysis that negates the possibility of the sensor axes cancelling each other out.

%In Fig \ref{fig:robotconfig-kinematicsingularity}, although the robot is in a kinematic singular configuration, the sensor observability $\bm{s} \neq 0$ albeit it is quite small.
%In c), although the robot is in a kinematic singular configuration, it is not in a sensor singularity.

%%%%%%%%%%%%%%%%%%%%%%%%%%%%%%%%%%%%%%%%%%%%%%%%%%%%%%%%%%%%%
\subsection{Advantages of Sensor Observability Analysis}
\label{sec:SOI-Advantages}

%\hl{Discussion section about flexibility of SOI vs regular for Jacobian?}
%\hl{Further differences and advantages are afforded through analysis using sensor observability.}

While the discussion above might lead one to think that sensor observability can be derived using the traditional kinematic analysis, the parallels drawn in Sec. \ref{sec:parallels} are applicable only in the special case where the sensor axes are collinear with the joint axes (e.g. standard serial manipulators). 
%there are some key differences.
The advantage of sensor observability analysis is that it is flexible and applicable to robot architectures using different and non-traditional sensor mounting styles. 
Once non-traditional sensor mounting styles are used and there is no longer the one-to-one mapping between joint axes and sensors axes, then the parallels with the traditional Jacobian formulation no longer apply.
For example, a robot using a load cell in the middle of a link, similar to the one shown in Fig. \ref{fig:simulation3Rrobot}, could be a lower cost alternative to using a joint-mounted torque sensor if interaction torques are not of importance.
In this case, since non-joint-mounted sensors are present, the formulation of $\bm{S}$ and $o$ will differ significantly from $\bm{J}$ and $w_k$.
An example highlighting this case will be discussed in Sec. \ref{sec:planarRRRrobot}.
%, $w_s$,
%\hl{The same information can no longer be gleaned from the traditional Jacobian matrix.}

Moreover, certain sensors may need to be interpreted differently than simply with axis direction, which is why the sensor transformation $T_\square(\bm{\hat{s}}^i, \bm{r}^i)$ and sensor observability $\Gamma_\square(\bm{S})$ functions are implemented.
For example, an articulated robot may be covered with an array of distributed laser distance sensors \cite{Stavridis2020ICHR-MobileManipDistributedDistanceSensors}, ultrasonic sensors, or magnetic directional proximity sensors \cite{Wu2016AIM-MagneticProximitySensorSphericalRobot}.
Depending on the joint configuration, these sensors may interact with each other (e.g. ultrasonic interference or crossing magnetic fields), and these interactions may affect sensing quality.
Their interactions could thus be modelled and captured by the $T_\square(\bm{\hat{s}}^i, \bm{r}^i)$ and $\Gamma_\square(\bm{S})$ functions and used to optimize joint configuration to minimize interference.
These complexities will be explored in future work.

Many limitations associated with manipulability analysis using the Jacobian matrix \cite{Patel2015JIRS-SurveyManipulatorPerfMeasures} are not present in sensor observability analysis.
For example, while the kinematic manipulability index $w_k$ suffers from unboundedness and both scale and dimensional dependencies, sensor observability using the row-wise \emph{max} function (\ref{eq:obsfunc-max}) does not since values are bounded between 0 and 1.
The caveat is that some other sensor observability functions, for example the row-wise \emph{sum} function (\ref{eq:obsfunc-sum}), are unbounded.
Calculations with sensor axes are generally normalized and unaffected by scale and dimensions\footnote{The exception is that there is potentially an indirect influence of scale and dimension dependency when sensor noise thresholding is used (Sec. \ref{sec:noise}).}. 

Thus, when analysing task space observability, our proposed sensor observability analysis can be viewed as a more generalized and more flexible analysis than the traditional Jacobian-based analysis.
%Although optimal sensor placement is a well-researched problem in distributed sensing, 
Sensor observability analysis can also be potentially used in the robot design phase to optimize the placement of sensors to create redundancy or minimize the number of sensors required, which is especially applicable to soft robots \cite{Spielberg2021RAL-LearningSensorPlacementSoftRobots}, and will be the subject of future work. 
Despite these comparisons with kinematic analysis using the Jacobian, it must be stressed that sensor observability analysis cannot be used for kinematic analysis and vice versa, and thus it is in fact inappropriate to compare the two directly.

\subsection{Sensor Observability Threshold and Sensor Noise}
\label{sec:noise}

As sensor observability is a continuous quality measure, it is difficult to pinpoint an exact threshold to state when observability has been lost.
An ideal sensor with infinite sensitivity and zero noise would be able to provide a usable reading for all non-zero sensor observability values.
In reality, sensors have finite sensitivity and are susceptible to sensor noise.
A poorly observed axis will have poor signal-to-noise ratio and will effectively be unable to provide meaningful values below a certain threshold $s^{i,*}_j$, denoted by the asterisk. 
Sensor observability for sensor $i$ in the $j$-th task-space axis could be flagged as lost when $s^i_j < s^{i,*}_j$, where the signal noise is greater than a defined required minimum detectable amount $\Phi^i_j$ as a function of the sensor sensitivity.
A system could have multiple types of sensors, e.g., a mix of different models of load cells and joint torque sensors, each with their own specifications.
There then exists individual threshold value $s^{i,*}_j$ for each sensor $i$ and each axis $j$ (see Appendix for derivation): 

\begin{equation}
s^{i,*}_j = \frac{\sigma^i_{\epsilon}}{\Phi^i_j}
%s^{i,*}_j = \frac{\sigma_{\epsilon}}{\kappa \Phi^i_j}
\label{eq:sensnoise}
\end{equation}

%$\kappa$ is the sensor sensitivity,
\noindent where $\sigma_{\epsilon}$ is the standard deviation of the sensor noise and $\Phi^i_j$ is the desired minimum observable quantity for sensor $i$ in the $j$-th task-space axis. 
$\Phi^i_j$ is a user-defined design parameter based on the task definition for each specific axis and the sensor specifications.
%This minimum quantity $\Phi^i_j$ is dependent on the sensor type and task-space axis.
%\hlc{If $\Phi^i_j\rightarrow 0$ then $s^{i,*}_j\rightarrow \infty$} %\hl{This shouldn't matter because phi should never be 0. Small phi should be balanced by the appropriate sensitivity (or else the hardware or desired characteristics combo is infeasible).}
%Also, make sure $s^*$ is never over 1, if $s^* > 1$, you only stay in 0 case. 
%for load cells (LC) with sensitivity $\kappa = 20$ mV/N and noise levels of $\sigma_{\epsilon} = 10$ mV
For example, we require that a system must be able to detect interaction forces of at least $\Phi^{LC} = F_{min} = 10$ N using a load cell (LC) with noise levels of $\sigma_{\epsilon} = 0.5$ N.
The threshold for sensor observability would then be:

\begin{equation*}
s^{LC,*} = \frac{\sigma_{\epsilon}}{\Phi^{LC}} = \frac{(0.5\text{ N})}{(10\text{ N})} = 0.05
%s^{LC,*} = \frac{\sigma_{\epsilon}}{\kappa \Phi^{LC}} = \frac{(10\text{ mV})}{(20\ \frac{\text{mV}}{\text{N}})(10\text{ N})} = 0.05
%\label{eq:sensnoise_example}
\end{equation*}

\noindent where sensor observability is considered lost if $s^{LC} < 0.05$. 

If $s^{i,*}_j > 1$, it means that it is impossible for that sensor to detect the minimum desired quantity $\Phi^i_j$ as $s^{i}_j$ has an upper bound of 1.
%Thus, the following constraint should be followed to ensure $s^{i,*}_j < 1$:
Thus, to ensure $s^{i,*}_j \leq 1$, the constraint $\sigma_{\epsilon} \leq \Phi^i_j$ should be followed.
%\begin{equation}
%\frac{\sigma_{\epsilon}}{\kappa} < \Phi^i_j
%\label{eq:sensnoise_phiconstraint}
%\end{equation}
Having $s^i_j$ below the threshold simply indicates that the \emph{minimum} quantity may not be individually observable by the $i$-th sensor anymore;
%It is important to note that $\Phi^i_j$ is the desired \emph{minimum} observable quantity for the $i$-th sensor; 
conversely, it is possible for much larger values $F_{j,actual} \gg \Phi^i_j$ to be observable even when $s^i_j < s^{i,*}_j$.
In addition, from the perspective of the entire system, other sensors may be able to compensate for any sensor $s^i_j < s^{i,*}_j$ if they are positioned correctly.

Although $\Phi^i_j$ is a design parameter, it is not necessarily defined in a straightforward manner for different types of sensors.
Recall that the force sensor transformation in (\ref{eq:senstransf}) has both translational and rotational components $\bm{\tilde{s}}^i = \begin{bmatrix} \bm{\tilde{s}}^i_{p} & \bm{\tilde{s}}^i_{\theta} \end{bmatrix}^T$.
For joint torque sensors, measuring torque is a straightforward transformation as $\bm{\tilde{s}}^{JTS}_{\theta} \approx \lvert\bm{\hat{s}}^{JTS}_{\theta}\rvert$ such that $\Phi^{JTS}_\theta = \tau_{min}$, similar to $\Phi^{LC}_p = F_{min}$.

Conversely, measuring linear forces using a torque sensor is influenced by the moment arm, which cannot be ignored.
We can calculate the effect of the cross product as well as the signal amplification as a result of the moment arm using the quantity $\bm{c}^{JTS}_{p} = \lvert\bm{\hat{s}}^{JTS}_{\theta} \times \bm{r}^{JTS}\rvert$.
%, which is similar to (\ref{eq:senstransf}) but without the normalization.
Thus, for minimum linear forces measured by torque sensors in the $j$-th axis, we have $\Phi^{JTS}_{p,j} = F_{min}\ c^{JTS}_{p,j}$.
%\odot where $\odot$ is the element-wise Hadamard product.
Other definitions of $s^{i,*}_j$ are possible and other external factors could also be factored into the noise term when calculating the minimum required sensor observability threshold, e.g., ultrasonic distance sensors interfering with each other.
%Additionally, numerical robustness may play a factor as well, e.g. very high condition numbers for $\bm{S}$ could affect the trustworthiness of the sensor values.

To factor in the sensor observability threshold, an extra step is added after calculating the transformed sensor axes $\bm{\tilde{s}}^i = T_\square(\bm{\hat{s}}^i, \bm{r}^i)$ in (\ref{eq:senstransf}), as shown in Algorithm \ref{alg:sens_obs}:

\begin{equation}
%o = \prod_{j = 1}^{n_t} f(s_j, s^*_j) \\
\tilde{s}^i_j = f(\tilde{s}^i_j, s^{i,*}_j)\quad \forall\ i,j
\label{eq:obsindex_noise}
\end{equation}

\begin{equation*}
\text{where } f(\tilde{s}^i_j, s^{i,*}_j) = 
	\begin{cases}
      0  \text{,} &\text{if } \tilde{s}^i_j \leq s^{i,*}_j \\
      \frac{\tilde{s}^i_j - s^{i,*}_j}{1 - s^{i,*}_j}  \text{,} &\text{if } \tilde{s}^i_j > s^{i,*}_j \\ 
	\end{cases}
\end{equation*}

The piece-wise defined function is used to threshold each individual sensor along each task-space axis.
Sensor observability values below the threshold are set to 0 and the range above the threshold is scaled back to between 0 and 1.
All subsequent steps from (\ref{eq:sensmatrix}) onwards in calculating the system sensor observability $\bm{s}$ and index $o$ remain the same, where either $s_j = 0$ or $o = 0$ indicates that at least one task-space axis is no longer observable.
The effect of sensor observability thresholding is similar to a deadband. 

It is important to note that for the purpose of sensor observability analysis, noise is only considered to affect the calculation of $s^{i,*}_j$; its direct effect on the sensor readout itself and the need for filtering are not considered. 
In Sec. \ref{sec:physicalexperiments}, a real-world force reconstruction case demonstrates that the reconstructed value only matches the ground truth when $s^i_j$ is greater than a specific value, though this value could be unique to the particular robot, sensor type, joint configuration, experiment, and reconstruction method.

%\hlc{Example with gaussian white noise? Can $J^T$ do the same thing?}

%\hlc{Can you add a gaussian white noise to the signal in the experiment section to support this point? Do you think sensor observability will be able to flag an observability loss while $J^T$ cannot in this case? If my understanding is correct, the sensor observability depends on the sensor axes and not the reading from the sensors, so how the sensor noise can be considered?}

%%%%%%%%%%%%%%%%%%%%%%%%%%%%%%%%%%%%%%%%%%%%%%%%%%%%%%%%%%%%
\subsection{Non-Traditional Robot Architectures} \label{sec:planarRRRrobot}

\begin{figure}[t]
\centering
\subfigure[]{
	\includegraphics[width = 0.36\textwidth]{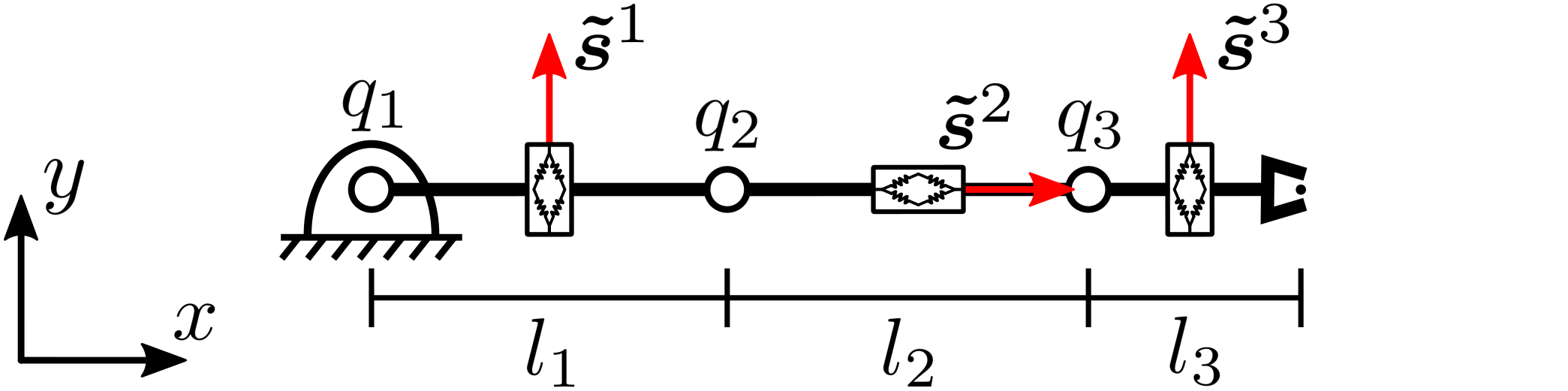}
	\label{fig:simulation3Rrobot-structure}}
\caption{Special planar RRR robot with 3 revolute joints (without joint torque sensing) and 3 single axis load cells located on each link.
$\bm{\tilde{s}}^1$ and $\bm{\tilde{s}}^3$ are aligned perpendicularly to the link while $\bm{\tilde{s}}^2$ is parallel with the second link.
}
\label{fig:simulation3Rrobot}
\end{figure}

To demonstrate the utility of sensor observability analysis, we simulate the following planar RRR robot shown in Fig. \ref{fig:simulation3Rrobot}, which has 3 revolute joints (without joint torque sensing) and 3 single axis load cells located on each link.
$\bm{\hat{s}}^1$ and $\bm{\hat{s}}^3$ are perpendicular to their respective links while $\bm{\hat{s}}^2$ is parallel with the second link.
Such a structure could be a potential design for cheaper upper body rehabilitation robots \cite{Demofonti2021TMRB-AffordableStrokeRehabRobots}. The use of single-axis load cells aligned non-traditionally could be a cheaper alternative to using joint torque sensors or a multi-axis sensor.
In this example, we are only interested in linear forces in the $x$- and $y$-axes and not torques in $z$. 
Thus, $n_t = 2$ and we have $\bm{\hat{s}}^1 = \bm{\hat{s}}^1_{p} = \begin{bmatrix} 0 & 1 \end{bmatrix}^T$, $\bm{\hat{s}}^2 = \bm{\hat{s}}^2_{p} = \begin{bmatrix} 1 & 0 \end{bmatrix}^T$, and $\bm{\hat{s}}^3 = \bm{\hat{s}}^3_{p} = \begin{bmatrix} 0 & 1 \end{bmatrix}^T$.
We then obtain the following according to the force sensor transformation $T_{f}(\bm{\hat{s}}^i, \bm{r}^i)$:

\begin{equation}
\bm{\tilde{s}}^i = T_{f}(\bm{\hat{s}}^i, \bm{r}^i) = \lvert\bm{\hat{s}}^i_{p}\rvert
\label{eq:sim3R-transform}
\end{equation}

\noindent Since these are only linear sensors, $\bm{\hat{s}}^i_{\theta}$ from (\ref{eq:senstransf}) does not exist. 
If we include the rotation into the task frame, we obtain:

\begin{equation}
\begin{gathered}
\bm{\tilde{s}}^1 = \lvert \bm{R}_0^1 \bm{\hat{s}}^1_{p}\rvert = \begin{bmatrix} \lvert -\mathcal{S}_1 \rvert \\ \lvert \mathcal{C}_1 \rvert \end{bmatrix} \\
 %= T_{f}(\bm{\hat{s}}^1)
\bm{\tilde{s}}^2 = \lvert \bm{R}_0^2 \bm{\hat{s}}^2_{p}\rvert = \begin{bmatrix} \lvert \mathcal{C}_1 \mathcal{C}_2 - \mathcal{S}_1 \mathcal{S}_2 \rvert \\ \lvert \mathcal{C}_1 \mathcal{S}_2 + \mathcal{C}_2 \mathcal{S}_1 \rvert \end{bmatrix} \\
\bm{\tilde{s}}^3 = \lvert \bm{R}_0^3 \bm{\hat{s}}^3_{p}\rvert = \begin{bmatrix} \lvert -\mathcal{C}_1 \mathcal{C}_2 \mathcal{S}_3 + \mathcal{C}_3 \mathcal{S}_2 - \mathcal{S}_1 \mathcal{C}_2 \mathcal{C}_3 - \mathcal{S}_2 \mathcal{S}_3 \rvert \\ 
\lvert \mathcal{C}_1 \mathcal{C}_2 \mathcal{C}_3 - \mathcal{S}_2 \mathcal{S}_3 - \mathcal{S}_1 \mathcal{C}_2 \mathcal{S}_3 + \mathcal{C}_3 \mathcal{S}_2 \rvert \end{bmatrix} \\
%\text{where } s_i = \text{sin}(q_i) \text{ and } c_i = \text{cos}(q_i)
\end{gathered}
\label{eq:sim3R-transform2}
\end{equation}

\noindent where $\mathcal{S}_i = \text{sin}(q_i)$, $\mathcal{C}_i = \text{cos}(q_i)$ and $\bm{R}_a^b$ is the rotation from frame $a$ to frame $b$.
The sensor observability matrix becomes:

\begin{equation}
\begin{aligned}
\bm{S} &= \begin{bmatrix} \bm{\tilde{s}}^1 & \bm{\tilde{s}}^2 & \bm{\tilde{s}}^3 \end{bmatrix} \\
&= \left[\begin{matrix} 
\lvert -\mathcal{S}_1 \rvert, & \lvert \mathcal{C}_1 \mathcal{C}_2 - \mathcal{S}_1 \mathcal{S}_2 \rvert, \\
\lvert \mathcal{C}_1 \rvert, & \lvert \mathcal{C}_1 \mathcal{S}_2 + \mathcal{C}_2 \mathcal{S}_1 \rvert, \end{matrix} \right.\\
&\qquad\qquad
\left. \begin{matrix} 
\lvert -\mathcal{C}_1 \mathcal{C}_2 \mathcal{S}_3 + \mathcal{C}_3 \mathcal{S}_2 - \mathcal{S}_1 \mathcal{C}_2 \mathcal{C}_3 - \mathcal{S}_2 \mathcal{S}_3 \rvert \\
\lvert \mathcal{C}_1 \mathcal{C}_2 \mathcal{C}_3 - \mathcal{S}_2 \mathcal{S}_3 - \mathcal{S}_1 \mathcal{C}_2 \mathcal{S}_3 + \mathcal{C}_3 \mathcal{S}_2 \rvert \end{matrix}\right]
\end{aligned}
\label{eq:sim3R-transform3}
\end{equation}

\noindent Conversely, if we calculate the standard kinematic Jacobian for linear motion in the $x$- and $y$-axes only, we obtain:

\begin{equation}
\begin{aligned}
\bm{J} &= \begin{bmatrix} \bm{\hat{z}}_1 \times \bm{r}_1 & \bm{\hat{z}}_2 \times \bm{r}_2 & \bm{\hat{z}}_3 \times \bm{r}_3 \end{bmatrix} \\
&= \left[\begin{matrix} 
& - l_1\mathcal{S}_1 - l_2\mathcal{S}_{12} - l_3\mathcal{S}_{123}, \\
&  l_1\mathcal{C}_1 + l_2\mathcal{C}_{12} + l_3\mathcal{C}_{123}, \end{matrix} \right.\\
&\qquad\qquad
\left. \begin{matrix} 
& - l_2\mathcal{S}_{12} - l_3\mathcal{S}_{123}, & - l_3\mathcal{S}_{123} \\
&   l_2\mathcal{C}_{12} + l_3\mathcal{C}_{123}, & + l_3\mathcal{C}_{123}  \end{matrix}\right]
\end{aligned}
\label{eq:sim3R-transform4}
\end{equation}

\noindent where the multiple subscripts indicate angle summation, e.g., $\mathcal{S}_{123} = \text{sin}(q_1 + q_2 + q_3)$.
Clearly, we can see that see that the sensor observability matrix $\bm{S}$ and the standard kinematic Jacobian $\bm{J}$ no longer match as the robot in Fig. \ref{fig:simulation3Rrobot} does not have a one-to-one mapping between joints and sensors.
The system sensor observability $\bm{s}$ and sensor observability index $o$ are then calculated using using their respective equations described in Sec. \ref{sec:sensobs}.
 %$= \Gamma_{sum}(\bm{S})$ and $o$}, (\ref{eq:obsfunc-sum}) and (\ref{eq:obsindex}) respectively

%%%%%%%%%%%%%%%%%%%%%%%%%%%%%%%%%%%%%%%%%%%%%%%%%%%%%%%%%%%%%
%%%%%%%%%%%%%%%%%%%%%%%%%%%%%%%%%%%%%%%%%%%%%%%%%%%%%%%%%%%%%
%\section{Simluations \& Experiments}
%\label{sec:experiments}

%%%%%%%%%%%%%%%%%%%%%%%%%%%%%%%%%%%%%%%%%%%%%%%%%%%%%%%%%%%%%
%%%%%%%%%%%%%%%%%%%%%%%%%%%%%%%%%%%%%%%%%%%%%%%%%%%%%%%%%%%%%
%\section{\hl{SOA applied to a }3 DOF Planar Robot}
\section{Maximizing Sensor Observability} \label{sec:SOoptimization}
In this section, our focus is on exploring the integration of maximizing sensor observability in conjunction with solving a kinematics task. 
We provide two separate formulations: 1) in the null space of the Jacobian matrix and 2) as an optimization problem. 

%%%%%%%%%%%%%%%%%%%%%%%%%%%%%%%%%%%%%%%%%%%%%%%%%%%%%%%%%%%%%
\subsection{Null Space Formulation} \label{sec:optNullSpace}
\label{sec:simulation3Rrobot}
%\subsubsection{\hl{Structure}}

\begin{figure*}[t]
\centering
\subfigure[Minimizing joint motion using null space formulation]{
	\includegraphics[width = 0.94\textwidth]{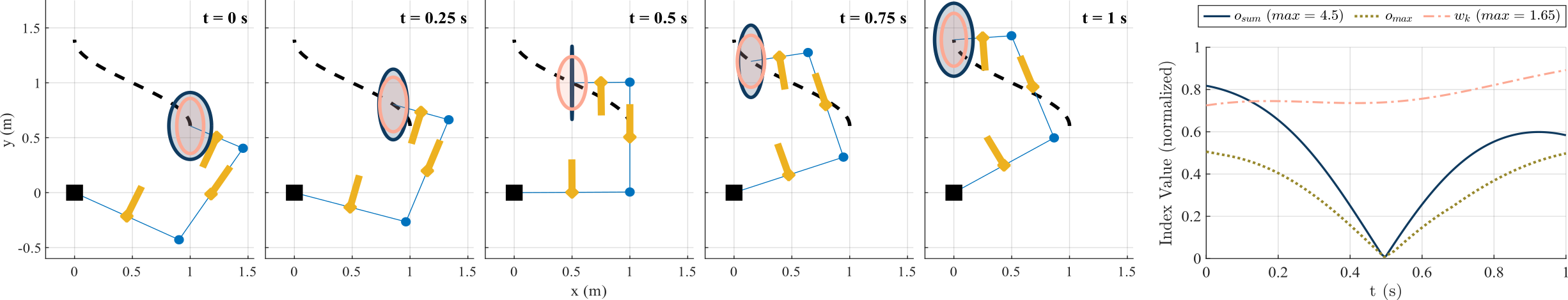}
	\label{fig:sim3dof_OptNone}}
\par\medskip
\subfigure[Maximizing kinematic manipulability $w_k$ using null space formulation]{
	\includegraphics[width = 0.94\textwidth]{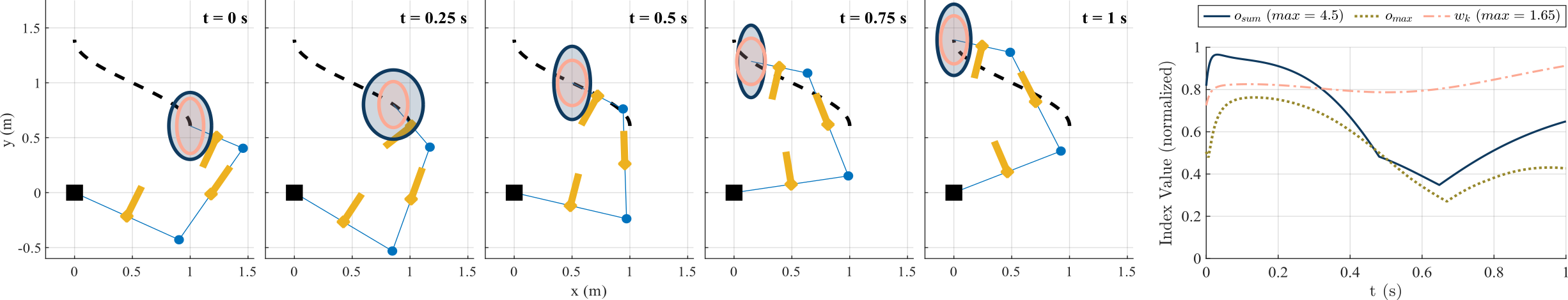}
	\label{fig:sim3dof_OptKin}}
\par\medskip
\subfigure[Maximizing sensor observability $o$ using null space formulation]{
	\includegraphics[width = 0.94\textwidth]{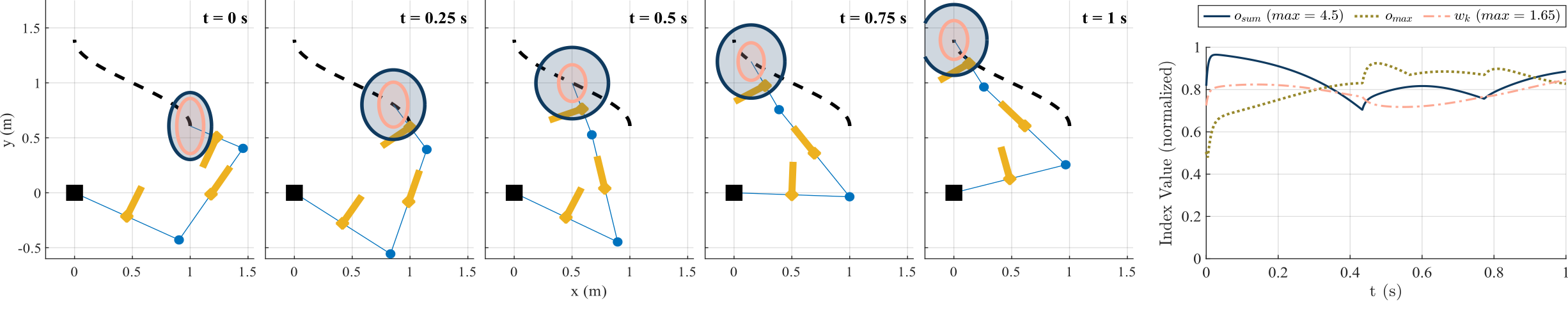}
	\label{fig:sim3dof_OptSO}}
\par\medskip
\subfigure[Maximizing sensor observability $o$ as an optimization problem using quadratic programming]{
	\includegraphics[width = 0.94\textwidth]{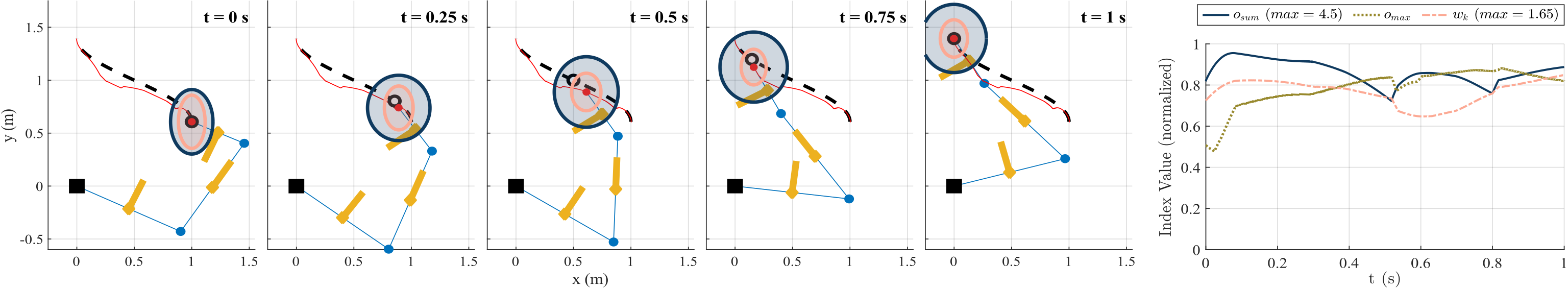}
	\label{fig:sim3dof_OptQP}}
	%with the same initial conditions 
	%different redundancy resolution strategies: 
\caption{Motion of a custom 3 DOF robot using the Jacobian null space to (a) minimize joint motion, (b) maximize kinematic manipulability, and (c) maximize sensor observability, and (d) maximize sensor observability as an optimization problem formulation. 
The black square is the base joint 1 while joints 2 and 3 are blue circles. 
Robot links are thin blue lines while the linear sensor axes are thick orange lines.
The blue ellipse is the sensor observability ellipsoid using $\bm{s}_{sum}$ and the pink ellipse is the kinematic manipulability ellipsoid.
The dotted black line is the desired EE trajectory.
In (d), the black outlined dot is the desired EE position at that time point, and the red line and red dot are respectively the actual EE trajectory and position as a result of the relaxation vector $\delta$.
The plots show the evolution of the sensor observability indices $o_{sum}$ and $o_{max}$ and the kinematic manipulability index $w_k$ through the motion. 
The sum index $o_{sum}$ and kinematic manipulability index $w_k$ are normalized according to their global maxima (4.5 and 1.65, respectively) for plotting purposes.
}
\label{fig:sim3dofmotion}
\end{figure*}

If a robot is considered kinematically redundant, then the null space of the Jacobian matrix can be exploited to satisfy secondary tasks \cite{Siciliano2008TEXT-RoboticsModellingPlanningandControl}, e.g., to maximize sensor observability:
%Given the redundancy of the robot, we can optimize secondary tasks using the 

\begin{equation}
\dot{\bm{q}} = \bm{J}^\dagger \dot{\bm{x}} + (\bm{I} - \bm{J}^\dagger\bm{J}) \dot{\bm{q_*}}
\label{eq:sim3R-Jyesopt}
\end{equation}

\noindent where $\dot{\bm{x}}$ is the end effector Cartesian velocity, $\bm{J}^\dagger$ is the right pseudoinverse of $\bm{J}$, $\bm{I}$ is the identity matrix, and $\dot{\bm{q_*}}$ are the joint velocities related to the secondary task. 
The first term is the solution to the kinematics task that minimizes the norm of joint velocities, while the second term can be used to maximize a secondary task.
To compare the effect of using different redundancy resolution strategies, we simulate the 3-DOF RRR robot shown in Sec. \ref{sec:planarRRRrobot} in MATLAB to follow a sinusoidal trajectory using velocity control. 
 %as shown in Fig. \ref{fig:sim3dofmotion}. 
%Four different redundancy resolution strategies, all from the same starting pose, to illustrate their different effects. 
The results are shown in Figs. \ref{fig:sim3dofmotion} and \ref{fig:sim3dofmotion-axisopt}.
The simulation script is provided online\footnote{ https://github.com/chrisywong/SensorObservabilityAnalysisDataset}.

%\noindent where the second term is the projection into the null space of $\bm{J}$, $\bm{I}$ is the identity matrix, and $\dot{\bm{q_*}}$ is the joint velocities used to increase sensor observability \hl{(wording is off)}.

\subsubsection{Minimize joint motion}
In a first trial, the joint trajectories simply use the right pseudoinverse of the Jacobian $\bm{J}^\dagger$ to minimize joint motion without a secondary task $\dot{\bm{q_*}} = \bm{0}$ such that (\ref{eq:sim3R-Jyesopt}) becomes:

\begin{equation}
\dot{\bm{q}} = \bm{J}^\dagger \dot{\bm{x}}
\label{eq:sim3R-Jnoopt}
\end{equation}

The resulting motion is shown in Fig. \ref{fig:sim3dof_OptNone}.
At $t = 0.5$ s, the robot passes through a sensor observability singularity and the sensor observability index $o \rightarrow 0$.
In this sensor observability singular configuration, linear forces in the $x$-axis cannot be detected by the sensors and may lead to undesirable consequences. 
It is important to note that although the robot is in a sensor observability singular configuration, the robot is not in a kinematically singular configuration as $w_k \neq 0$.
Once the robot moves past the sensor observability singularity at $t > 0.5$ s, forces in $x$ are observable once again.

\subsubsection{Maximize kinematic manipulability}
%In the second and third trials, the redundancy of the robot is used for a secondary task, respectively maximizing kinematic manipulability or sensor observability, through the null space of the Jacobian \cite{Siciliano2008TEXT-RoboticsModellingPlanningandControl}.

%\begin{equation}
%\dot{\bm{q}} = \bm{J}^\dagger \dot{\bm{x}} + (\bm{I} - \bm{J}^\dagger\bm{J}) \dot{\bm{q_*}}
%\label{eq:sim3R-Jyesopt}
%\end{equation}

%\noindent where the second term is the projection into the null space of $\bm{J}$, $\bm{I}$ is the identity matrix, and $\dot{\bm{q_*}}$ is the joint velocities used to increase sensor observability \hl{(wording is off)}.

In the second trial, the secondary task in (\ref{eq:sim3R-Jyesopt}) is used to maximize kinematic manipulability of the end effector.
As such, $\dot{\bm{q_*}}$ is defined as follows using the partial derivatives of the manipulability index $w_k$, as defined in (\ref{eq:manip_kin}),  w.r.t. the joints $\bm{q}$:
%In the second trial, kinematic manipulability is maximized using the null space projection where 

\begin{equation}
\dot{\bm{q_*}} = k_0\frac{\delta w_k}{\delta\bm{q}}
\label{eq:sim3R-q0kin}
\end{equation}

\noindent where $k_0$ is a positive scalar coefficient.
The resulting motion is shown in Fig. \ref{fig:sim3dof_OptKin}.
Although the effect is not overly pronounced, there are small differences in joint trajectories compared to the first trial that result in an overall increase in kinematic manipulability throughout the motion as well as a higher maximum $w_k$.
There is also the unintended effect that the robot no longer passes through the sensor observability singularity, but sensor observability is not explicitly maximized. 
The discontinuous $o$ profiles are inflection points in the robot motions that cause the sensor axes to change directions.

\begin{figure*}[t]
\centering
\subfigure[Maximizing sensor observability $o$ in $x$-axis using null space formulation]{
	\includegraphics[width = 0.98\textwidth]{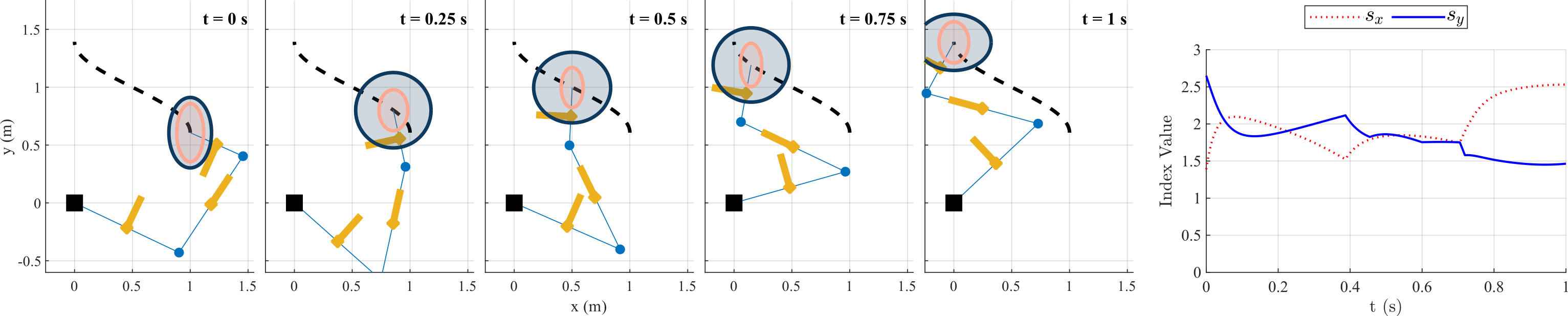}
	\label{fig:sim3dof_OptAxisX}}
\par\medskip
\subfigure[Maximizing sensor observability $o$ in $x$-axis as an optimization problem using quadratic programming]{
	\includegraphics[width = 0.98\textwidth]{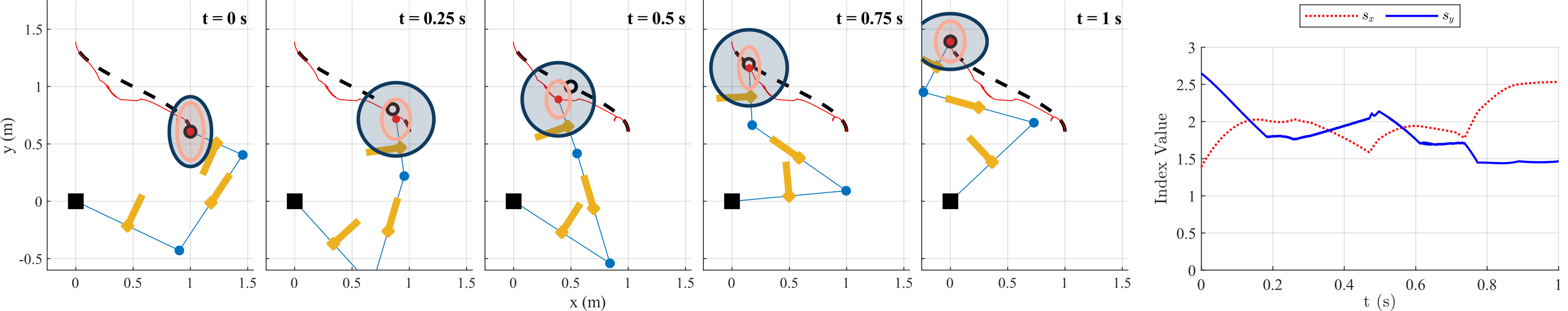}
	\label{fig:sim3dof_OptAxisXQP}}
\par\medskip
\subfigure[Maximizing sensor observability $o$ in $y$-axis using null space formulation]{
	\includegraphics[width = 0.98\textwidth]{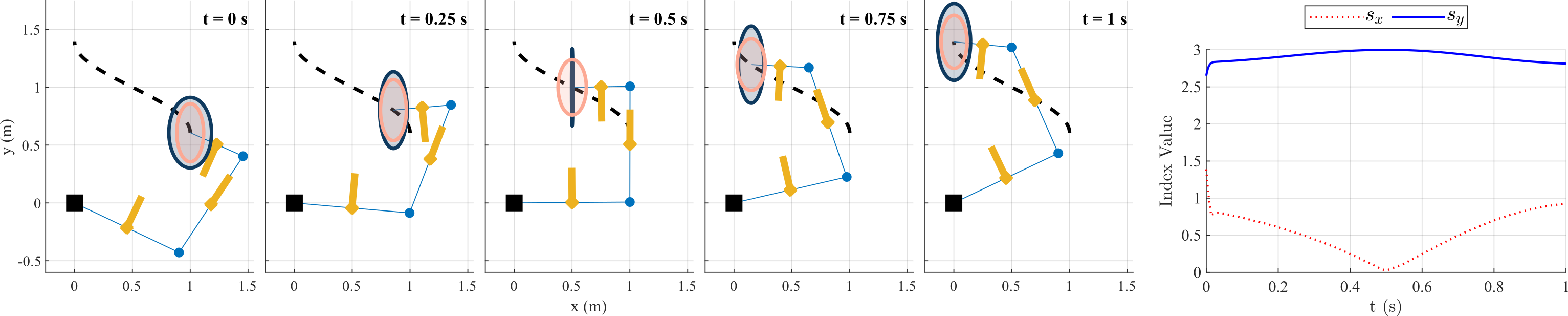}
	\label{fig:sim3dof_OptAxisY}}
\caption{Motion of a 3 DOF robot where sensor observability is maximized only in the $x$-axis using (a) null space formulation and (b) as an optimization problem, and (c) only in the $y$-axis using null space formulation. 
Given the similarities between the two methods in $y$-axis only maximization, only the null space formulation is shown.
Plot shows sensor observability in the $x$- and $y$-axes, $s_x$ and $s_y$ respectively, using the sum method in (\ref{eq:obsfunc-sum}).
}
\label{fig:sim3dofmotion-axisopt}
\end{figure*} 

\subsubsection{Maximize sensor observability} \label{sec:nullspaceSO}

In the third trial, the secondary task in (\ref{eq:sim3R-Jyesopt}) is used to maximize sensor observability instead of kinematic manipulability.
Joint velocities in the Jacobian null space $\dot{\bm{q_*}}$ now use the partial derivatives of the sum-based sensor observability index $o_{sum}$, defined in (\ref{eq:obsindex}): 

\begin{equation}
\dot{\bm{q_*}} = k_0\frac{\delta o_{sum}}{\delta\bm{q}}
\label{eq:sim3R-q0SO}
\end{equation}

The joint trajectories are now optimized to maximize overall sensor observability while maintaining the desired EE trajectory along the sinusoidal path.
The resulting motion is shown in Fig. \ref{fig:sim3dof_OptSO}.
Both $o_{sum}$ and $o_{max}$ have much higher values on average throughout the motion compared to the other two trials as well as a higher maximum value and smaller dips in the middle.
Conversely, $w_k$ is lower than the other two trials as an unintended consequence. 
As expected, maximizing kinematic manipulability in the second trial and maximizing sensor observability in the third trial do not yield the same results as they are different objectives. 

\subsubsection{Maximize sensor observability in specific axes}
%Other trajectory optimization techniques may be used to maximize specific sensor observability axes if needed. 
Sensor observability can also be maximized in specific axes using the system sensor observability $\bm{s}$ rather than as a whole using the sensor observability index $o$.
Fig. \ref{fig:sim3dofmotion-axisopt} shows the changes in robot trajectory when only a single sensor observability axis is included as the secondary task in (\ref{eq:sim3R-Jyesopt}).
Rather than using $\delta o / \delta \bm{q}$ in (\ref{eq:sim3R-q0SO}), only a specific axis $s_j$ is used.

\begin{equation}
\dot{\bm{q_*}} = k_0\frac{\delta s_j}{\delta\bm{q}}
\label{eq:sim3R-q0SOaxis}
\end{equation}

In this case, we use the sum function (\ref{eq:obsfunc-sum}) to calculate $s_x$ and $s_y$.
As expected, maximizing $s_x$ skews the sensor observability ellipsoid in the $x$-axis, as seen in Fig. \ref{fig:sim3dof_OptAxisX}.
While maximizing $s_y$ in Fig. \ref{fig:sim3dof_OptAxisY} has certain similarities to the first trial shown in Fig. \ref{fig:sim3dof_OptNone}, the joint trajectories differ slightly.
Ellipsoid shaping to simultaneously achieve specific $\bm{s}$ is also possible using the methods in \cite{Jaquier2021IJRR-GeometryAwareManipulabilityLearning} and will be the subject of future work.

\begin{figure}[t]
\centering
\subfigure[]{
	\includegraphics[width = 0.48\textwidth]{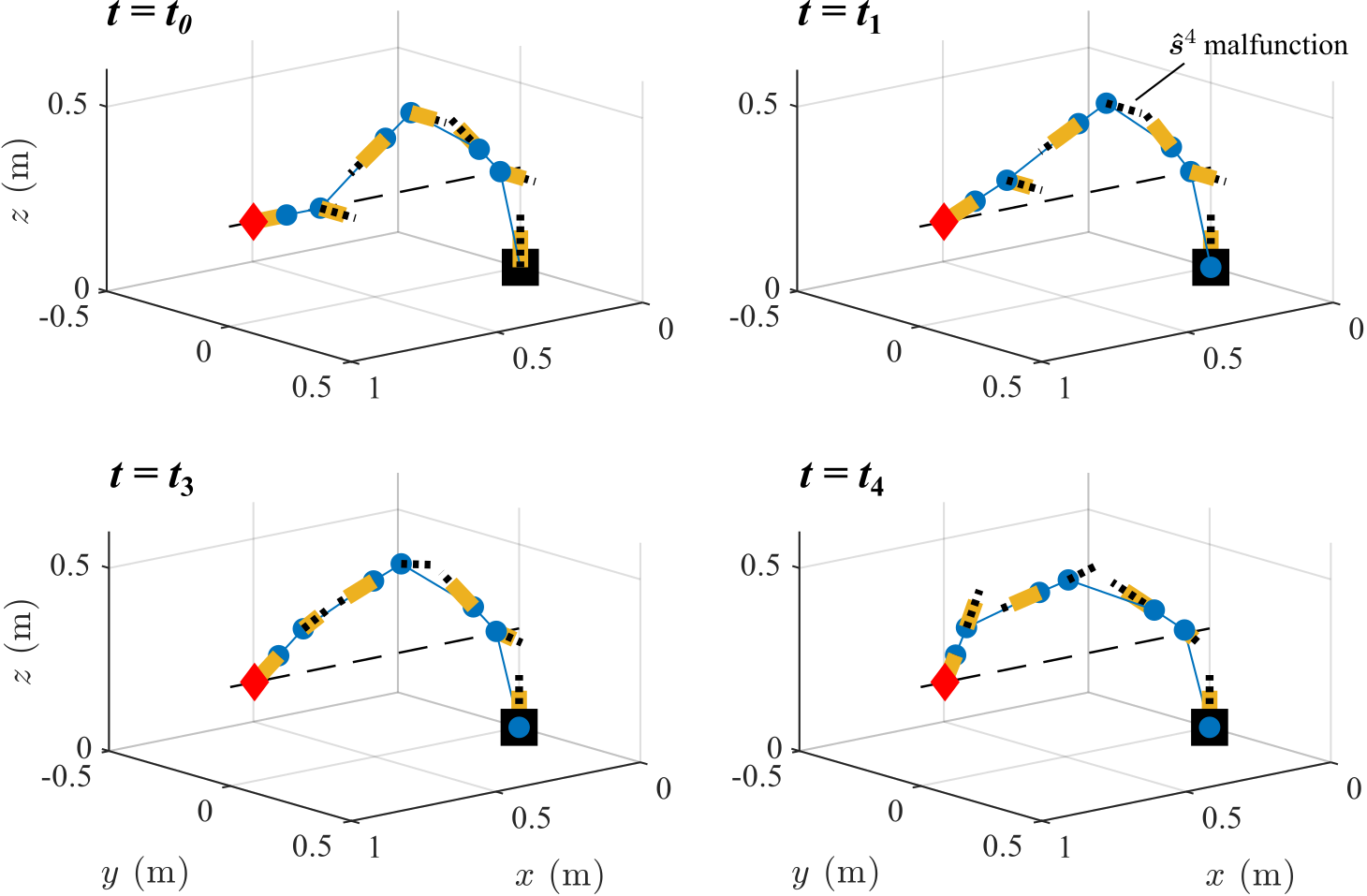}
	\label{fig:simBaxterJointDeficiencyA}}
\subfigure[]{
	\includegraphics[width = 0.36\textwidth]{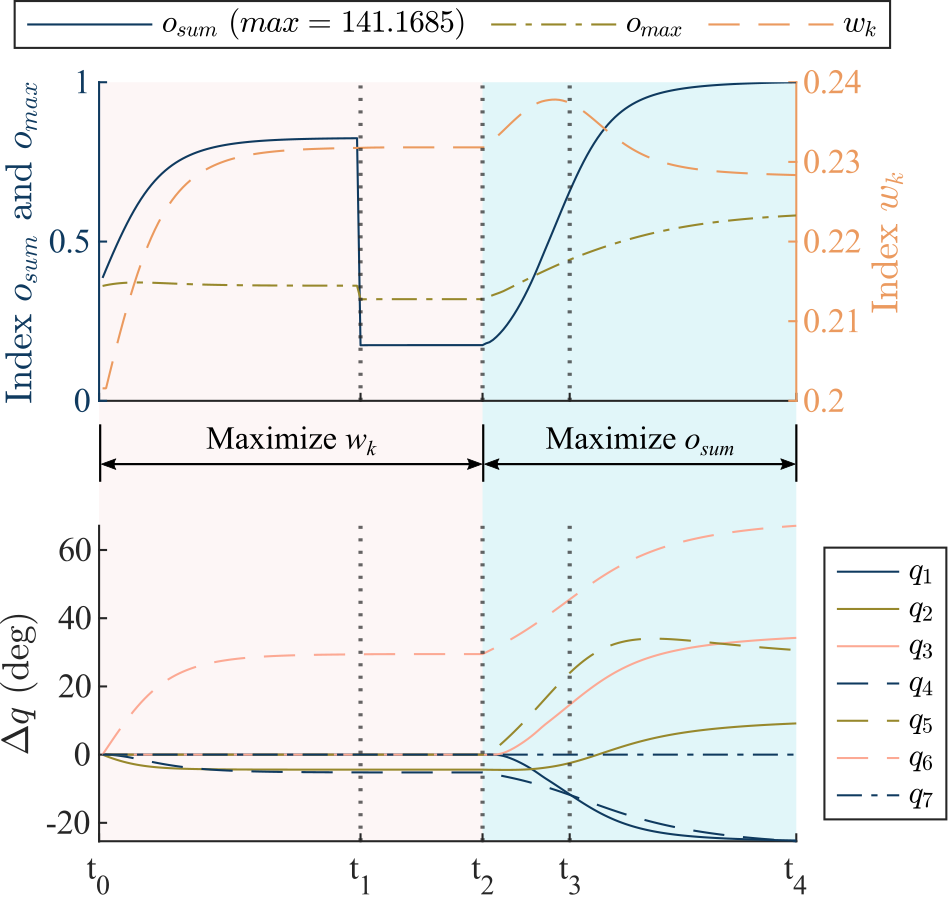}
	\label{fig:simBaxterJointDeficiencyB}}
\caption{(a) Simulation of Baxter robot posture optimization using null-space projection but with a sensor deficiency.
Blue lines and circles are the robot links and joints respectively with the black square base and red diamond end effector.
The black dotted lines are joint axes and the solid yellow lines are sensor axes (as Baxter uses joint torque sensors, each joint and sensor pair is collinear);
(b) Plot of various performance indices and the joint angles.
$o_{sum}$ is normalized to the maximum of 141.1685.
Between $t = \begin{bmatrix} t_0 & t_2 \end{bmatrix} $, the robot uses the null-space projection to maximize kinematic manipulability $w_k$.
At $t = t_1$, we simulate a removal of the joint torque sensor at joint $q_4$ such that the sensor no longer functions, but the joint is still able to move and be controlled.
Between $t = \begin{bmatrix} t_2 & t_4 \end{bmatrix} $, the robot switches the null-space projection to maximize sensor observability $o_{sum}$ instead.
}
\label{fig:simBaxterJointDeficiency}
\end{figure}

%To further illustrate the utility of sensor observability analysis in a more practical case, we simulate the Baxter robot \hl{while it spontaneously undergoes a sensor malfunction.}
To further illustrate the utility of sensor observability analysis in a more practical case, we simulate the Baxter robot when one of its sensors is removed if, for example, there was a sensor malfunction or if it was removed during cost- or weight-cutting measures.
The robot begins in the posture at $t = t_0$ shown in Fig. \ref{fig:simBaxterJointDeficiencyA}, which is similar to the posture shown in Fig. \ref{fig:expFS-norm} but not identical.
Throughout this simulation, the robot holds the end effector position and uses the null-space projection equations (\ref{eq:sim3R-Jyesopt})-(\ref{eq:sim3R-q0SO}) to maximize different performance metrics. 
During the period $t = \begin{bmatrix} t_0 & t_2 \end{bmatrix}$, the robot uses the null-space projection to increase kinematic manipulability $w_k$.
As expected, $w_k$ increases until it reaches a local maximum and plateaus, shown in Fig. \ref{fig:simBaxterJointDeficiencyB}. 
%At $t = t_1$, we simulate a malfunction with the joint torque sensor of joint $q_4$ such that the sensor no longer functions, i.e., $\bm{\hat{s}}^4 = \bm{0}$, but joint $q_4$ is still able to move and be controlled.
At $t = t_1$, we simulate the removal of the joint torque sensor at joint $q_4$ such that $\bm{\hat{s}}^4 = \bm{0}$, but joint $q_4$ is still able to move and be controlled.
As a result, there is a large drop in sensor observability index $o$ at $t = t_1$. 
Since only $w_k$ is being maximized and there are no kinematic changes to the robot joints, the controller does not adjust to compensate for the deficient sensor, as expected. 
From $t = t_2$ onwards, the robot switches the null-space projection to optimize sensor observability $o_{sum}$ instead.
The robot begins to compensate for the sensor deficiency and increases the sensor observability index $o$ by adjusting the orientation of the remaining sensor axes, eventually trending towards a maximum at $t = t_4$.
Despite the fact that $o_{sum}$ is the target of optimization, $w_k$ only changes slightly.
This experiment demonstrates the different purposes of the kinematic manipulability and sensor observability indices and how the sensor observability indices can be used to ensure that sensor observability is maximized and that interaction forces in the task-space can be observed properly.
This ability to at least temporarily recover from such a sensor malfunction is especially critical if the robot must continue to function and is positioned beyond immediate reach for repair, for instance in teleoperated situations or for robots out in the field.

%%%%%%%%%%%%%%%%%%%%%%%%%%%%%%%%%%%%%%%%%%%%%%%%%%%%%%%%%%%%%
\subsection{Optimization Formulation} \label{sec:optQP}

Maximizing sensor observability can also be formulated as an optimization problem. 
Leveraging Lemma 1 in \cite{Dufour2020JIRS-ManipulabilityQP}, and taking into consideration that the sensor observability is inherently a positive function, we formulate the subsequent optimization problem: 
 \newcommand \conf{\bm{q}}
\begin{equation}
\label{SO_QP}
\begin{aligned}
& \min_{\bm{\dot{q}}, \bm{\delta}, \epsilon_{o}} &  & \frac{1}{2}\,\bm{\dot{q}}^{T}\bm{Q}\bm{\dot{q}} + \frac{1}{2}\bm{\delta}^{T}\bm{Q}_{\delta}\bm{\delta}+ \frac{1}{2}\,\alpha\,\epsilon_{o}^{2} \\
& \text{subject to} &&  \bm{J}\bm{\dot{q}} + \bm{\delta}= \dot{\bm{x}}\\
& &  & o - T\, (\bm{\nabla} o)^{T}\, \bm{\dot{q}} = \epsilon_{o}\\
& &  & \bm{b^{-}} \leq \bm{A}\bm{\dot{\conf}}\leq \bm{b^{+}} \\
& & & \bm{{\dot{q}}^{-}} \leq \bm{\dot{q}} \leq \bm{{\dot{q}}^{+}} \\
&&& \bm{\delta^{-}} \leq \bm{\delta} \leq \bm{\delta^{+}}
\end{aligned}
\end{equation}

\noindent where $\bm{\delta} \in \mathbb{R}^{n_t}$ is a relaxation vector\footnote{E.g., allow the position and/or orientation of the end effector to vary from the desired trajectory. Please refer to \cite{Dufour2020JIRS-ManipulabilityQP} for in-depth explanations of the relaxation vector.},  $\bm{Q}$ and $\bm{Q}_{\delta}$ are positive semidefinite matrices\footnote{$\bm{Q}_{\delta}$ is a non-constant matrix that varies with the motion and allows the relaxation vector to take on different profiles. For example, it can force the EE motion to match the desired trajectory near the beginning and end of the trajectory, but allow the EE to deviate from the desired trajectory in the middle of the motion.} defined as \cite{Dufour2020JIRS-ManipulabilityQP}, $\alpha$ is a positive coefficient, $o$ is the sensor observability index or sensor observability in a specific axis, $\bm{\nabla} o=\frac{\partial o}{\partial \bm{q}}$ is the gradient of $o$, $T$ is the time period of the robot control loop, $\vrectangleA^{-}$ and $\vrectangleA^{+}$ are respectively the lower and upper limits of any particular variable $\vrectangleA$, $\bm{A} \in \mathbb{R}^{m\times n_q}$ and $\bm{b} \in \mathbb{R}^{m}$ can serve various purposes, such as collision avoidance or ensuring that the end-effector remains within the robot's visual field \cite{Dufour2020RAS-VisualSpatialAttention}. 
$m$ is therefore dependent on the number of inequality constraints.
% $\bm{{\dot{r}}} = \left[ {\bm{v}}^T \quad {\bm{\omega}}^T \right]^T$, 

The optimization problem \eqref{SO_QP} can be reformulated as the following standard quadratic programming (QP) problem:
	
%$\begin{bmatrix} \bm{J}_{n_t \times n_q} & \bm{I}_{n_t \times n_t} & \bm{0}_{n_t \times 1}\\ T\, (\bm{\nabla} o)^{T}& \bm{0}_{1 \times n_t}&1\end{bmatrix} \begin{bmatrix} \bm{\dot{q}} \\ \bm{\delta} \\ \epsilon_{o}\end{bmatrix}  =  \begin{bmatrix} \dot{\bm{x}}  \\ o \end{bmatrix} $

\begin{equation}
 	\begin{aligned}
	& \min_{\bm{\mathcal{Z}}\in \mathbb{R}^{n_q+n_t+1}}& & \frac{1}{2}\bm{\mathcal{Z}^{T}\mathcal{Q}\mathcal{Z}} \\%+ \mathcal{G}^{T}\mathcal{Z} \\
	& \text{subject to} & & \bm{\mathcal{J}\,\mathcal{Z}} = \begin{bmatrix} \dot{\bm{x}}  \\ o \end{bmatrix}\\
	%{}\\
	& & & \bm{\mathcal{B}^{-}}\ \leq \bm{\mathcal{A}\mathcal{Z}} \leq \bm{\mathcal{B}^{+}}\\
	& & & \bm{\mathcal{Z}^{-}} \leq \bm{\mathcal{Z}}  \leq \bm{\mathcal{Z}^{+}}
	\label{SO_QP_reform}
	\end{aligned}
	\end{equation}
where:
	\begin{gather*}
	\bm{\mathcal{Z}}= \begin{bmatrix} \bm{\dot{q}} \\ \bm{\delta} \\ \epsilon_{o}\end{bmatrix},
		 \quad \bm{\mathcal{Q}} = \begin{bmatrix} \bm{Q} &\quad  \bm{0}_{n_q \times n_t} &\quad  \bm{0}_{n_q \times 1} \\ \bm{0}_{n_t \times n_q} &\quad \bm{Q}_{\delta} &\quad \bm{0}_{n_t \times 1} \\ \bm{0}_{1 \times n_q} &\quad \bm{0}_{1 \times n_t} &\quad \alpha\end{bmatrix},\\
		% \quad \mathcal{G} =  \begin{bmatrix} \bm{0} \\ \beta \end{bmatrix}  \\
	\bm{\mathcal{J}} =  \begin{bmatrix} \bm{J}_{n_t \times n_q} & \bm{I}_{n_t \times n_t} & \bm{0}_{n_t \times 1}\\ T\, (\bm{\nabla} o)^{T}& \bm{0}_{1 \times n_t}&1\end{bmatrix}, \\ 
	\bm{\mathcal{A}} =  \begin{bmatrix} \bm{A} & \bm{0}_{m \times (n_t+1)} \end{bmatrix},\quad \bm{\mathcal{B}}^+= \bm{b}^+,\quad \bm{\mathcal{B}}^-= \bm{b}^-, \\
	 \bm{\mathcal{Z}}^{+}= \begin{bmatrix} \bm{\dot{q}^{+}} \\ \bm{\delta}^{+}  \\ \epsilon_{o}^+ \end{bmatrix}, \quad \bm{\mathcal{Z}}^{-}=\begin{bmatrix} \bm{\dot{q}^{-}} \\  \bm{\delta}^{-} \\0 \end{bmatrix} 
	\end{gather*}

\noindent where $\epsilon_{o}^+ \gg 1$ is a user-defined coefficient.
As a consequence, any off-the-shelf QP solver can be used to solve \eqref{SO_QP_reform}, thereby solving the optimization problem in \eqref{SO_QP}.
The resulting robot motion, using the same scenario as Sec. \ref{sec:simulation3Rrobot}, is shown in Fig. \ref{fig:sim3dof_OptQP} and implemented using \texttt{quadprog()} in MATLAB.
There are clear differences when compared to maximizing sensor observability as a secondary task using the null space of the Jacobian (Sec. \ref{sec:nullspaceSO} and Fig. \ref{fig:sim3dof_OptSO}). 
Posing sensor observability as an optimization problem with the relaxation vector generates a different trajectory that results in a flatter $o_{sum}$ profile that has a slightly higher mean ($\bar{o}_{sum,\ opt} = 3.8856$ vs $\bar{o}_{sum,\ null} = 3.7817$) at the expense of deviating from the desired trajectory ($\bar{x}_{err}$ = 0.074 m). % and max $x_{err}$ = 0.159 m
Similar differences are obtained when optimizing sensor observability only for a single axis, as shown in Fig. \ref{fig:sim3dof_OptAxisXQP}, when compared to the null space formulation in \eqref{eq:sim3R-q0SOaxis} in Fig. \ref{fig:sim3dof_OptAxisX}.
In both cases, the relaxation vector\ $\bm{\delta}^+= -\bm{\delta}^- = [0.1,\ 0.1]^T$ allows for deviations from the prescribed path in order to allow for slightly different maximization profiles of $o$.

If the end effector should not deviate from the prescribed path, then the relaxation limits should be set close to zero: $\bm{\delta}^+= -\bm{\delta}^- = [\eta,\ \eta]^T$ with $0<\eta\ll1$. 
The resulting motion is then similar to the null space formulation, but \eqref{SO_QP} still allows for other optimization variables to be used if desired.
%\hl{change from 6 into task space axes variable $n_t$ and $n_q$}

%%%%%%%%%%%%%%%%%%%%%%%%%%%%%%%%%%%%%%%%%%%%%%%%%%%%%%%%%%%%%
%%%%%%%%%%%%%%%%%%%%%%%%%%%%%%%%%%%%%%%%%%%%%%%%%%%%%%%%%%%%%

%\section{\hl{Physical Interaction and Sensor Observability}} \label{sec:physicalexperiments}
\section{Practical Implications of Sensor Observability} \label{sec:physicalexperiments}

%\subsection{Baxter Robot Sensor Deficiency}
%\subsection{\hl{Physical Interactions and Sensor Observability Maximization}}

\begin{figure}[t]
\centering
\subfigure[]{
	\includegraphics[width = 0.225\textwidth]{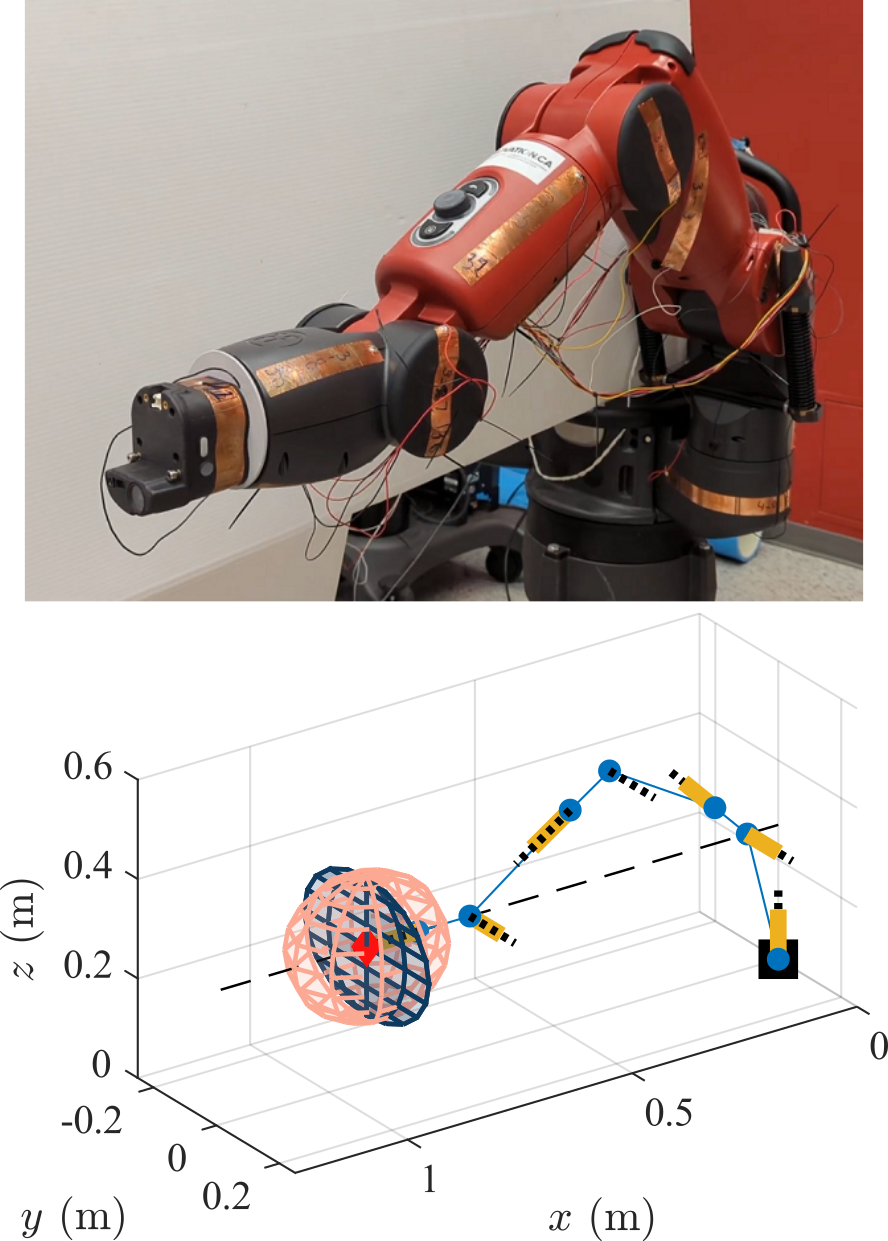}
	\label{fig:simBaxterJointDeficiencyv2A}}
\subfigure[]{
	\includegraphics[width = 0.225\textwidth]{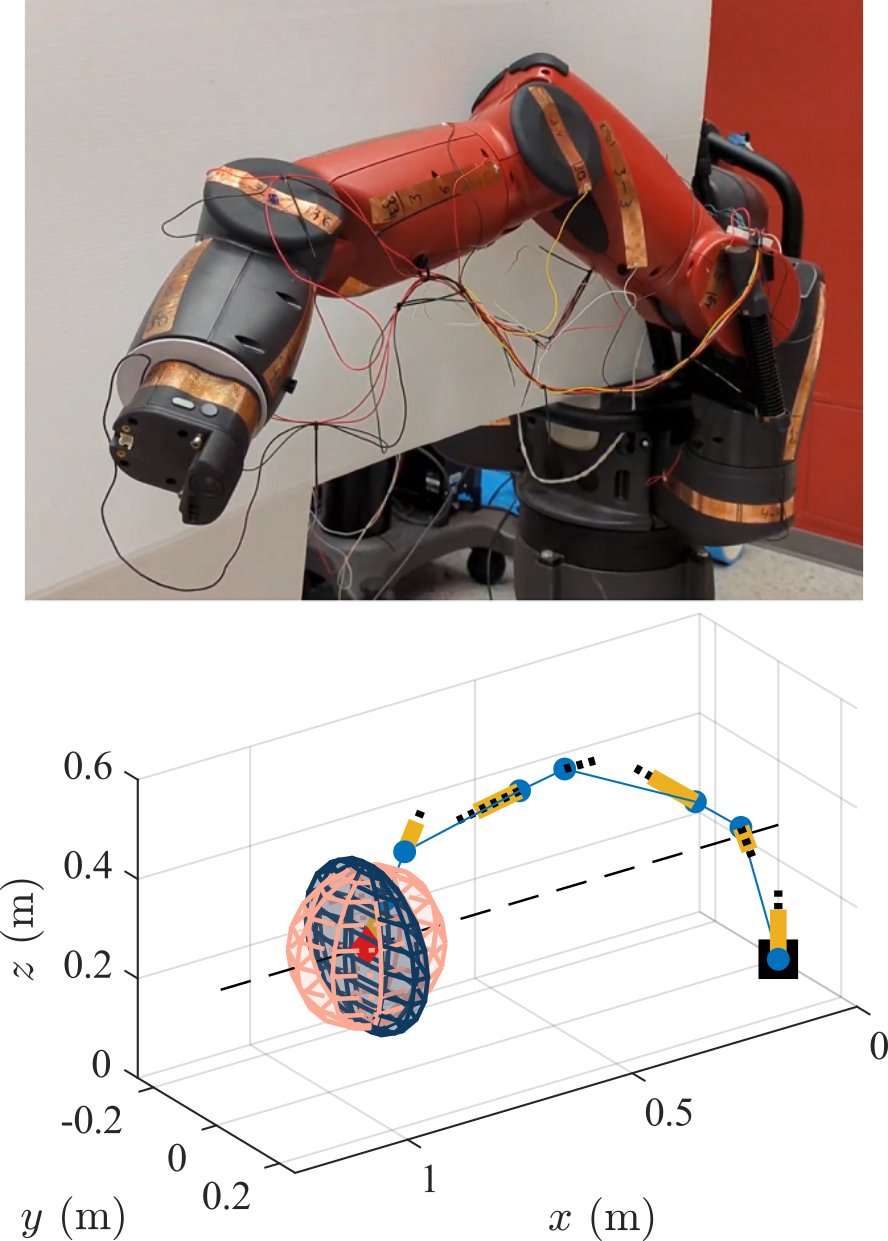}
	\label{fig:simBaxterJointDeficiencyv2B}}
\subfigure[]{
	\includegraphics[width = 0.45\textwidth]{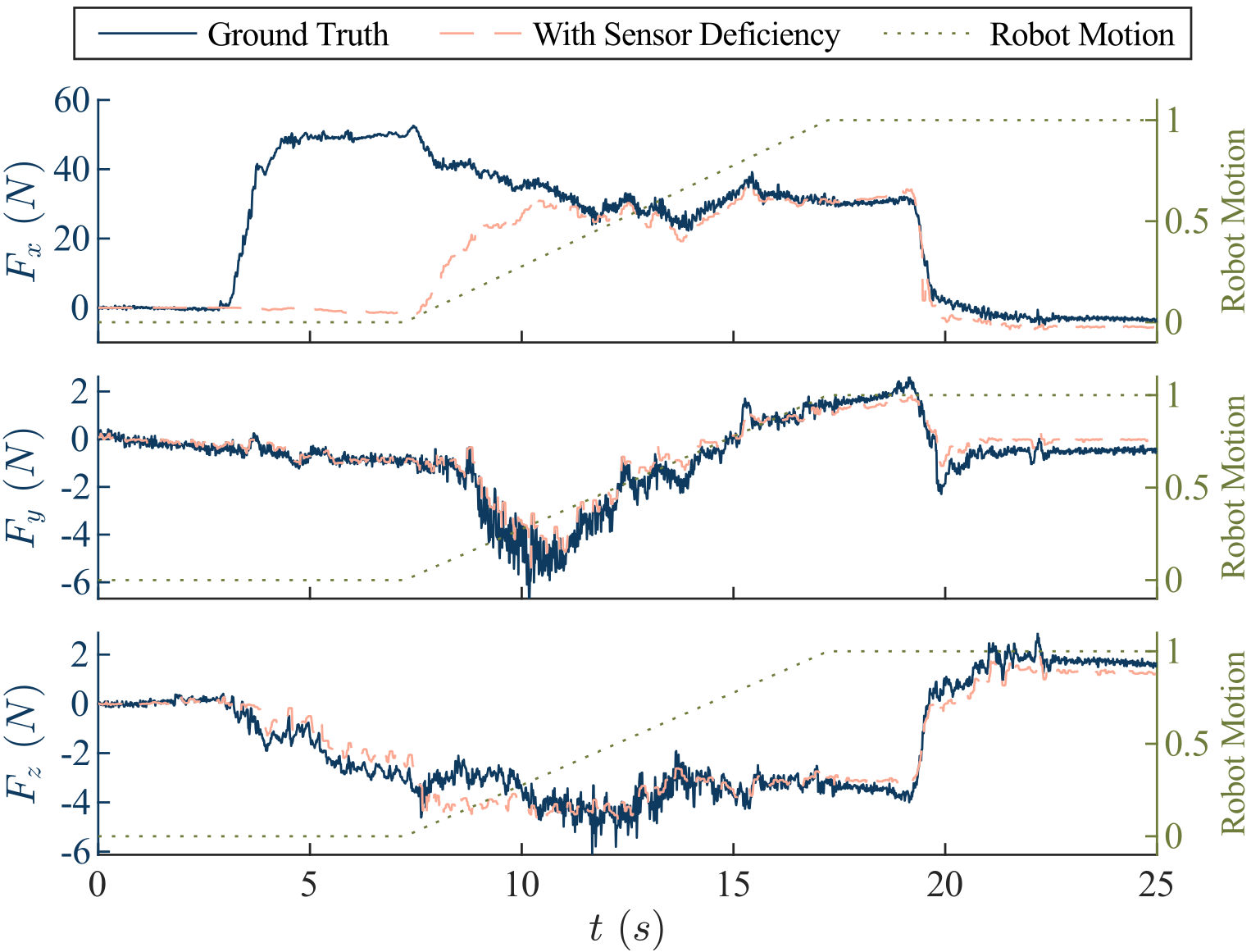}
	\label{fig:simBaxterJointDeficiencyv2-plot}}
\subfigure[]{
	\includegraphics[width = 0.45\textwidth]{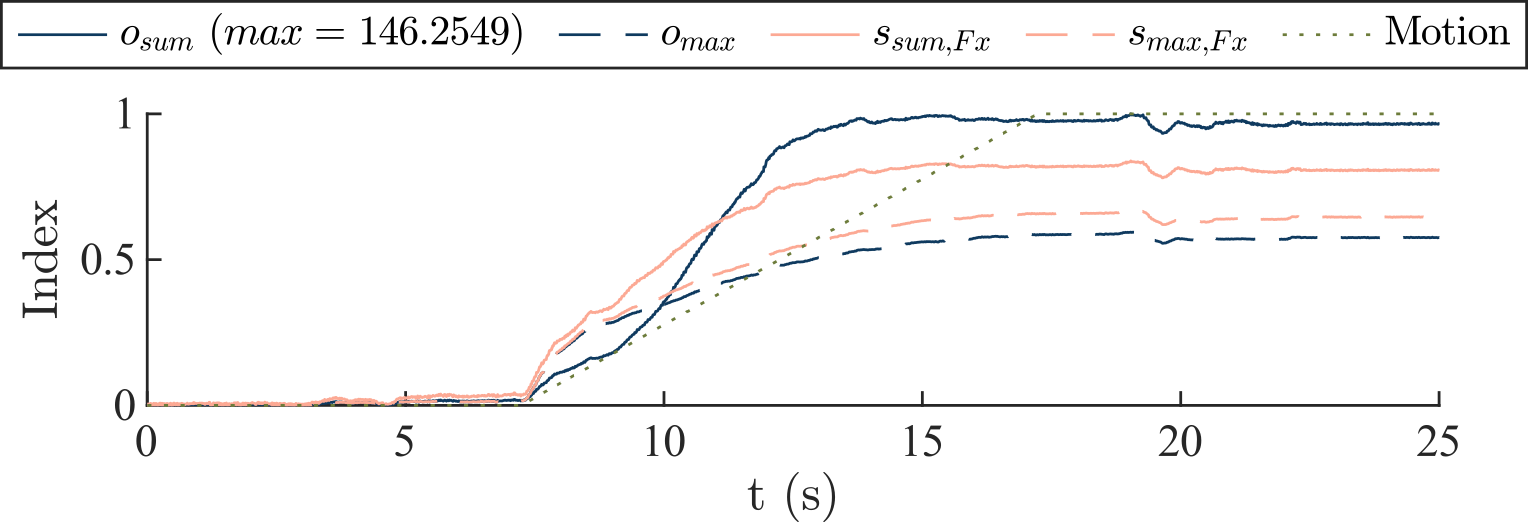}
	\label{fig:simBaxterJointDeficiencyv2-SOIplot}}
\caption{Baxter robot experiment with sensor deficiency at $\bm{\hat{s}}^4$.
Blue and pink ellipsoids are the force and torque sensor observability ellipsoids, respectively. 
(a) Initial Position. Joints 2, 6, and 7 form a line that is parallel with the $x$-axis. 
As $\bm{\hat{s}}^4$ is the only sensor that can detect forces along the $x$-axis, the robot is now in a sensor observability singular configuration as witnessed by the collapsed force sensor observability ellipsoid. 
(b) Final position for optimized sensor observability, similar to $t = t_4$ of Fig. \ref{fig:simBaxterJointDeficiencyA}.
(c) Reconstructed end effector forces with ground truth (all sensors) and with sensor deficiency (without $\bm{\hat{s}}^4$).
(d) Sensor observability index $o$ and system sensor observability vector $\bm{s}$ throughout the experiment. 
System sensor observability $\bm{s}$ values for the other axes are not shown as they are well above 0 and not of interest.
}
\label{fig:simBaxterJointDeficiencyv2}
\end{figure}

In this section, we perform a physical interaction experiment using the real Baxter robot and demonstrate the practical implications of sensor observability singularity and the effect on force reconstruction over a range of sensor observability values, seen in Fig. \ref{fig:simBaxterJointDeficiencyv2}. 
Following the previous sensor deficiency example shown in Fig. \ref{fig:simBaxterJointDeficiency}, the initial joint configuration has joints 2, 6, and 7 form a line that is parallel with the $x$-axis. 
As a result, only joint 4 is able to detect forces in $x$.
Let us say that the sensor deficiency at $\bm{\hat{s}}^4$ now occurs immediately at $t = 0$ s.
The sensor deficiency renders the current initial sensor configuration into a sensor observability singularity as none of the remaining sensors are capable of sensing forces along the $x$-axis, which is shown by the collapsed force sensor observability ellipsoid in Fig. \ref{fig:simBaxterJointDeficiencyv2A}.
In this experiment, forces are applied to the end effector as the robot moves from the sensor observability singular configuration in Fig. \ref{fig:simBaxterJointDeficiencyv2A} to a more optimized pose shown in Fig. \ref{fig:simBaxterJointDeficiencyv2B}.

The goal is to reconstruct the end effector forces $\bm{F}_{EE}$ using the equation $\bm{\tau} = \bm{J}^T \bm{F}_{EE}$, where the joint torques $\bm{\tau}$ and Jacobian transpose $\bm{J}^T$ are known. 
The deficiency is modelled by removing elements related to the 4th joint in the \emph{deficient} joint torque vector $\bm{\tau}_{def} = \begin{bmatrix} \tau_1 & \tau_2 & \tau_3 & \tau_5 & \tau_6 & \tau_7 \end{bmatrix}$ and the corresponding 4th column to obtain the \emph{deficient} Jacobian $\bm{J}_{def} \in \mathbb{R}^{6 \times 6}$.
A least squares approximation of end effector forces is performed using the remaining joint torque readouts $\bm{\tau}_{def} = \bm{J}_{def}^T \bm{F}_{EE}$ and compared to the ground truth using all sensors $\bm{\tau} = \bm{J}^T \bm{F}_{EE}$. 
The MATLAB function \texttt{lsqr()} is used with a higher tolerance and a preconditioner matrix to stabilize against the small singular values of $\bm{J}_{def}$.
The evolution of the reconstructed end effector forces as the robot transitions between the two poses is shown in Fig. \ref{fig:simBaxterJointDeficiencyv2-plot}.

As expected, $\bm{\tau}_{def}$ and $\bm{J}_{def}$ are unable to reconstruct the forces in $x$ in the sensor observability singular configuration from $t = $ 0 s to $t = $ 7.2 s.
Conversely, the forces in $y$ and $z$ are correctly reconstructed throughout the entirety of the experiment. 
Starting from $t = 7.2$ s, sensor observability optimization repositions the remaining sensors to compensate for the deficient one and regain sensing in the $x$-axis and completes the transition at $t = 17.2$ s.
As the robot moves away from the singular position, forces in $x$ slowly become visible starting from $t = 7.6$ s (4\% into the robot motion and $s_{sum, F_x} \approx 0.14$).
The reconstructed forces in $x$ eventually coincide with the ground truth starting from $t = 11.4$ s.
From $t = 11.4$ s onwards, all forces are reconstructed properly for the sensor deficient case. 
In Fig. \ref{fig:simBaxterJointDeficiencyv2-SOIplot}, the sensor observability indices and system observability vectors show that they are all close to 0 until the robot begins to move away from the singular configuration.
While $F_x$ is fully reconstructed when $s_{sum, F_x} \approx 0.66$ at $t = 11.4$ s, this value does not necessarily represent a universal threshold and could be unique to this particular robot, sensor type, joint configuration, experiment, and reconstruction method.

%\begin{equation}
%\begin{aligned}
%\bm{S} & = \begin{bmatrix} \bm{\tilde{s}}^1 & \bm{\tilde{s}}^2 & \bm{\tilde{s}}^3 & \bm{\tilde{s}}^4 & \bm{\tilde{s}}^5 & \bm{\tilde{s}}^6 & \bm{\tilde{s}}^7 \end{bmatrix} \\
%\bm{S}_{def} & = \begin{bmatrix} \bm{\tilde{s}}^1 & \bm{\tilde{s}}^2 & \bm{\tilde{s}}^3 & \bm{0} & \bm{\tilde{s}}^5 & \bm{\tilde{s}}^6 & \bm{\tilde{s}}^7 \end{bmatrix}
%\label{eq:SensDef-1}
%\end{aligned}
%\end{equation}

%%%%%%%%%%%%%%%%%%%%%%%%%%%%%%%%%%%%%%%%%%%%%%%%%%%%%%%%%%%%%
%%%%%%%%%%%%%%%%%%%%%%%%%%%%%%%%%%%%%%%%%%%%%%%%%%%%%%%%%%%%%
\section{Conclusion} \label{sec:concl}

%I think the background should be shortened in the conclusion, explaining just the drawbacks of the conventional methods that are solved in this paper. Instead, clarify the limitations of the paper and explain how those issues will be addressed in the future work, at the last part of the conclusion. As noted above, change the paragraph and put more details on the ongoing and future work.

In this work, we introduce the novel concept of the sensor observability for analysing the quality of a specific joint configuration for observing task-space quantities.
We believe that this is the first work in quantifying the cumulating effect of distributed axial sensor positioning in multi-DOF articulated robots and provides a novel performance metric as well as the base framework for developing further tools related to sensor analysis.
%\hl{This metric quantifies the alignment of onboard sensors to determine whether task}
In special cases related to force sensing, there exists parallels between traditional kinematics analysis and the proposed sensor observability analysis, but sensor observability has certain advantages related to generalization. 
A deeper analysis shows the need to distinguish between the two and use sensor observability to augment kinematic manipulability, particularly in irregular robot structures where joints and sensors do not have a one-to-one mapping.
While sensor observability analysis is most intuitively applied to force sensing, the concept may potentially be applied to other axial sensors such as accelerometers or distance sensors.

Future work, as mentioned throughout the paper, will include further generalization of the concept to include other sensor types, unidirectional sensors, and sensor performance at joint limits.
The concept will also be extended to multi-contact robots, multi-limbed robots, flexible/soft robots, and floating base robots.
While we have demonstrated the optimization of task-space observability along a single direction, future work would extend the concept for sensor observabiliy ellipsoid shaping to achieve specific sensor observability profiles, as shown in \cite{Jaquier2021IJRR-GeometryAwareManipulabilityLearning} for the manipulability ellipsoid shaping.
Additionally, possible parallels between sensor observability versus controllability and observability in the state-space sense will also be explored in future work.
Lastly, sensor observability analysis has the potential to be used in the robot design phase to optimize the placement of sensors to create redundancy or minimize the number of sensors required and will be the subject of future work.
%\hl{more formal comparisons with conventional methods}

%Furthermore, observability is currently frame dependent, but yet certain configurations, while ... ... \hl{this is not true if it's in EE frame (and not in world frame)}.

%\section*{Acknowledgements}
%We would like to thank the Fonds de Recherche du Quebec - Nature et technologies and Mitacs Globalink Research Award for supporting in part this research.

\begin{comment}
Acknowledgements:
At the end of the article one or more statements should specify
(a) contributions that do not justify authorship;
(b) technical help;
(c) financial and material support, specifying the nature of the support;
(d) financial relationships that may pose a conflict of interest. 
\end{comment}
%funding provided by the Natural Sciences and Engineering Research Council of Canada (NSERC) and the Fonds de recherche du Qu\'ebec - Nature et technologies (FRQNT)

%\IEEEtriggeratref{8}
% The "triggered" command can be changed if desired:
%\IEEEtriggercmd{\enlargethispage{-5in}}
\bibliographystyle{IEEEtran}
\bibliography{HumanoidReferences}

%\begin{IEEEbiography}[{\includegraphics[width=1in,height=1.25in,clip,keepaspectratio]{figures/ChrisWong}}]{Christopher Yee Wong} (M'15)
 %received his B.Eng (2011) and M.Eng. (2014) from McGill University in Montreal, Canada and Ph.D. (2017) in mechanical engineering from University of Toronto in Toronto, Canada. He received postdoctoral fellowships to perform research at AIST in Tsukuba, Japan (2018-2019), at LIRMM in Montpellier, France (2019), and at l'Universit\'{e} de Sherbrooke in Sherbrooke, Canada (2019-2023). He joined McGill University as a research associate in 2023, where his research interest focuses on physical human-robot interaction using humanoids, particularly on improving methods for robot cognition with regards to safety and intention detection related to human touch.
%\end{IEEEbiography}

\begin{IEEEbiography}[{\includegraphics[width=1in,height=1.25in,clip,keepaspectratio]{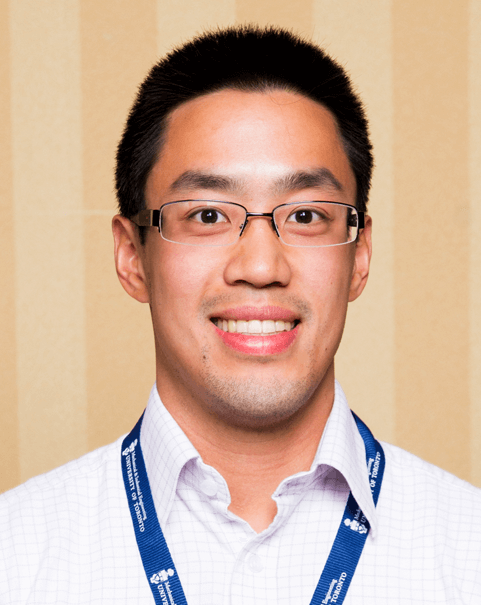}}]{Christopher Yee Wong} (M'15)
 received his B.Eng ('11) and M.Eng. ('14) from McGill University in Montreal, Canada and Ph.D. ('17) in mechanical engineering from Univ. of Toronto in Toronto, Canada. He was a postdoctoral researcher at AIST in Tsukuba, Japan ('18-'19), at LIRMM in Montpellier, France ('19), and at l'Univ. de Sherbrooke in Sherbrooke, Canada ('19-'23) and is currently at McGill University since 2023. 
His research focuses on improving robot cognition in physical human-robot interaction with regards to safety, human analysis, and intention detection related to touch.
%received postdoctoral fellowships to perform research 
\end{IEEEbiography}

\begin{IEEEbiography}[{\includegraphics[width=1in,height=1.25in,clip,keepaspectratio]{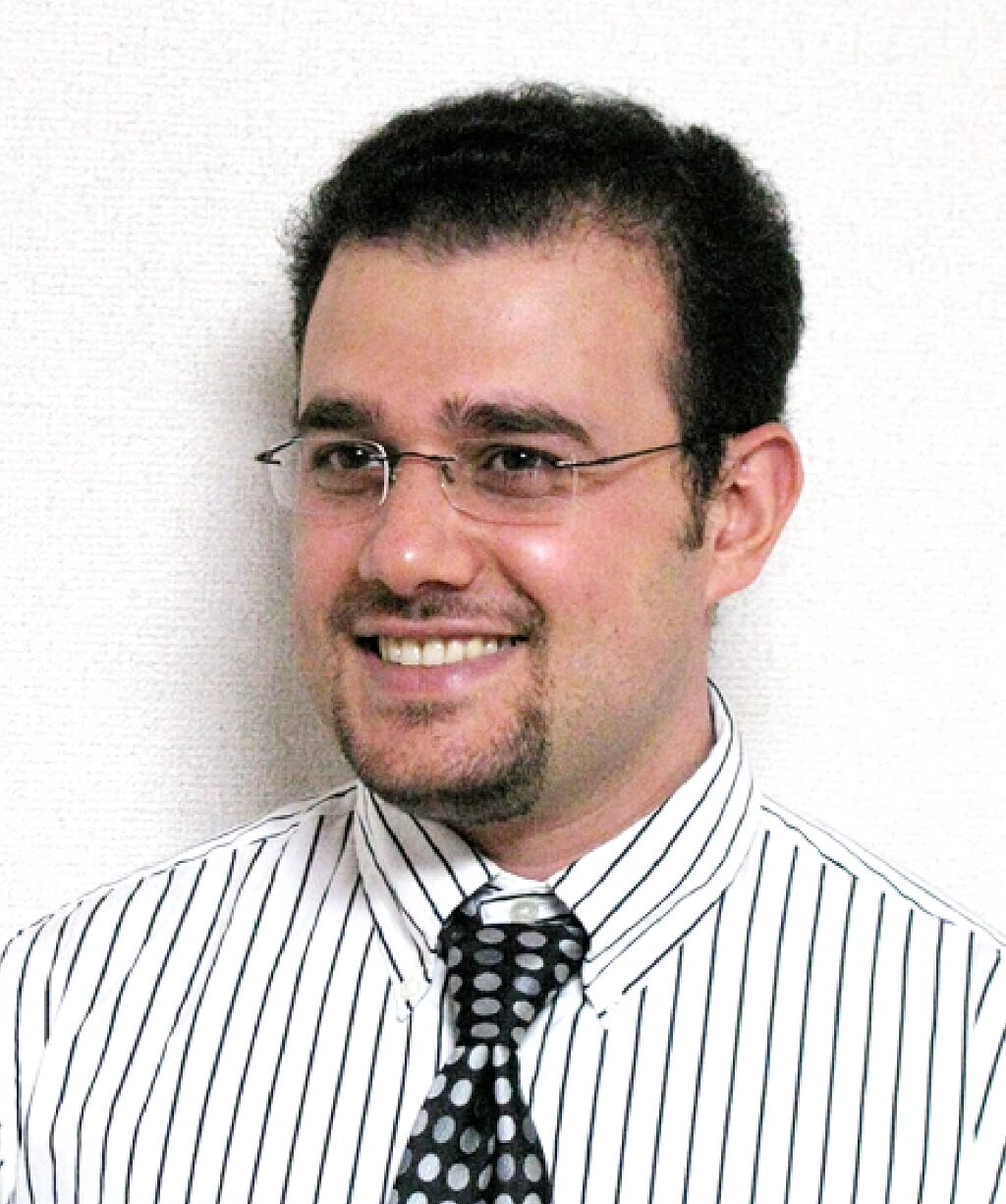}}]{Wael Suleiman} received the Master's and Ph.D. degrees in automatic control from Paul Sabatier University, Toulouse, France in 2004 and 2008, respectively. He has been Postdoctoral researcher at AIST, Tsukuba, Japan from 2008 to 2010, and at Heidelberg University, Germany from 2010 to 2011. He joined University of Sherbrooke, Quebec, Canada, in 2011, and is currently Full Professor at Electrical and Computer Engineering Department. His research interests include collaborative and humanoid robots, motion planning, nonlinear system identification and control and numerical optimization. 
\end{IEEEbiography}

%%%%%%%%%%%%%%%%%%%%%%%%%%%%%%%%%%%%%%%%%%%%%%%%%%%%%%%%%%%%%
%%%%%%%%%%%%%%%%%%%%%%%%%%%%%%%%%%%%%%%%%%%%%%%%%%%%%%%%%%%%%
\appendix{
\subsection{Sensor observability threshold for sensor noise}
\label{app:sensornoise}

\begin{figure}[!ht]
\centering
\subfigure[]{
	\includegraphics[width = 0.15\textwidth]{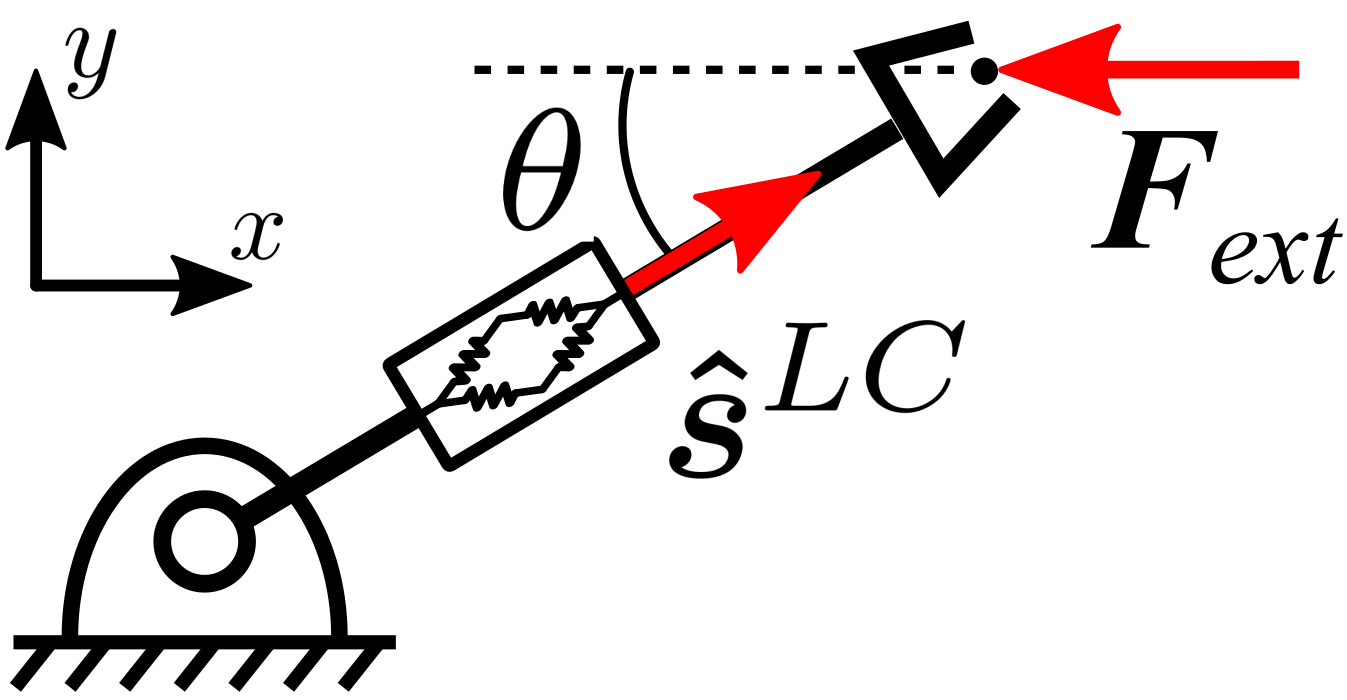}
	\label{fig:app-fig}}
\subfigure[]{
	\includegraphics[width = 0.30\textwidth]{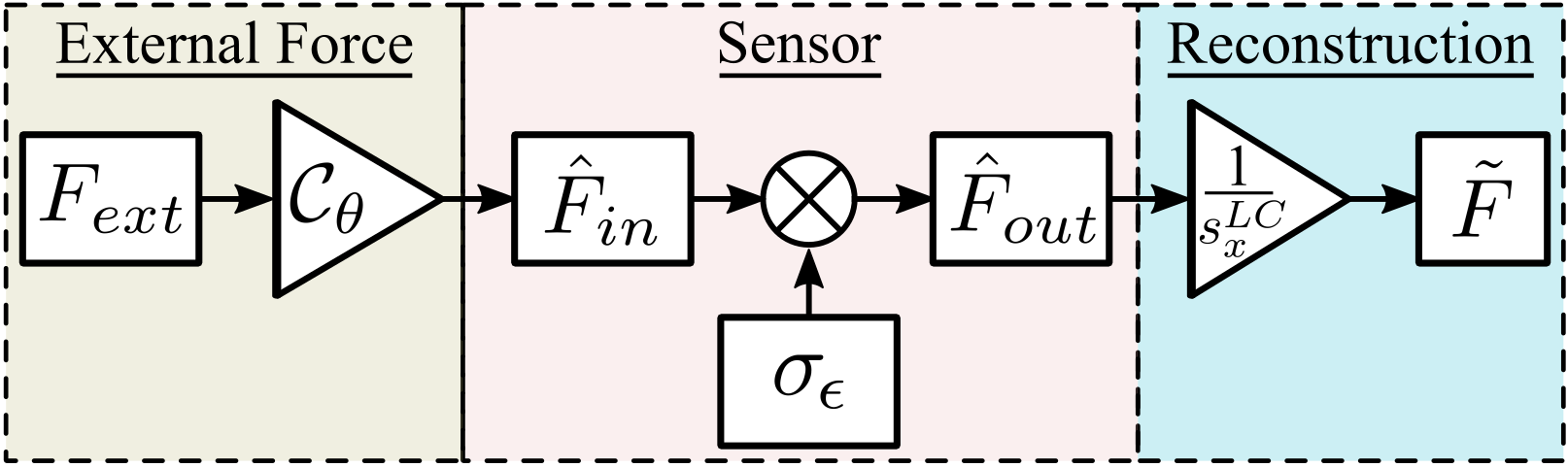}
	\label{fig:app-chart}}
\caption{a) Simplified example for the derivation of sensor observability thresholding in the $x$-axis; b) flowchart of the formulation.}
\label{fig:appendix}
\end{figure}

To illustrate the derivation of the sensor observability threshold, we examine the scenario shown in Fig. \ref{fig:appendix} where an external force $\bm{F}_{ext}$ is applied to the 1DOF robot with a single load cell $\bm{\hat{s}}^{LC}$ that is aligned with the link.
The analysis only considers the $x$-axis for simplicity.
%The formulation follows the flowchart shown in Fig. \ref{fig:app-chart}.
The force projected along the axis of the load cell, which is seen by the sensor, is $\hat{F}_{in} = F_{ext} \lvert\mathcal{C}_\theta\rvert$, where $\mathcal{C}_\theta = \text{cos}(\theta)$.
The output signal of the load cell $\hat{F}_{out}$ is corrupted by sensor noise $\sigma_{\epsilon}$, a property of the sensor hardware, resulting in $\hat{F}_{out} = \hat{F}_{in} + \sigma_{\epsilon}$.
In order to reconstruct the applied external force from the sensor output, the output of the sensor must be divided by the angle offset of the sensor $\mathcal{C}_\theta$.
The reconstructed external force $\tilde{F}$ is then:

\begin{equation}
\tilde{F} = \frac{\hat{F}_{out}}{\lvert\mathcal{C}_\theta\rvert} = \frac{\hat{F}_{in} + \sigma_{\epsilon}}{\lvert\mathcal{C}_\theta\rvert} = \frac{F_{ext}\lvert\mathcal{C}_\theta\rvert + \sigma_{\epsilon}}{\lvert\mathcal{C}_\theta\rvert} = F_{ext} + \frac{\sigma_{\epsilon}}{\lvert\mathcal{C}_\theta\rvert}
\end{equation}

As the alignment of the sensor axis is equivalent to the angle offset, i.e., $\lvert\mathcal{C}_\theta\rvert = s^{LC}_x$, the second term is then rewritten as: %$\frac{\sigma_{\epsilon}}{s^{LC}}$, resulting in the following: 

\begin{equation}
\tilde{F} = F_{ext} + \frac{\sigma_{\epsilon}}{s^{LC}_x}
\end{equation}

Thus, sensing of the external force and reconstructing it precisely is highly dependent on the degree of alignment between $F_{ext}$ and $s^{LC}_x$.
If the alignment is poor, then the noise term $\sigma_{\epsilon}$ will be amplified by $\frac{1}{s^{LC}_x}$ and mask any small $F_{ext}$.
Thus, for a given alignment $s^{LC}_x$, $\Phi$ is considered the minimum force that can be detected. 

\begin{equation}
\Phi = \frac{\sigma_{\epsilon}}{s^{LC}_x}
\label{eq:app-init-phi}
\end{equation}

By considering (\ref{eq:app-init-phi}) in reverse, we set $\Phi_{min}$ as the minimum that must be detected and then determine the sensor observability threshold $s^{LC,*}_x$:
%Additionally, we define the sensor noise $\sigma_{\epsilon}$ as the standard deviation of the electrical signal noise $\sigma_{\epsilon}$ divided by the sensitivity of the sensor $\kappa$: %, thus (\ref{eq:app-final-force}) becomes:

\begin{equation}
s^{LC,*}_x = \frac{\sigma_{\epsilon}}{\Phi_{min}}% = \frac{\sigma_{\epsilon}}{\kappa \Phi_{min}} %= \frac{\sigma^2}{\kappa \Phi_{min}}
\label{eq:app-final-force}
\end{equation}

While this derivation is for load cells, formulations for other sensor types may differ and will be explored in future work.

%When considering moment arms, basically have $\hat{F}_{in} = F_{ext}\text{cos}(\theta) = F_{ext}s$ where $F_{ext} = F \times r$

%\begin{equation}
%s^{LC,*}_x = \frac{\sigma^2}{\kappa \Phi_{min}}
%\end{equation}

%\hl{WIP}

%=====================
%Variance of the signal noise $\sigma^2$ divided by the sensitivity of the sensor $\kappa$ results in the noise in units of force 
%Let us say that any minimum reading $\Phi$ must be greater than the amplified amplitude of $\frac{F_\epsilon}{s^{LC}}$.

%\begin{equation}
%\begin{gathered}
%\hat{F}_{out} = \hat{F}_{in} + F_\epsilon \\
%\tilde{F} = \frac{\hat{F}_{out}}{\text{cos}(\theta)} = \frac{\hat{F}_{out}}{s_j} \\
%\therefore \tilde{F}s = \hat{F}_{out} \\
%\text{but } \hat{F}_{in} = F_{ext}\text{cos}(\theta) = F_{ext}s \\ 
%\tilde{F}s = F_{ext}s + F_\epsilon \\
%\tilde{F} = F_{ext} + \frac{F_\epsilon}{s} = F_{ext} + \frac{\sigma^2/\kappa}{s}
%\end{gathered}
%\label{eq:appendix1}
%\end{equation}
} %end of appendix

\end{document}